\documentclass{article}
\usepackage{spconf,amsmath,graphicx}
\usepackage{algorithm}
\usepackage[noend]{algpseudocode}
\usepackage{mathtools}
\usepackage{url}
\usepackage{makecell}
\usepackage{subcaption}
\usepackage{amssymb}
\usepackage{amsfonts}
\usepackage{dsfont}
\usepackage{amsthm}
\usepackage{multirow}

\DeclareMathOperator*{\minimize}{minimize}

\DeclareMathOperator*{\argmin}{arg\,min}
\DeclareMathOperator*{\argmax}{arg\,max}

\usepackage{wasysym}

\theoremstyle{definition}

\title{Reverse Engineering Imperceptible Backdoor Attacks on Deep Neural Networks for Detection and Training Set Cleansing}
\name{Zhen Xiang, David J. Miller and George Kesidis\thanks{Supported by an AFOSR DDDAS grant and Cisco URP gift.}
}
\address{Anomalee Inc., State College, PA\\
 Pennsylvania State University, University Park, PA\\
\{zux49,djm25,gik2\}@psu.edu}
\begin{document}
%\ninept
%
\maketitle
\begin{abstract}
Backdoor data poisoning is an emerging form of adversarial attack usually against deep neural network image classifiers. The attacker poisons the training set with a relatively small set of images from one (or several) source class(es), embedded with a backdoor pattern and labeled to a target class. For a successful attack, during operation, the trained classifier will: 1) misclassify a test image from the source class(es) to the target class whenever the same backdoor pattern is present; 2) maintain a high classification accuracy for backdoor-free test images. In this paper, we make a break-through in defending backdoor attacks with {\it imperceptible} backdoor patterns (e.g. watermarks) {\it before/during the training phase}. This is a challenging problem because it is a priori unknown which subset (if any) of the training set has been poisoned. We propose
% Zhen -- delete "the first unsupervised approach"
% the first unsupervised approach, 
an optimization-based reverse-engineering defense, that jointly: 1) detects whether the training set is poisoned; 2) if so, identifies the target class and the training images with the backdoor pattern embedded; and 3) additionally, reversely engineers an estimate of the backdoor pattern used by the attacker. In benchmark experiments on CIFAR-10, for a large variety of attacks, our defense achieves a new state-of-the-art by reducing the attack success rate to no more than 4.9\% after removing detected suspicious training images.
\end{abstract}
\section{Introduction}\label{sec:intro}

As deep neural network (DNN) classifiers have achieved state-of-the-art performance in many research areas, they have also been shown to be vulnerable to adversarial attacks \cite{Szegedy_seminal}. This has inspired adversarial learning research, including work in devising both formidable attacks as well as defenses against same \cite{Review}. Perhaps the most well-known adversarial attack is a test-time evasion (TTE) attack, where the attacker optimizes an image-specific, human-imperceptible perturbation for a given test image, in order to induce misclassifications during the classifier's operation \cite{FGSM,CW,ADA}.

Recently, a new form of adversarial attack, a backdoor data poisoning (DP) attack, was proposed, aiming to have a deep neural network (DNN) image\footnote{While the image domain is the focus of most works, backdoor attacks for other data domains have also been considered, e.g. \cite{Trojan,LSTMBD}.} classifier learn to (mis)classify to a target class during its operation, whenever a backdoor pattern (used to poison the training set) is present in a test image from one or several source classes \cite{BadNet,Targeted,Trojan}. 
% Zhen -- removed "note that"
% Note that 
The backdoor pattern, the source class(es), and the target class are all specified by the attacker. The attacker's goal is typically achieved by poisoning the training set of the classifier with a relatively small set of ``backdoor training images'' -- these images are originally from the source class(es) but with the backdoor pattern embedded and labeled to the target class.

Compared with traditional DP attacks, e.g. \cite{Xiao15,Tygar11}, which aim to degrade accuracy of the trained classifier, a successful backdoor attack will not degrade accuracy on clean test images. This is ensured by the fact that the vast majority of the training set is backdoor-free and the DNN has a large ``capacity'' -- thus, the DNN can learn to recognize the backdoor pattern without compromising accuracy on clean images. Hence, validation set accuracy degradation, e.g. \cite{Roni}, cannot be reliably used for detecting backdoors. Compared with TTE attacks which require knowledge of the victim classifier or a surrogate classifier \cite{Papernot3}, backdoor attacks can be devised with few requirements -- only the ability to poison the training set. Such poisoning capability is facilitated by the need in practice to obtain ``big data'' suitable for accurately training a DNN
% Zhen -- removed the following
% for a given domain
-- to do so, the training authority may need to seek data from as many sources as possible (some of which could be attackers). Thus, backdoor attacks are indeed emergent threats to deep learning systems.

Existing defenses against backdoor attacks are deployed either before/during training or post-training. In the post-training defense scenario, the defender is assumed to be the user of the DNN classifier who has no access to the training phase (including the training set). But a small, independent set of clean images is available to the defender to detect whether the classifier has been backdoored \cite{FP,NC,Post-TNNLS,Post-TIFS,STRIP}. In this paper, we consider the before/during training defense scenario, where the defender has full control of the training process, but none of the training set is guaranteed to be clean (i.e. without the backdoor pattern) \cite{SS,AC,CI}. This problem is challenging because the defender needs to not only detect if the training set is poisoned, but also to accurately identify and remove backdoor training images if there are any -- it is practically infeasible to reject the entire training set. Since there is no independent/held-out clean dataset as assumed for the post-training scenario, there is no reliable baseline reference for the ``behavior'' of the clean samples. Moreover, liberally removing suspicious training images, which could possibly cause a high fraction of clean training images to be falsely removed, should be avoided; otherwise, the performance of the resulting trained classifier will be degraded.

We propose an unsupervised, optimization-based reverse-engineering defense against backdoor attacks with human-imperceptible backdoor patterns that are widely considered in existing attacks \cite{Targeted,Haoti} and defenses \cite{SS,AC,Post-TNNLS}. We detect whether the training set is poisoned and the target class (if there is poisoning) by reverse engineering the backdoor pattern used by the attacker. This estimated pattern is further used to identify the backdoor training images. Our defense is much more effective than existing defenses for the before/during training defense scenario as shown by our experiments. After cleaning the poisoned training set (by removing the detected backdoor training images) and retraining, the attack success rate is reduced to no more than 4.9\%.

\vspace{-0.025in}
\section{Threat Model}\label{sec:threatmodel}

\subsection{Backdoor Pattern}

To launch a backdoor attack, the attacker needs to first specify a target class $t^{\ast}$, a set of source class(es) ${\mathcal S}^{\ast}$, and, most importantly, a backdoor pattern ${\bf v}^{\ast}$. Similar to the stealthiness required by TTE attacks, a backdoor pattern, when embedded into clean test images during operation, should not be easily noticeable to humans. Hence {\bf effective backdoor patterns in existing works} mainly fall into two categories: 1) a human-imperceptible, additive perturbation applied to clean images (dubbed here an {\it imperceptible} backdoor pattern), e.g. a watermark \cite{Targeted,Haoti,SS,CI,Post-TNNLS}; 2) a seemingly innocuous object in a scene (dubbed here a {\it perceptible} backdoor pattern), e.g. a pair of glasses on a face \cite{Targeted,BadNet,NC,Tabor}. In this paper, we consider backdoor attacks with an {\it imperceptible backdoor pattern}. Such pattern can be easily embedded in a clean image ${\bf x}$ by the following function:
\begin{equation}\label{eq:bd_embed}
m({\bf x}; {\bf v}^{\ast}) = [{\bf x}+{\bf v}^{\ast}]_c
\end{equation}
where $[\cdot]_c$ is a clipping operation that constrains the pixel intensity values to their valid range. The visual stealthiness of an imperceptible backdoor pattern is usually achieved by setting a very small maximum perturbation size $||{\bf v}^{\ast}||_{\infty}$ if the pattern is global (i.e. image-wide) \cite{Targeted,Haoti,Post-TNNLS}; or, in the case of a local pattern, setting a small $||{\bf v}^{\ast}||_{0}$, i.e. perturbing just a few pixels \cite{SS,CI,Post-TNNLS}. 
% Zhen -- removed the figures and refer to the backdoor patterns used by our experiments instead
%As an example, Figure \ref{fig:bd_example} shows two ``truck'' images embedded with a global ``chessboard'' pattern and a local ``cross'' pattern, respectively, and the originally clean ``truck'' image. The backdoor embedding is barely visible to humans.
Our experiments address most of the {\it imperceptible} backdoor patterns that have been considered by existing works (see Figure \ref{fig:bd_pattern}).
%DJM -- changed involve to "address"

%\begin{figure}
%	\centering
%	\begin{minipage}[b]{.3\linewidth}
%		\centering
%		\centerline{\includegraphics[width=\linewidth]{figures/bd_image_cb.png}}
%		\subcaption{``chessboard''}
%	\end{minipage}
%	\begin{minipage}[b]{.3\linewidth}
%		\centering
%		\centerline{\includegraphics[width=\linewidth]{figures/bd_image_cross.png}}
%		\subcaption{``cross''}
%	\end{minipage}
%	\begin{minipage}[b]{.3\linewidth}
%		\centering
%		\centerline{\includegraphics[width=\linewidth]{figures/bd_image_original.png}}
%		\subcaption{clean}
%	\end{minipage}
%	\caption{Example images embedded with a ``chessboard'' pattern, a ``cross'' , and the original clean (backdoor-free) training image.}
%	\label{fig:bd_example}
%	\vspace{-0.2in}
%\end{figure}

\subsection{Backdoor Data Poisoning}

Our proposed defense in this paper is designed to target standard backdoor attacks against DNN images classifiers, launched by poisoning the training set \cite{Targeted,BadNet}. To be specific, the actual (poisoned) training set used for the victim classifier's training is the union of the benign training set ${\mathcal D}_{\rm train}$ and a small ``backdoor set'' ${\mathcal B}=\{(m({\bf x}; {\bf v}^{\ast}),y)|{\bf x}\sim P_{{\mathcal S}^{\ast}}, y=t^{\ast}\}$, which is a set of backdoor training images embedded with the backdoor pattern ${\bf v}^{\ast}$, labeled to the target class $t^{\ast}$, and with the originally clean images (collected by the attacker) following $P_{{\mathcal S}^{\ast}}$, i.e. the distribution of images from the source class(es) ${\mathcal S}^{\ast}$. %Note that we reasonably assume the poisoning capability of the attacker who could possibly be a contributor of the training data, we do not assume that the attacker has access to the training data contributed by other sources. Hence the clean version of each backdoor training image ${\mathcal B}$ is likely not appearing in ${\mathcal D}_{\rm train}$.

We notice that recently there are more sophisticated variants of backdoor attacks that are designed to be less noticeable to human experts \cite{Haoti,Hidden-trigger}. But these attacks require the additional knowledge of a surrogate classifier to optimize the backdoor pattern, which greatly reduces their practical feasibility. Regardless, we show that our defense performs well against one of these attacks in the supplementary material.

\subsection{Attacker's Goals}

The goals of a backdoor attacker are two-fold. First, the attacker aims to maximize the misclassification rate to the target class $t^{\ast}$ during testing, when the same backdoor pattern ${\bf v}^{\ast}$ used during training is embedded into any test image following distribution $P_{{\mathcal S}^{\ast}}$, i.e. to maximize
\begin{equation}\label{eq:goal1}
{\mathbb E}_{{\bf x}\sim P_{{\mathcal S}^{\ast}}}[{\mathds 1}(f(m({\bf x}; {\bf v}^{\ast});\Theta)=t^{\ast})],
\end{equation}
where $f(\cdot;{\Theta}):{\mathcal X}\rightarrow{\mathcal Y}$ is the DNN classifier with parameters $\Theta$ that maps from the image domain ${\mathcal X}$ to the corresponding set of class labels ${\mathcal Y}$, and ${\mathds 1}(\cdot)$ is a logical indicator function. Second, for any clean test image ${\bf x}$ (ground truth) from any class $y\in{\mathcal Y}$ (i.e. following image distribution $P_y$), the classifier is supposed to make a correct classification, when there is no backdoor pattern embedded, i.e. to maximize
\begin{equation}\label{eq:goal2}
{\mathbb E}_{{\bf x}\sim P_y} [{\mathds 1}(f({\bf x};\Theta)=y)], \quad \forall y\in{\mathcal Y}.
\end{equation}
Although the attacker does not have access to the training process, these two goals can be jointly and automatically achieved by regular training using the poisoned training set:
\begin{equation}\label{eq:bd_training}
\minimize_{\Theta} \,\sum_{({\bf x},y)\in{\mathcal D}_{\rm train}\cup{\mathcal B}} l({\bf p}({\bf x};\Theta),y)
\end{equation}
Here, ${\bf p}({\bf x};\Theta)$ maps from the image domain ${\mathcal X}$ to the $(K-1)$-simplex, where $K=|{\mathcal Y}|$ is the total number of classes -- its $i$-th element $p_i({\bf x};\Theta)$ represents the posterior probability of class $i$. $l(\cdot,\cdot)$ is the training loss function, e.g. the cross-entropy loss. Clearly, the two goals of the attacker rely on the training involving samples from both ${\mathcal B}$ and ${\mathcal D}_{\rm train}$, respectively. %Moreover, our reverse engineering defense leverages the mechanism how the ``backdoor mapping'', i.e. Eq. (\ref{eq:goal1}), is learned, as will be detailed in Section XXX.

\subsection{Defender's Assumptions and Goals}

In this paper, we focus on backdoor defenses before/during the classifier's training phase, with the following assumptions about the defender: 1) The defender has full control of the training process. 2) The defender has full access to the entire training set, but none of the training samples are guaranteed to be backdoor-free. 3) The defender does not possess any independent clean dataset free of backdoor patterns. These assumptions are widely adopted by many existing backdoor defenses before/during training \cite{SS,AC,CI}, and even for some traditional DP defenses.

The ultimate goal of the defender is to help provide the user with a classifier without backdoors; and the classifier should perform well during its operation. Hence the defender's goals are also two-fold: 1) {\bf detection} -- detect whether the training set is poisoned; 2) {\bf training set cleansing} -- accurately, with high sensitivity and specificity, identify and remove the backdoor training images if there are any. Note that the defender should limit the number of clean images being falsely removed, such that there are sufficient clean images on which adequate training can be performed.

\vspace{-0.025in}
\section{Related Works}\label{sec:relatedwork}

Most existing backdoor defenses deployed before/during training first train a classifier using the possibly poisoned training set. The ``spectral signature'' (SS) defense projects the internal layer (e.g. the penultimate layer) activations of the training images, extracted using the trained classifier, for each class, onto the principal eigenvector and removes the outliers \cite{SS}. As a seminal work in this field, SS does not involve any detection of backdoor attacks or of the target class being poisoned. The ``activation clustering'' (AC) defense \cite{AC}, however, involves both backdoor detection (based on Silhouette score) and training set cleansing. For each putative target class, the penultimate layer activations of the associated training images are projected onto a low dimensional (e.g. 10-dimensional) space  using PCA and clustered using k-means with $k=2$. If the activations are well-fitted by the two-cluster model (evaluated by comparing with a threshold on the Silhouette score), a backdoor attack is detected and this class is deemed the target class of the attack. Then, for the detected target class, the training images associated with the cluster with the smaller mass are identified as the backdoor training images and are removed before retraining. However, in many cases, the backdoor training images are not clearly separable (using e.g. 2-means) from the clean training images labeled to class $t^{\ast}$ in the projected low-dimensional space of internal layer features.

In comparison, a more recent ``cluster impurity'' (CI) defense first models the penultimate layer activations (without projection to low-dimensional spaces) for each class using Gaussian mixture models (GMM), with the number of Gaussian components determined by Bayesian information Criterion (BIC) \cite{CI}. Then, for each component, the associated training images are blurred by applying an average filter and fed into the same trained classifier. If a component contains mainly backdoor training images, their embedded backdoor pattern will likely be destroyed by blurring; hence there will be a high fraction of images from this component whose predicted label is altered by blurring. Although the clustering procedure of CI allows multiple components, such that the backdoor training images may form their own clusters, the number of components selected by BIC is very sensitive to the dimension of the activations. When the dimension is too high, BIC always selects one component, even for the true target class being poisoned, as shown by our experiments. Moreover, both CI and AC require setting a detection threshold, which depends on the data domain -- while the method proposed in this paper requires no such careful threshold choice.

Although backdoor defenses deployed before/during training are not directly comparable with a backdoor defense deployed post-training, the current work is inspired by a {\it post-training} defense which detects backdoor attacks and the target class by optimizing a perturbation that induces high group misclassification fraction for each candidate (source, target) pair using a clean, independent dataset \cite{Post-TNNLS}. In this work, however, the defender does not possess any images that are guaranteed to be clean. More importantly, our goals here are not only to detect if an attack exists, but also to {\it identify the backdoor training images}, which is not achievable by the method in \cite{Post-TNNLS}, as will be shown in the supplementary material.

\vspace{-0.025in}
\section{Method}\label{sec:defense}

\subsection{Key Ideas}

\subsubsection{Detection}
Suppose $f(\cdot;\Theta^{\ast})$ is a DNN classifier trained on the possibly poisoned training set to be inspected, where the parameters $\Theta^{\ast}$ are obtained by solving (\ref{eq:bd_training}). If there is successful backdoor data poisoning, i.e. ${\mathcal B}\neq\O$ and the expectation (\ref{eq:goal1}) is large enough, for any backdoor class pair $(s, t^{\ast})$ with $s\in{\mathcal S}^{\ast}$, a large fraction of training images from class $s$, since they also follow the distribution $P_{{\mathcal S}^{\ast}}$, will be misclassified to class $t^{\ast}$ when the backdoor pattern ${\bf v}^{\ast}$ is embedded. Thus, for this class pair, high misclassifications can be achieved with a {\it small} perturbation. However, for any non-backdoor class pair $(s, t)$, to achieve a similarly high misclassification rate from $s$ to $t$, the minimum required perturbation norm will be very large\footnote{As is experimentally supported by several works, when there is no backdoor, the more images from one class that need to be perturbed to another class using a common perturbation, the larger the minimum perturbation norm is required \cite{DeepFool_Univ,Post-TNNLS}.} -- likely much larger than the norm of the imperceptible backdoor pattern ${\bf v}^{\ast}$ \cite{Post-TNNLS}.

\subsubsection{Training Set Cleansing}
It is straightforward that sequentially embedding a pattern ${\bf v}$ and then $-{\bf v}$ to an image ${\bf x}$ will induce limited changes to the image, i.e.,
\begin{equation}\label{eq:bd_similarity}
||m(m({\bf x};{\bf v}); -{\bf v})-{\bf x}||_{\infty} \leq ||{\bf v}||_{\infty}
\end{equation}
Moreover, if the clipping function is not activated, the left hand side of (\ref{eq:bd_similarity}) will be zero -- no difference is induced. Hence, if there is an attack, i.e. ${\mathcal B}\neq\O$, for any $({\bf x}, y)\in{\mathcal B}$ where ${\bf x}=m(\tilde{\bf x};{\bf v}^{\ast})$ for some $\tilde{\bf x}\sim P_{{\mathcal S}^{\ast}}$ and $y=t^{\ast}$, with high probability
\begin{equation}\label{eq:label_recover}
f(m({\bf x};-{\bf v}^{\ast});\Theta^{\ast}) = f(\tilde{\bf x};\Theta^{\ast}) \in {\mathcal S}^{\ast}
\end{equation}
i.e., the label of the originally clean image of a backdoor training image can be recovered by {\it removing} the embedded backdoor pattern. However, for clean training images labeled to class $t^{\ast}$, ``embedding'' the pattern $-{\bf v}^{\ast}$, or any arbitrary pattern with similarly small norm, will likely not change the label predicted by $f(\cdot;\Theta^{\ast})$.

\subsection{Reverse Engineering Defense}

Our defense (summarized as Algorithm \ref{alg:main}) consists of three main procedures: pattern estimation, detection inference, and training set cleansing. It is designed for both detection of backdoor attacks and training set cleansing, and in addition, estimation of the true backdoor pattern ${\bf v}^{\ast}$.

\subsubsection{Pattern Estimation}

%Based on the key ideas of our detection, when there is a successful backdoor data poisoning, the existence of a small size perturbation that induces high misclassification from any $s\in{\mathcal S}^{\ast}$ to $t^{\ast}$ is guaranteed by the existence of ${\bf v}^{\ast}$. Moreover, as experimentally verified by \cite{Post-TNNLS,Qiao}, ${\bf v}^{\ast}$ is not a unique perturbation that induces high misclassification for backdoor class pairs. This motivates our search of a small pattern inducing high misclassification (similar to the true backdoor pattern ${\bf v}^{\ast}$), i.e. reverse engineering the backdoor pattern, for the backdoor class pairs.

Based on the key ideas of our detection, when there is a successful backdoor data poisoning, the existence of a small-sized perturbation that induces high misclassification from any $s\in{\mathcal S}^{\ast}$ to $t^{\ast}$ is guaranteed by the existence of ${\bf v}^{\ast}$. This motivates our search of such a small-sized perturbation, i.e. reverse engineering the backdoor pattern
% Zhen -- removed the footenote as you suggested
%\footnote{For future attacks with other imperceptible backdoor embedding mechanisms, our defense may be generalized by estimating a perturbation on internal layer activations \cite{Post-TNNLS}.}
, for the backdoor class pairs.

To serve both the {\it detection} and {\it training set cleansing} purposes, for any class pair $(s,t)\in{\mathcal Y}\times{\mathcal Y}$ and $s\neq t$, under the hypothesis that $(s,t)$ is a backdoor class pair (i.e. $s\in{\mathcal S}^{\ast}$ and $t=t^{\ast}$), we expect the following to hold:\\
1) High misclassification from $s$ to $t$ is induced by a well-chosen small perturbation.\\
2) The class decision of the backdoor training images is no longer $t$ when this {\it same} perturbation is ``removed''.\\
3) It is not possible for any small perturbation to change the class decision of a significant proportion of the clean training images labeled to $t$, when the pattern is ``removed'' from these images.\\
Hence we search for a small pattern that induces at least $\pi\in(0,1)$ misclassification fraction from $s$ to $t$ (on the training images labeled to $s$) and maximizes the class decision changes for the training images labeled to $t$ when ``removed'' from them, i.e.
\begin{equation}\label{eq:opt_raw}
\begin{aligned}
& \underset{\bf v}{\text{maximize}}
& & \frac{1}{|\mathcal{D}_t|}\sum_{({\bf x},y)\in\mathcal{D}_t}{\mathds 1} (f(m({\bf x};-{\bf v});\Theta^{\ast})\neq t) \\
& \text{subject to}
& & \frac{1}{|\mathcal{D}_s|}\sum_{({\bf x},y)\in\mathcal{D}_s}{\mathds 1} (f(m({\bf x};{\bf v});\Theta^{\ast})=t) ~\geq~ \pi,\\
& & & d({\bf v}) \leq \Delta_{st}
\end{aligned}
\end{equation}
Here, $d(\cdot)$ is any reasonable metric representing the perturbation size (e.g. $l_2$ norm), and $\mathcal{D}_s$, $\mathcal{D}_t$ represent the subsets of the training set labeled to class $s$ and class $t$, respectively. $\Delta_{st}$ constrains the perturbation size (with reference to metric $d(\cdot)$) to be small.

Since $\Delta_{st}$ and the feasible set of ${\bf v}$ for solving (\ref{eq:opt_raw}) cannot be easily specified, we use a gradient-based search algorithm that seeks to maximize the following Lagrangian objective:
\begin{equation}\label{eq:opt_intermediate}
\begin{aligned}
\underset{\bf v}{\text{maximize}}
\quad \frac{1}{|\mathcal{D}_t|}\sum_{({\bf x},y)\in\mathcal{D}_t}{\mathds 1} (f(m({\bf x};-{\bf v});\Theta^{\ast})\neq t) \\
+ \lambda \times \frac{1}{|\mathcal{D}_s|}\sum_{({\bf x},y)\in\mathcal{D}_s}{\mathds 1} (f(m({\bf x};{\bf v});\Theta^{\ast})=t)
\end{aligned}
\end{equation}
Since ${\mathds 1}(\cdot)$ in (\ref{eq:opt_intermediate}) is not differentiable, we adopt the following differentiable surrogate objective function
\begin{equation}\label{eq:obj}
\begin{aligned}
J_{st}({\bf v})= \frac{1}{|{\mathcal D}_t|}\sum_{({\bf x},y)\in{\mathcal D}_t} \log(1-p_t(m({\bf x};-{\bf v}),\Theta^{\ast}))\\
+\lambda \times\frac{1}{|{\mathcal D}_s|} \sum_{({\bf x},y)\in{\mathcal D}_s} \log(p_t(m({\bf x};{\bf v}),\Theta^{\ast}))
\end{aligned}
\end{equation}
Here, we take logarithm to make the objective function smoother. We search for ${\bf v}$ along the gradient of (\ref{eq:obj}) with a {\it relatively small} step size $\delta$, starting from ${\bf v}=\boldsymbol{0}$. The searching is stopped immediately once ${\bf v}$ successfully induces at least $\pi$ fraction of misclassifications from $s$ to $t$ on all the training images labeled to $s$. This termination condition limits the size of the perturbation. Thus, the resulting perturbation is expected to satisfy both constraints of (\ref{eq:opt_raw}), i.e. inducing at least $\pi$ fraction of misclassification from $s$ to $t$ and having as small size as possible. The pattern estimation algorithm is detailed in lines 3-7 of Algorithm \ref{alg:main}. It is applied to all class pairs independently and we denote the estimated pattern for any $(s,t)$ pair as $\hat{{\bf v}}_{st}$. The step size $\delta$ and the target misclassification fraction $\pi$ can be easily chosen without any supervision. The Lagrange multiplier is set to $\lambda=1$ by default. The implementation details will be described in the experiment section.

%, as described in Section XXX. The Lagrange multiplier $\lambda$ is set to $\lambda=1$ by default, though its choice is not critical to the performance of our defense as shown in Section XXX.

%The target mislcassification fraction $\pi$ should be large enough, based on the key ideas, for distinguishing backdoor class pairs from non-backdoor class pairs. The step size $\delta$ can be easily scheduled to ensure that the $\pi$-misclassification is achieved smoothly -- if $\delta$ is overly large, $\pi$-misclassification may be achieved in a single iteration, with the estimated pattern 

\subsubsection{Detection Inference}

Our detection inference uses the same hypothesis test as in \cite{Post-TNNLS}. For each class pair $(s,t)$, we obtain a reciprocal statistic $r_{st}=d(\hat{{\bf v}}_{st})^{-1}$. The null hypothesis here is that there is no attack -- all the reciprocal statistics are from the same population. For each putative target class $c\in\{1,\ldots,K\}$, we exclude all the $(K-1)$ reciprocals with $t=c$, and fit a null parameterized (e.g. Gamma) density $g_{R_c}(r)$ using maximum likelihood estimation (MLE) with the remaining reciprocals. Then we evaluate the probability that the largest of the $(K-1)$ reciprocals being excluded is no less than their observed maximum, $r_{c,{\rm max}}$, under the null density, i.e. an order statistic p-value:
\begin{equation}\label{eq:order_pv_1}
{\rm pv}_{c,\rm max} =1-G_{R_c}(r_{c,\rm max})^{K-1}, \quad c=1, \ldots, K
\end{equation}
where $G_{R_c}$ is the cdf corresponding to $g_{R_c}$. Under the null hypothesis, each of the $K$ order statistic p-values follows a uniform distribution on the interval $[0,1]$. But if there is an attack with target class $t^{\ast}$, based on the key ideas of our detection, ${\rm pv}_{t^{\ast},\rm max}$ will be abnormally small with high probability. Hence we evaluate the probability (under the uniform distribution) that the smallest of the $K$ order statistic p-values is no larger than the observed minimum, i.e. an order statistic {\it on} order statistics:
\begin{equation}\label{eq:order_pv_2}
{\rm pv} = 1-(1-{\rm pv}_{\rm min})^K,
\end{equation}
where ${\rm pv}_{\rm min} = \min_c {\rm pv}_{c,\rm max}$. We infer that there is an attack with confidence $\theta$ (e.g. $\theta=0.05$) if ${\rm pv}<\theta$. Also, $\hat{t} = \argmin_c {\rm pv}_{c,\rm max}$ is inferred to be the target class involved; and $\hat{s} = \argmax_s r_{s\hat{t}}$ is inferred as one of the source classes involved in the attack. Moreover, $\hat{\bf v}=\hat{\bf v}_{\hat{s}\hat{t}}$ is our estimation of the backdoor pattern ${\bf v}^{\ast}$.

\subsubsection{Training Set Cleansing}

In fact, the key of our training set cleansing is embedded in the formulation of our pattern estimation -- we estimate a pattern for each putative target class to induce label changes when the pattern is removed from the training images from the class. Hence if an attack is detected and the associated $(\hat{s}, \hat{t})$ are obtained, we identify backdoor training images from ${\mathcal D}_{\hat{t}}$ as those ${\bf x}$ such that
\begin{equation}\label{eq:bd_image_identify}
f(m({\bf x};-\hat{\bf v});\Theta^{\ast}) \neq \hat{t}.
\end{equation}
These images are removed before retraining.

\begin{algorithm}[t]
	\caption{Reverse engineering backdoor defense.}\label{alg:main}
	\begin{algorithmic}[1]
		\State {\bf Inputs}: labeled training set $\{{\mathcal D}_1,\ldots, {\mathcal D}_K\}$, classifier $f(\cdot;\Theta^{\ast})$
		\State {\bf Pattern estimation}:
		\For{all $(s,t)$ class pair}
		\State ${\bf v}\gets \boldsymbol{0}$, $\rho\gets \frac{1}{|\mathcal{D}_s|}\sum_{({\bf x},y)\in\mathcal{D}_s}{\mathds 1}(f({\bf x};\Theta^{\ast})=t)$
		\While{$\rho < \pi$}
		\State ${\bf v} \gets {\bf v} + \delta\cdot\nabla J_{st}({\bf v})$
		\State $\rho\gets \frac{1}{|\mathcal{D}_s|}\sum_{({\bf x},y)\in\mathcal{D}_s}{\mathds 1}(f(m({\bf x};{\bf v});\Theta^{\ast})=t)$
		\EndWhile
		\State $\hat{\bf v}_{st} = {\bf v}$, $r_{st}=d(\hat{{\bf v}}_{st})^{-1}$
		\EndFor
		\State {\bf Detection inference}:
		\For{all class $c$}
		\State get $g_{R_c}(r)$ by MLE using all $r_{st}$ with $t\neq c$
		\State evaluate ${\rm pv}_{c,\rm max}$ by Eq. (\ref{eq:order_pv_1})
		\EndFor
		\State evaluate ${\rm pv}$ by Eq. (\ref{eq:order_pv_2})
		\If{${\rm pv}>\theta$}
		\State there is no attack; defense terminated
		\Else
		\State backdoor attack detected; $\hat{t} = \argmin_c {\rm pv}_{c,\rm max}$ is the target class; $\hat{s} = \argmax_s r_{s\hat{t}}$ is one of the source classes; $\hat{\bf v}=\hat{\bf v}_{\hat{s}\hat{t}}$ is the estimation of ${\bf v}^{\ast}$
		\EndIf
		\State {\bf Training set cleansing}:
		\For{all $({\bf x},y)\in{\mathcal D}_{\hat{t}}$}
		\If{$f(m({\bf x};-\hat{\bf v});\Theta^{\ast}) \neq \hat{t}$}
		\State remove $({\bf x},y)$ from training set
		\EndIf
		\EndFor
		\State proceed to retraining if there are samples being removed
	\end{algorithmic}
\end{algorithm}

\vspace{-0.05in}
\section{Experiments}\label{sec:experiments}

We evaluate the performance of our proposed reverse engineering (RE) defense in comparison with other before/during-training backdoor defenses on CIFAR-10 (with 10 classes) \cite{CIFAR10}. In this baseline experiment, we first craft a large variety of backdoor attacks and show their effectiveness against defenseless DNNs. Then we apply our RE defense and its competitors to these attacks respectively to demonstrate RE's superior capability in detection and training set cleansing. Finally, we retrain the classifier with the problematic training images identified by RE being removed, and show that the attack is no longer effective. Due to page limitations, we leave substantial experiments (including experiments on other datasets, robustness analysis, etc.) and discussions to the supplementary material.

\subsection{Attacks Crafting}

\subsubsection{Backdoor Patterns}

\begin{figure}
	\centering
	\begin{minipage}[b]{.3\linewidth}
		\centering
		\centerline{\includegraphics[width=\linewidth]{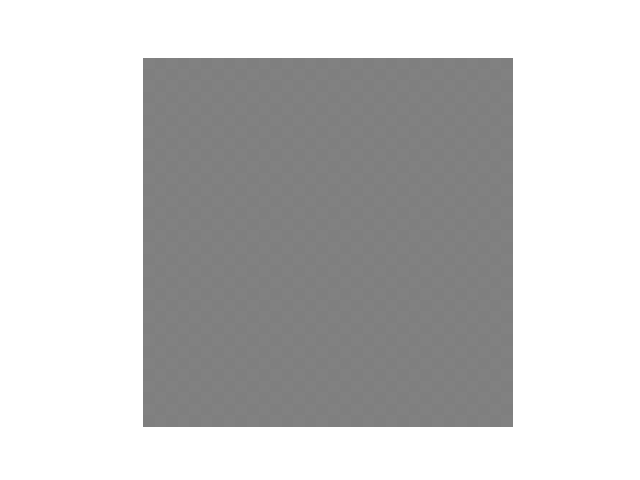}}
		\subcaption{pattern A}
	\end{minipage}
	\begin{minipage}[b]{.3\linewidth}
		\centering
		\centerline{\includegraphics[width=\linewidth]{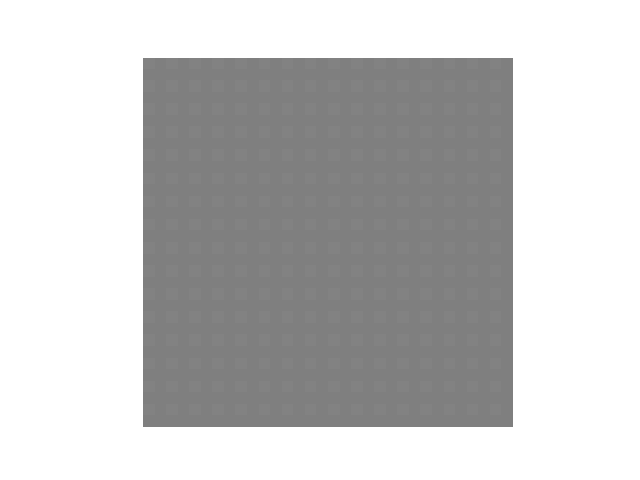}}
		\subcaption{pattern B}
	\end{minipage}
	\begin{minipage}[b]{.3\linewidth}
		\centering
		\centerline{\includegraphics[width=\linewidth]{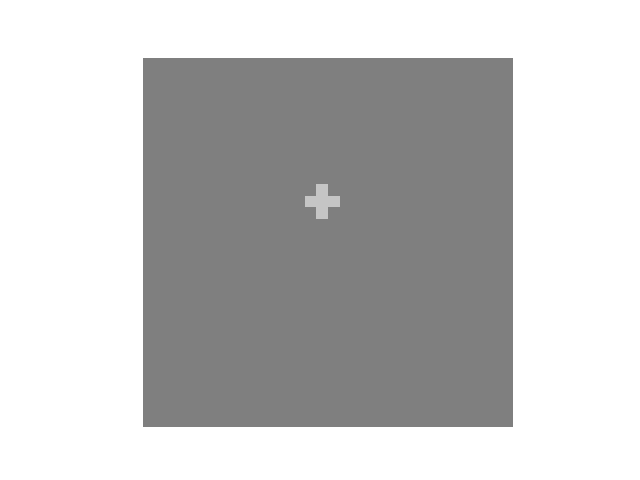}}
		\subcaption{pattern C}
	\end{minipage}
	\begin{minipage}[b]{.3\linewidth}
		\centering
		\centerline{\includegraphics[width=\linewidth]{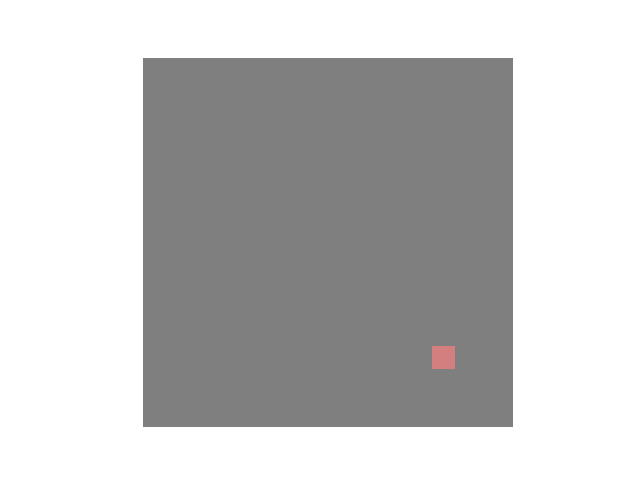}}
		\subcaption{pattern D}
	\end{minipage}
	\begin{minipage}[b]{.3\linewidth}
		\centering
		\centerline{\includegraphics[width=\linewidth]{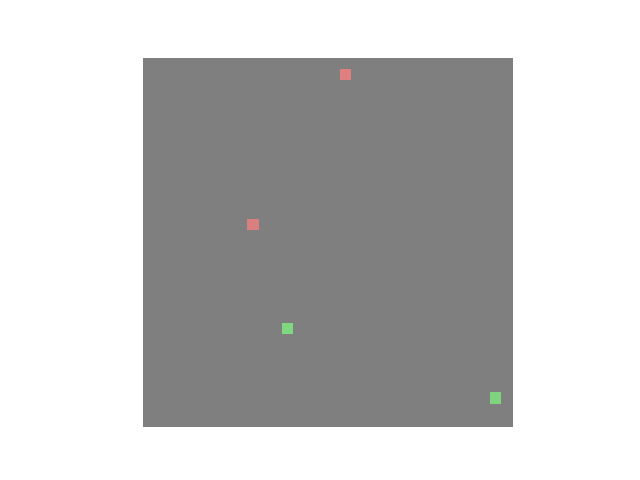}}
		\subcaption{pattern E}
	\end{minipage}
	\begin{minipage}[b]{.3\linewidth}
		\centering
		\centerline{\includegraphics[width=\linewidth]{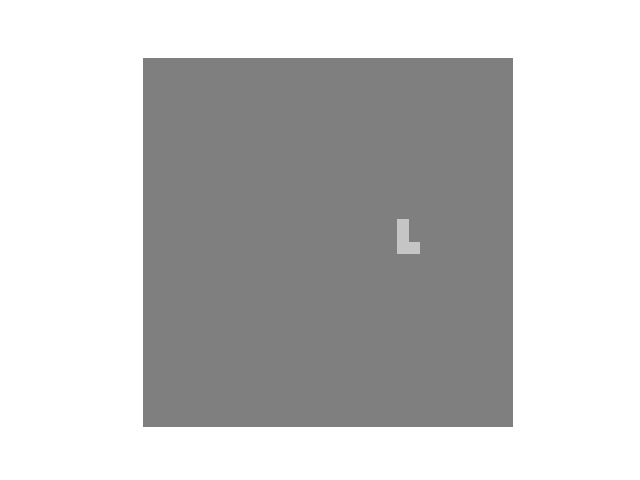}}
		\subcaption{pattern F}
	\end{minipage}
	\caption{Illustration of the backdoor patterns (with a 127/255 offset to manifest possible negative perturbations).}
	\label{fig:bd_pattern}
	\vspace{-0.1in}
\end{figure}

We consider six {\it imperceptible} backdoor patterns (shown in Figure \ref{fig:bd_pattern}) that have appeared in the backdoor literature, including two ``global'' patterns (pattern A, B) and four ``local patterns'' (pattern C-F). Note that pattern A and B are barely visible to humans because the maximum absolute perturbation size is only 2/255\footnote{Although our defense is designed for the general case where the pixel values are continuous, we mimic real-world attackers by imposing an 8-bit finite precision to the pixel values, such that a valid image (with or without a backdoor pattern) should have pixel values in the finite set $\{0, 1/255, \ldots, 254/255, 1\}$.} and 3/255, respectively. 
% Zhen -- removed the following since the original Figure 1, the "trucks" are gone
% Poisoned images embedded with pattern A (a ``chessboard'') and pattern C (a ``cross'') respectively are shown in Figure \ref{fig:bd_example}. 
Detailed generating mechanism of the six backdoor patterns and examples of poisoned images are shown in the supplementary material.

\subsubsection{Attack Configurations}

For each backdoor pattern, we create three backdoor attacks with one source class (1SC), three source classes (3SC), and nine source classes (i.e. all classes except for the target class are the source classes) (9SC), respectively. The source class(es) and the target class for each attack are arbitrarily chosen, with the detailed choices provided in the supplementary material. For each backdoor pattern, 500, 200, and 60 clean training images per source class are used to create backdoor training images for the 1SC, 3SC, and 9SC attacks, respectively.

\subsubsection{Attack Effectiveness on Defenseless DNNs}

We consider two slightly different architectures from the standard ResNet model family for the classification task on CIFAR-10. Both architectures contain 4 groups of residual blocks, with 2 blocks per group (see the ``18-layer'' structure in Table 1 of \cite{ResNet}). However, one architecture uses 64, 128, 256, and 512 filters for each convolutional layer for the 4 groups of residual blocks, respectively (i.e. a wide architecture), and achieves \url{~92}\% accuracy on the test set of CIFAR-10 when there is no attack; while the other architecture uses 16, 32, 64, and 64 filters instead (i.e. a compact, economic architecture), and achieves \url{~91}\% test accuracy on CIFAR-10 without an attack. The details of the DNN architectures and the training configurations are provided in the supplementary material. Here, we emphasize that the DNN architecture and the training configurations are {\it not} selected by the attacker. Experiments on CIFAR-10 with other DNN architectures are also reported in the supplementary material.

In Table \ref{tab:ASR_ACC}, we show that all 18 attacks (1SC, 3SC, and 9SC attacks for 6 backdoor patterns) crafted for evaluating the performance of defenses are successful against defenseless DNNs for both the wide and the compact architectures. The attack success rate (ASR) for each attack is the percentage of the test images from the source class(es) being classified to the target class when the backdoor pattern is embedded. For each backdoor pattern and each DNN architecture, we also train a DNN on the clean (i.e. not backdoor poisoned) training set and obtain a benchmark accuracy (ACC) on the clean test set (with no backdoor patterns). For all attacks, both of the attacker's goals -- a high ASR and a negligible degradation in ACC -- are achieved.

\begin{table}[t]
	\begin{center}
		\caption{Attack success rate (ASR) and the accuracy (ACC) on clean test set (jointly represented by ASR/ACC) for each of the 18 attacks (1SC, 3SC, and 9SC attacks for pattern A-F) against defenseless DNNs for both wide and compact architectures; and the test ACC of the clean benchmark DNNs (ASR is not applicable (represented by n.a.) to clean DNNs).}
		\resizebox{0.48\textwidth}{!}{
			\begin{tabular}{ |c|c|c|c|c|c|c|c| }
				\hline 
				\multicolumn{2}{|c|}{pattern} & A & B & C & D & E & F\\
				\hline
				\multirow{4}{*}{\thead{wide\\DNN}} 
				& \multicolumn{1}{|c|}{clean} & \multicolumn{1}{|c|}{n.a./92.2} & \multicolumn{1}{|c|}{n.a./91.8} & \multicolumn{1}{|c|}{n.a./91.9} & \multicolumn{1}{|c|}{n.a./91.7} & \multicolumn{1}{|c|}{n.a./92.3} & \multicolumn{1}{|c|}{n.a./91.3}\\
				\cline{2-8}
				& \multicolumn{1}{|c|}{1SC} & \multicolumn{1}{|c|}{99.2/92.1} & \multicolumn{1}{|c|}{97.3/92.0} & \multicolumn{1}{|c|}{98.9/92.2} & \multicolumn{1}{|c|}{96.2/92.1} & \multicolumn{1}{|c|}{97.0/91.8} & \multicolumn{1}{|c|}{86.1/91.7}\\
				\cline{2-8}
				& \multicolumn{1}{|c|}{3SC} & \multicolumn{1}{|c|}{99.5/91.6} & \multicolumn{1}{|c|}{98.5/92.0} & \multicolumn{1}{|c|}{99.3/91.8} & \multicolumn{1}{|c|}{99.5/91.8} & \multicolumn{1}{|c|}{99.9/92.1} & \multicolumn{1}{|c|}{94.2/90.8}\\
				\cline{2-8}
				& \multicolumn{1}{|c|}{9SC} & \multicolumn{1}{|c|}{98.8/91.7} & \multicolumn{1}{|c|}{97.1/91.9} & \multicolumn{1}{|c|}{98.4/91.7} & \multicolumn{1}{|c|}{92.6/91.9} & \multicolumn{1}{|c|}{99.4/91.7} & \multicolumn{1}{|c|}{89.4/92.0}\\
				\hline
				\multirow{4}{*}{\thead{com-\\pact\\DNN}} 
				& \multicolumn{1}{|c|}{clean} & \multicolumn{1}{|c|}{n.a./90.4} & \multicolumn{1}{|c|}{n.a./90.7} & \multicolumn{1}{|c|}{n.a./91.3} & \multicolumn{1}{|c|}{n.a./91.2} & \multicolumn{1}{|c|}{n.a./90.4} & \multicolumn{1}{|c|}{n.a./90.8}\\
				\cline{2-8}
				& \multicolumn{1}{|c|}{1SC} & \multicolumn{1}{|c|}{99.1/90.8} & \multicolumn{1}{|c|}{99.5/90.5} & \multicolumn{1}{|c|}{96.4/90.3} & \multicolumn{1}{|c|}{92.4/90.6} & \multicolumn{1}{|c|}{96.0/90.7} & \multicolumn{1}{|c|}{89.4/90.1}\\
				\cline{2-8}
				& \multicolumn{1}{|c|}{3SC} & \multicolumn{1}{|c|}{99.3/90.1} & \multicolumn{1}{|c|}{91.0/90.9} & \multicolumn{1}{|c|}{98.1/90.2} & \multicolumn{1}{|c|}{99.5/90.4} & \multicolumn{1}{|c|}{99.6/90.6} & \multicolumn{1}{|c|}{90.5/89.8}\\
				\cline{2-8}
				& \multicolumn{1}{|c|}{9SC} & \multicolumn{1}{|c|}{99.1/90.3} & \multicolumn{1}{|c|}{97.7/90.5} & \multicolumn{1}{|c|}{97.3/90.8} & \multicolumn{1}{|c|}{86.5/90.6} & \multicolumn{1}{|c|}{98.2/90.0} & \multicolumn{1}{|c|}{87.6/90.2}\\
				\hline
			\end{tabular}\label{tab:ASR_ACC}}
	\end{center}
	\vspace{-0.15in}
\end{table}

\subsection{Defense Performance Evaluation}

\subsubsection{Implementation Details of the Proposed RE Defense}

In principle, the step size $\delta$ for pattern estimation should be relatively small such that $\pi$-misclassification can be achieved with as small a perturbation size as possible. If $\delta$ is chosen too large, $\pi$-misclassification will be achieved in very few iterations with a large overshoot, and the norm of the estimated pattern will likely be overly large. In practice, one can choose $\delta$ by line search for each iteration of pattern estimation. Here, we set $\delta=10^{-4}$ for simplicity. Based on the key ideas of our defense, the target mislcassification fraction $\pi$ for pattern estimation should be relatively large in order to distinguish backdoor class pairs from non-backdoor class pairs. Here, we set $\pi=0.95$, but this choice is not critical to the defense. Similarly, we use the default $\lambda=1$. Robustness analysis of the choices of $\pi$ and $\lambda$ are provided in the supplementary material. Moreover, in each iteration of pattern estimation, we compute the gradient of (\ref{eq:obj}) on two batches of training images (with batch size 256) randomly sampled from ${\mathcal{D}}_t$ and ${\mathcal{D}}_s$, respectively. This will largely reduce computational complexity and help avoid poor local optima. For detection inference, we use Gamma as the null density form; we use $l_2$ norm as the metric for perturbation size; and the confidence-level threshold is set to the classical $\theta=0.05$. Finally, there is no constraint on the DNN architecture or training configurations for the classifier trained on the possibly poisoned training set (as part of our defense). Here, we consider both the wide and the compact DNN architectures, and use the same configuration as for training the defenseless classifiers in the previous section, but without training data augmentation -- this ensures that the true backdoor pattern (but not the resized/shifted version) is well learned.

\begin{table}[t]
	\caption{Detection performance evaluation of (a) RE defense, in comparison with (b) AC and (c) CI, on the 18 poisoned training sets and the clean training sets for both wide and compact DNN architectures. Symbols $\otimes$, $\oslash$, and $\odot$ represent: attack is not detected (or falsely detected for clean training set), attack is detected but the target class is incorrectly inferred, and attack is detected and the target class is correctly inferred (or no attack is detected for clean training set), respectively.}\label{tab:exp_detection}
	\centering
	\begin{subtable}[t]{1\columnwidth}
		\centering
		\resizebox{0.6\textwidth}{!}{
			\begin{tabular}{ |c|c|c|c|c|c|c|c| }
				\hline 
				\multicolumn{2}{|c|}{pattern} & A & B & C & D & E & F\\
				\hline
				\multirow{4}{*}{\thead{wide\\DNN}} 
				& \multicolumn{1}{|c|}{clean} & \multicolumn{1}{|c|}{$\odot$} & \multicolumn{1}{|c|}{$\odot$} & \multicolumn{1}{|c|}{$\odot$} & \multicolumn{1}{|c|}{$\odot$} & \multicolumn{1}{|c|}{$\otimes$} & \multicolumn{1}{|c|}{$\odot$}\\
				\cline{2-8}
				& \multicolumn{1}{|c|}{1SC} & \multicolumn{1}{|c|}{$\odot$} & \multicolumn{1}{|c|}{$\odot$} & \multicolumn{1}{|c|}{$\odot$} & \multicolumn{1}{|c|}{$\odot$} & \multicolumn{1}{|c|}{$\odot$} & \multicolumn{1}{|c|}{$\odot$}\\
				\cline{2-8}
				& \multicolumn{1}{|c|}{3SC} & \multicolumn{1}{|c|}{$\odot$} & \multicolumn{1}{|c|}{$\odot$} & \multicolumn{1}{|c|}{$\odot$} & \multicolumn{1}{|c|}{$\odot$} & \multicolumn{1}{|c|}{$\odot$} & \multicolumn{1}{|c|}{$\odot$}\\
				\cline{2-8}
				& \multicolumn{1}{|c|}{9SC} & \multicolumn{1}{|c|}{$\odot$} & \multicolumn{1}{|c|}{$\odot$} & \multicolumn{1}{|c|}{$\odot$} & \multicolumn{1}{|c|}{$\odot$} & \multicolumn{1}{|c|}{$\odot$} & \multicolumn{1}{|c|}{$\odot$}\\
				\hline
				\multirow{4}{*}{\thead{com-\\pact\\DNN}} 
				& \multicolumn{1}{|c|}{clean} & \multicolumn{1}{|c|}{$\odot$} & \multicolumn{1}{|c|}{$\odot$} & \multicolumn{1}{|c|}{$\odot$} & \multicolumn{1}{|c|}{$\odot$} & \multicolumn{1}{|c|}{$\odot$} & \multicolumn{1}{|c|}{$\odot$}\\
				\cline{2-8}
				& \multicolumn{1}{|c|}{1SC} & \multicolumn{1}{|c|}{$\odot$} & \multicolumn{1}{|c|}{$\odot$} & \multicolumn{1}{|c|}{$\odot$} & \multicolumn{1}{|c|}{$\odot$} & \multicolumn{1}{|c|}{$\odot$} & \multicolumn{1}{|c|}{$\odot$}\\
				\cline{2-8}
				& \multicolumn{1}{|c|}{3SC} & \multicolumn{1}{|c|}{$\odot$} & \multicolumn{1}{|c|}{$\odot$} & \multicolumn{1}{|c|}{$\odot$} & \multicolumn{1}{|c|}{$\odot$} & \multicolumn{1}{|c|}{$\odot$} & \multicolumn{1}{|c|}{$\odot$}\\
				\cline{2-8}
				& \multicolumn{1}{|c|}{9SC} & \multicolumn{1}{|c|}{$\odot$} & \multicolumn{1}{|c|}{$\odot$} & \multicolumn{1}{|c|}{$\odot$} & \multicolumn{1}{|c|}{$\odot$} & \multicolumn{1}{|c|}{$\odot$} & \multicolumn{1}{|c|}{$\odot$}\\
				\hline
		\end{tabular}}
		\subcaption{RE detection}\label{subtab:exp_detection_RE}
	\end{subtable}
	\begin{subtable}[t]{0.6\columnwidth}
		\centering
		\resizebox{\textwidth}{!}{
			\begin{tabular}{ |c|c|c|c|c|c|c|c| }
				\hline 
				\multicolumn{2}{|c|}{pattern} & A & B & C & D & E & F\\
				\hline
				\multirow{4}{*}{\thead{wide\\DNN}} 
				& \multicolumn{1}{|c|}{clean} & \multicolumn{1}{|c|}{$\odot$} & \multicolumn{1}{|c|}{$\odot$} & \multicolumn{1}{|c|}{$\odot$} & \multicolumn{1}{|c|}{$\odot$} & \multicolumn{1}{|c|}{$\odot$} & \multicolumn{1}{|c|}{$\odot$}\\
				\cline{2-8}
				& \multicolumn{1}{|c|}{1SC} & \multicolumn{1}{|c|}{$\otimes$} & \multicolumn{1}{|c|}{$\odot$} & \multicolumn{1}{|c|}{$\otimes$} & \multicolumn{1}{|c|}{$\otimes$} & \multicolumn{1}{|c|}{$\otimes$} & \multicolumn{1}{|c|}{$\otimes$}\\
				\cline{2-8}
				& \multicolumn{1}{|c|}{3SC} & \multicolumn{1}{|c|}{$\otimes$} & \multicolumn{1}{|c|}{$\otimes$} & \multicolumn{1}{|c|}{$\otimes$} & \multicolumn{1}{|c|}{$\odot$} & \multicolumn{1}{|c|}{$\odot$} & \multicolumn{1}{|c|}{$\otimes$}\\
				\cline{2-8}
				& \multicolumn{1}{|c|}{9SC} & \multicolumn{1}{|c|}{$\otimes$} & \multicolumn{1}{|c|}{$\odot$} & \multicolumn{1}{|c|}{$\otimes$} & \multicolumn{1}{|c|}{$\odot$} & \multicolumn{1}{|c|}{$\odot$} & \multicolumn{1}{|c|}{$\odot$}\\
				\hline
				\multirow{4}{*}{\thead{com-\\pact\\DNN}} 
				& \multicolumn{1}{|c|}{clean} & \multicolumn{1}{|c|}{$\odot$} & \multicolumn{1}{|c|}{$\odot$} & \multicolumn{1}{|c|}{$\odot$} & \multicolumn{1}{|c|}{$\odot$} & \multicolumn{1}{|c|}{$\odot$} & \multicolumn{1}{|c|}{$\odot$}\\
				\cline{2-8}
				& \multicolumn{1}{|c|}{1SC} & \multicolumn{1}{|c|}{$\odot$} & \multicolumn{1}{|c|}{$\oslash$} & \multicolumn{1}{|c|}{$\odot$} & \multicolumn{1}{|c|}{$\odot$} & \multicolumn{1}{|c|}{$\odot$} & \multicolumn{1}{|c|}{$\odot$}\\
				\cline{2-8}
				& \multicolumn{1}{|c|}{3SC} & \multicolumn{1}{|c|}{$\odot$} & \multicolumn{1}{|c|}{$\odot$} & \multicolumn{1}{|c|}{$\odot$} & \multicolumn{1}{|c|}{$\oslash$} & \multicolumn{1}{|c|}{$\odot$} & \multicolumn{1}{|c|}{$\odot$}\\
				\cline{2-8}
				& \multicolumn{1}{|c|}{9SC} & \multicolumn{1}{|c|}{$\odot$} & \multicolumn{1}{|c|}{$\odot$} & \multicolumn{1}{|c|}{$\odot$} & \multicolumn{1}{|c|}{$\odot$} & \multicolumn{1}{|c|}{$\odot$} & \multicolumn{1}{|c|}{$\odot$}\\
				\hline
		\end{tabular}}
		\subcaption{AC detection}\label{subtab:exp_detection_AC}
	\end{subtable}
	\begin{subtable}[t]{0.6\columnwidth}
		\centering
		\resizebox{\textwidth}{!}{
			\begin{tabular}{ |c|c|c|c|c|c|c|c| }
				\hline 
				\multicolumn{2}{|c|}{pattern} & A & B & C & D & E & F\\
				\hline
				\multirow{4}{*}{\thead{wide\\DNN}} 
				& \multicolumn{1}{|c|}{clean} & \multicolumn{1}{|c|}{$\odot$} & \multicolumn{1}{|c|}{$\odot$} & \multicolumn{1}{|c|}{$\odot$} & \multicolumn{1}{|c|}{$\odot$} & \multicolumn{1}{|c|}{$\odot$} & \multicolumn{1}{|c|}{$\odot$}\\
				\cline{2-8}
				& \multicolumn{1}{|c|}{1SC} & \multicolumn{1}{|c|}{$\otimes$} & \multicolumn{1}{|c|}{$\otimes$} & \multicolumn{1}{|c|}{$\otimes$} & \multicolumn{1}{|c|}{$\otimes$} & \multicolumn{1}{|c|}{$\otimes$} & \multicolumn{1}{|c|}{$\otimes$}\\
				\cline{2-8}
				& \multicolumn{1}{|c|}{3SC} & \multicolumn{1}{|c|}{$\otimes$} & \multicolumn{1}{|c|}{$\otimes$} & \multicolumn{1}{|c|}{$\otimes$} & \multicolumn{1}{|c|}{$\otimes$} & \multicolumn{1}{|c|}{$\otimes$} & \multicolumn{1}{|c|}{$\otimes$}\\
				\cline{2-8}
				& \multicolumn{1}{|c|}{9SC} & \multicolumn{1}{|c|}{$\otimes$} & \multicolumn{1}{|c|}{$\otimes$} & \multicolumn{1}{|c|}{$\otimes$} & \multicolumn{1}{|c|}{$\otimes$} & \multicolumn{1}{|c|}{$\otimes$} & \multicolumn{1}{|c|}{$\otimes$}\\
				\hline
				\multirow{4}{*}{\thead{com-\\pact\\DNN}} 
				& \multicolumn{1}{|c|}{clean} & \multicolumn{1}{|c|}{$\odot$} & \multicolumn{1}{|c|}{$\odot$} & \multicolumn{1}{|c|}{$\odot$} & \multicolumn{1}{|c|}{$\odot$} & \multicolumn{1}{|c|}{$\odot$} & \multicolumn{1}{|c|}{$\odot$}\\
				\cline{2-8}
				& \multicolumn{1}{|c|}{1SC} & \multicolumn{1}{|c|}{$\odot$} & \multicolumn{1}{|c|}{$\odot$} & \multicolumn{1}{|c|}{$\odot$} & \multicolumn{1}{|c|}{$\otimes$} & \multicolumn{1}{|c|}{$\odot$} & \multicolumn{1}{|c|}{$\odot$}\\
				\cline{2-8}
				& \multicolumn{1}{|c|}{3SC} & \multicolumn{1}{|c|}{$\odot$} & \multicolumn{1}{|c|}{$\odot$} & \multicolumn{1}{|c|}{$\odot$} & \multicolumn{1}{|c|}{$\odot$} & \multicolumn{1}{|c|}{$\odot$} & \multicolumn{1}{|c|}{$\odot$}\\
				\cline{2-8}
				& \multicolumn{1}{|c|}{9SC} & \multicolumn{1}{|c|}{$\odot$} & \multicolumn{1}{|c|}{$\odot$} & \multicolumn{1}{|c|}{$\odot$} & \multicolumn{1}{|c|}{$\odot$} & \multicolumn{1}{|c|}{$\odot$} & \multicolumn{1}{|c|}{$\odot$}\\
				\hline
		\end{tabular}}
		\subcaption{CI detection}\label{subtab:exp_detection_CI}
	\end{subtable}
\end{table}

\subsubsection{Detection Performance Evaluation}

We evaluate the detection performance of our RE defense in comparison with AC and CI (since no detection approach is proposed by SS) on the 18 attacks for the two DNN architectures. We set the detection thresholds for AC and CI as 0.52 and 0.62, respectively, for the wide DNN architecture; 0.29 and 0.65, respectively, for the compact DNN architecture. These thresholds are chosen to {\it maximize} the performance of AC and CI in detecting whether the training set is poisoned, while maintaining zero false detections when there is no poisoning. In contrast, our RE defense uses a statistical confidence threshold ($\theta=0.05$); it does not require this supervised hyperparameter tuning. As shown in Table \ref{tab:exp_detection}, RE outperforms AC and CI for both DNN architectures by successfully detecting all attacks, with correct inference of the target class\footnote{If the target class is incorrectly inferred, there is no way to remove any backdoor training images. Moreover, clean training images will be falsely removed.} for each attack, with only one false detection. Note that CI fails to detect any attacks for the wide DNN architecture because only one component is obtained by Gaussian mixture modeling with BIC, for all classes, including the true target class.

\subsubsection{Training Set Cleansing Performance Evaluation}

We compare the training set cleansing performance of RE with SS, AC, and CI on the 18 attacks for the two DNN architectures. Although AC did not detect all the attacks like our RE defense, and SS does not even propose any detection approach, we assume for AC and SS that the presence of the attack and the true target class are known to the defender. We further assist SS by setting the number of training images to be removed from the true target class to be 1000 (nearly doubling the number of the backdoor training images embedded in the training set). Note that this number cannot be easily determined by a practical defender without knowing the number of backdoor training images.

\begin{table}[t]
	\caption{Training set cleansing true positive rate (TPR) and false positive rate (FPR) of SS, AC, CI, and our RE (represented in TPR/FPR form), for the 18 attacks, for (a) the wide DNN architecture, and (b) the compact DNN architecture. TPRs $\geq90\%$ and FPRs $\leq10\%$ are in bold.}\label{tab:exp_cleansing}
	\begin{subtable}[t]{1\columnwidth}
		\centering
		\resizebox{\textwidth}{!}{
			\begin{tabular}{ |c|c|c|c|c|c|c|c| }
				\hline 
				\multicolumn{2}{|c|}{pattern} & A & B & C & D & E & F\\
				\hline
				\multirow{3}{*}{1SC} 
				& \multicolumn{1}{|c|}{SS} & \multicolumn{1}{|c|}{{\bf 98.2}/10.2} & \multicolumn{1}{|c|}{{\bf 100}/{\bf 10.0}} & \multicolumn{1}{|c|}{49.8/15.0} & \multicolumn{1}{|c|}{{\bf 98.2}/10.2} & \multicolumn{1}{|c|}{60.2/14.0} & \multicolumn{1}{|c|}{79.4/12.1}\\
				\cline{2-8}
				& \multicolumn{1}{|c|}{AC} & \multicolumn{1}{|c|}{88.4/{\bf 0}} & \multicolumn{1}{|c|}{{\bf 98.6}/{\bf 0}} & \multicolumn{1}{|c|}{83.2/29.1} & \multicolumn{1}{|c|}{{\bf 93.2}/{\bf 0}} & \multicolumn{1}{|c|}{78.6/24.4} & \multicolumn{1}{|c|}{89.4/13.8}\\
				\cline{2-8}
				& \multicolumn{1}{|c|}{RE} & \multicolumn{1}{|c|}{{\bf 94.8}/{\bf 8.4}} & \multicolumn{1}{|c|}{{\bf 97.2}/{\bf 0.3}} & \multicolumn{1}{|c|}{{\bf 95.6}/{\bf 8.6}} & \multicolumn{1}{|c|}{{\bf 92.8}/{\bf 2.8}} & \multicolumn{1}{|c|}{83.0/{\bf 0.2}} & \multicolumn{1}{|c|}{{\bf 98.6}/10.8}\\
				\hline
				\multirow{3}{*}{3SC} 
				& \multicolumn{1}{|c|}{SS} & \multicolumn{1}{|c|}{{\bf 99.7}/{\bf 8.4}} & \multicolumn{1}{|c|}{{\bf 100}/{\bf 8.4}} & \multicolumn{1}{|c|}{68.2/12.2} & \multicolumn{1}{|c|}{{\bf 99.7}/{\bf 8.4}} & \multicolumn{1}{|c|}{{\bf 92.5}/{\bf 9.3}} & \multicolumn{1}{|c|}{85.7/10.1}\\
				\cline{2-8}
				& \multicolumn{1}{|c|}{AC} & \multicolumn{1}{|c|}{{\bf 98.3}/{\bf 0}} & \multicolumn{1}{|c|}{{\bf 98.3}/{\bf 0}} & \multicolumn{1}{|c|}{87.2/27.2} & \multicolumn{1}{|c|}{{\bf 96.3}/{\bf 0}} & \multicolumn{1}{|c|}{87.7/{\bf 0}} & \multicolumn{1}{|c|}{{\bf 87.8}/{\bf 0}}\\
				\cline{2-8}
				& \multicolumn{1}{|c|}{RE} & \multicolumn{1}{|c|}{{\bf 98.0}/12.9} & \multicolumn{1}{|c|}{{\bf 98.7}/{\bf 1.6}} & \multicolumn{1}{|c|}{{\bf 99.2}/{\bf 6.2}} & \multicolumn{1}{|c|}{{\bf 98.2}/{\bf 0}} & \multicolumn{1}{|c|}{{\bf 90.3}/{\bf 0}} & \multicolumn{1}{|c|}{{\bf 98.5}/{\bf 2.8}}\\
				\hline	
				\multirow{3}{*}{9SC} 
				& \multicolumn{1}{|c|}{SS} & \multicolumn{1}{|c|}{{\bf 98.7}/{\bf 9.5}} & \multicolumn{1}{|c|}{{\bf 100}/{\bf 9.3}} & \multicolumn{1}{|c|}{{\bf 90.7}/10.3} & \multicolumn{1}{|c|}{{\bf 96.1}/{\bf 9.8}} & \multicolumn{1}{|c|}{{\bf 97.8}/{\bf 9.6}} & \multicolumn{1}{|c|}{{\bf 92.4}/10.2}\\
				\cline{2-8}
				& \multicolumn{1}{|c|}{AC} & \multicolumn{1}{|c|}{{\bf 97.6}/{\bf 0}} & \multicolumn{1}{|c|}{{\bf 98.7}/{\bf 0}} & \multicolumn{1}{|c|}{{\bf 92.6}/{\bf 0.1}} & \multicolumn{1}{|c|}{89.1/{\bf 0}} & \multicolumn{1}{|c|}{{\bf 92.4}/{\bf 0.4}} & \multicolumn{1}{|c|}{89.3/{\bf 0.1}}\\
				\cline{2-8}
				& \multicolumn{1}{|c|}{RE} & \multicolumn{1}{|c|}{{\bf 96.1}/{\bf 4.2}} & \multicolumn{1}{|c|}{{\bf 98.7}/{\bf 7.9}} & \multicolumn{1}{|c|}{{\bf 99.3}/{\bf 0.4}} & \multicolumn{1}{|c|}{{\bf 94.1}/{\bf 0}} & \multicolumn{1}{|c|}{88.7/{\bf 0}} & \multicolumn{1}{|c|}{{\bf 99.1}/{\bf 5.1}}\\
				\hline
		\end{tabular}}
		\subcaption{wide DNN architecture}\label{subtab:exp_cleansing_wide}
	\end{subtable}
	\begin{subtable}[t]{1\columnwidth}
		\centering
		\resizebox{\textwidth}{!}{
			\begin{tabular}{ |c|c|c|c|c|c|c|c| }
				\hline 
				\multicolumn{2}{|c|}{pattern} & A & B & C & D & E & F\\
				\hline
				\multirow{4}{*}{1SC} 
				& \multicolumn{1}{|c|}{SS} & \multicolumn{1}{|c|}{29.6/17.0} & \multicolumn{1}{|c|}{{\bf 98.2}/10.2} & \multicolumn{1}{|c|}{45.8/15.4} & \multicolumn{1}{|c|}{75.8/12.4} & \multicolumn{1}{|c|}{25.8/17.4} & \multicolumn{1}{|c|}{54.2/14.6}\\
				\cline{2-8}
				& \multicolumn{1}{|c|}{AC} & \multicolumn{1}{|c|}{30.6/45.0} & \multicolumn{1}{|c|}{{\bf 96.0}/{\bf 0}} & \multicolumn{1}{|c|}{81.8/39.4} & \multicolumn{1}{|c|}{{\bf 92.2}/20.2} & \multicolumn{1}{|c|}{59.6/40.0} & \multicolumn{1}{|c|}{89.0/35.4}\\
				\cline{2-8}
				& \multicolumn{1}{|c|}{CI} & \multicolumn{1}{|c|}{55.8/{\bf 7.5}} & \multicolumn{1}{|c|}{{\bf 99.8}/{\bf 0}} & \multicolumn{1}{|c|}{{\bf 93.8}/13.1} & \multicolumn{1}{|c|}{n.a./n.a.} & \multicolumn{1}{|c|}{{\bf 96.2}/55.6} & \multicolumn{1}{|c|}{{\bf 100}/8.4}\\
				\cline{2-8}
				& \multicolumn{1}{|c|}{RE} & \multicolumn{1}{|c|}{{\bf 93.6}/20.9} & \multicolumn{1}{|c|}{{\bf 100}/{\bf 9.5}} & \multicolumn{1}{|c|}{{\bf 91.0}/10.6} & \multicolumn{1}{|c|}{{\bf 98.8}/12.2} & \multicolumn{1}{|c|}{87.8/{\bf 5.1}} & \multicolumn{1}{|c|}{{\bf 98.4}/16.7}\\
				\cline{2-8}
				\hline
				\multirow{4}{*}{3SC} 
				& \multicolumn{1}{|c|}{SS} & \multicolumn{1}{|c|}{39.2/15.7} & \multicolumn{1}{|c|}{30.7/16.7} & \multicolumn{1}{|c|}{16.5/18.4} & \multicolumn{1}{|c|}{86.7/{\bf 9.9}} & \multicolumn{1}{|c|}{55.8/13.7} & \multicolumn{1}{|c|}{58.7/13.3}\\
				\cline{2-8}
				& \multicolumn{1}{|c|}{AC} & \multicolumn{1}{|c|}{61.2/36.1} & \multicolumn{1}{|c|}{72.8/35.9} & \multicolumn{1}{|c|}{61.8/43.4} & \multicolumn{1}{|c|}{{\bf 90.0}/{\bf 0}} & \multicolumn{1}{|c|}{78.5/28.9} & \multicolumn{1}{|c|}{87.3/26.9}\\
				\cline{2-8}
				& \multicolumn{1}{|c|}{CI} & \multicolumn{1}{|c|}{89.7/{\bf 1.9}} & \multicolumn{1}{|c|}{{\bf 97.0}/{\bf 0.1}} & \multicolumn{1}{|c|}{{\bf 98.7}/{\bf 0}} & \multicolumn{1}{|c|}{{\bf 99.0}/{\bf 0}} & \multicolumn{1}{|c|}{{\bf 97.5}/54.6} & \multicolumn{1}{|c|}{{\bf 97.7}/{\bf 2.2}}\\
				\cline{2-8}
				& \multicolumn{1}{|c|}{RE} & \multicolumn{1}{|c|}{{\bf 94.8}/11.3} & \multicolumn{1}{|c|}{{\bf 99.2}/12.6} & \multicolumn{1}{|c|}{{\bf 99.0}/{\bf 4.5}} & \multicolumn{1}{|c|}{{\bf 95.8}/{\bf 0.1}} & \multicolumn{1}{|c|}{{\bf 90.2}/{\bf 2.8}} & \multicolumn{1}{|c|}{{\bf 99.0}/{\bf 2.4}}\\
				\hline	
				\multirow{4}{*}{9SC} 
				& \multicolumn{1}{|c|}{SS} & \multicolumn{1}{|c|}{16.9/18.3} & \multicolumn{1}{|c|}{18.0/18.2} & \multicolumn{1}{|c|}{21.3/17.8} & \multicolumn{1}{|c|}{45.6/15.2} & \multicolumn{1}{|c|}{57.4/13.9} & \multicolumn{1}{|c|}{73.9/12.2}\\
				\cline{2-8}
				& \multicolumn{1}{|c|}{AC} & \multicolumn{1}{|c|}{23.0/41.7} & \multicolumn{1}{|c|}{57.0/40.5} & \multicolumn{1}{|c|}{60.0/42.7} & \multicolumn{1}{|c|}{75.9/32.4} & \multicolumn{1}{|c|}{83.5/30.8} & \multicolumn{1}{|c|}{{\bf 91.5}/23.5}\\
				\cline{2-8}
				& \multicolumn{1}{|c|}{CI} & \multicolumn{1}{|c|}{{\bf 96.5}/{\bf 0}} & \multicolumn{1}{|c|}{{\bf 95.7}/{\bf 0}} & \multicolumn{1}{|c|}{{\bf 96.7}/{\bf 0}} & \multicolumn{1}{|c|}{{\bf 95.9}/{\bf 1.1}} & \multicolumn{1}{|c|}{{\bf 98.3}/43.9} & \multicolumn{1}{|c|}{{\bf 99.1}/{\bf 0.2}}\\
				\cline{2-8}
				& \multicolumn{1}{|c|}{RE} & \multicolumn{1}{|c|}{{\bf 98.3}/{\bf 9.5}} & \multicolumn{1}{|c|}{{\bf 94.6}/10.2} & \multicolumn{1}{|c|}{{\bf 97.6}/{\bf 1.0}} & \multicolumn{1}{|c|}{{\bf 92.2}/{\bf 1.0}} & \multicolumn{1}{|c|}{{\bf 91.5}/{\bf 2.0}} & \multicolumn{1}{|c|}{{\bf 98.5}/{\bf 4.6}}\\
				\hline
		\end{tabular}}
		\subcaption{compact DNN architecture}\label{subtab:exp_cleansing_compact}
	\end{subtable}
	\vspace{-0.2in}
\end{table}

In Table \ref{tab:exp_cleansing}, we show the true positive rate (TPR) and false positive rate (FPR) of training set cleansing for SS, AC, CI, and our RE against the 18 attacks for the two DNN architectures. TPR is defined as the percentage of backdoor training images being correctly identified and removed; FPR is defined as the percentage of the clean training images labeled to the true target class being falsely removed. For CI, training set cleansing is performed simultaneously with attack detection -- all training images from a GMM component are removed if its ``cluster impurity'' measure exceeds the detection threshold \cite{CI}. Since for all the 18 attacks, CI estimates only one component for the true target class if the wide DNN architecture is used by the defender, even if the true target class is known to the defender, CI cannot identify or remove any backdoor training images. Hence, we do not include the failing results of CI in Table \ref{subtab:exp_cleansing_wide} due to space limitation. Although we recognize that CI works well (for both detection and training set cleansing) for most of the attacks if the defender chooses to use the compact DNN architecture (with a lower dimension for the internal layer features), wide architectures may be preferred in some applications. In contrast, SS and AC achieve high TPRs and low FPRs for a significant number of attacks for the wide DNN architecture, but fail for most of the attacks if the compact DNN architecture is used. More insights/discussions on the results regarding SS, AC, and CI are provided in the supplementary material. In comparison, our RE is uniformly effective, regardless of which DNN architecture is used for defense -- it outperforms its competitors in both {\bf detection} and {\bf training set cleansing}.

\subsubsection{Backdoor Pattern Estimation}

\begin{figure}
	\centering
	\begin{minipage}[b]{.3\linewidth}
		\centering
		\centerline{\includegraphics[width=\linewidth]{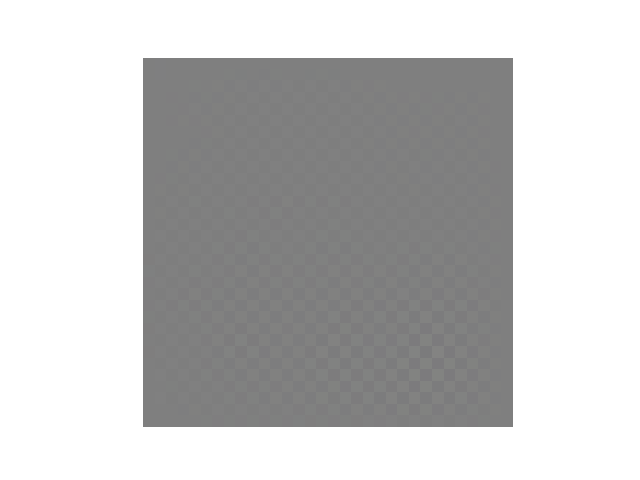}}
		\subcaption{pattern A}
	\end{minipage}
	\begin{minipage}[b]{.3\linewidth}
		\centering
		\centerline{\includegraphics[width=\linewidth]{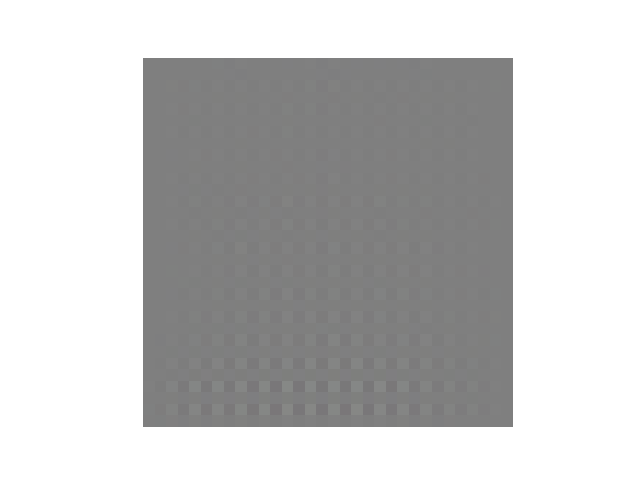}}
		\subcaption{pattern B}
	\end{minipage}
	\begin{minipage}[b]{.3\linewidth}
		\centering
		\centerline{\includegraphics[width=\linewidth]{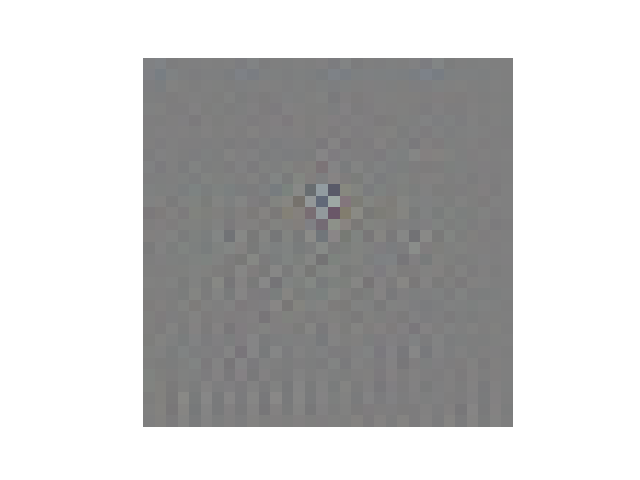}}
		\subcaption{pattern C}
	\end{minipage}
	\begin{minipage}[b]{.3\linewidth}
		\centering
		\centerline{\includegraphics[width=\linewidth]{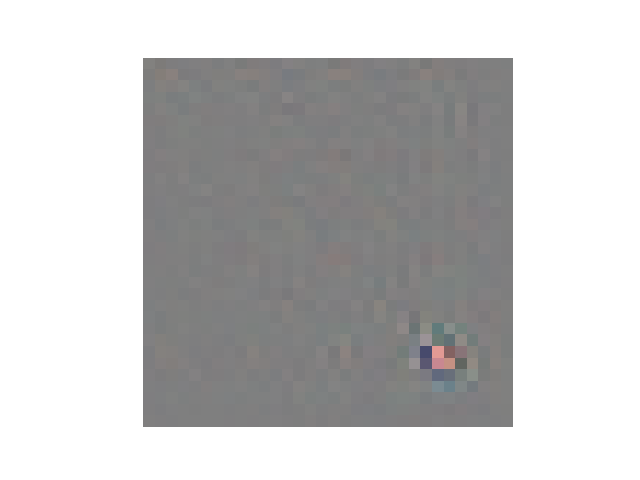}}
		\subcaption{pattern D}
	\end{minipage}
	\begin{minipage}[b]{.3\linewidth}
		\centering
		\centerline{\includegraphics[width=\linewidth]{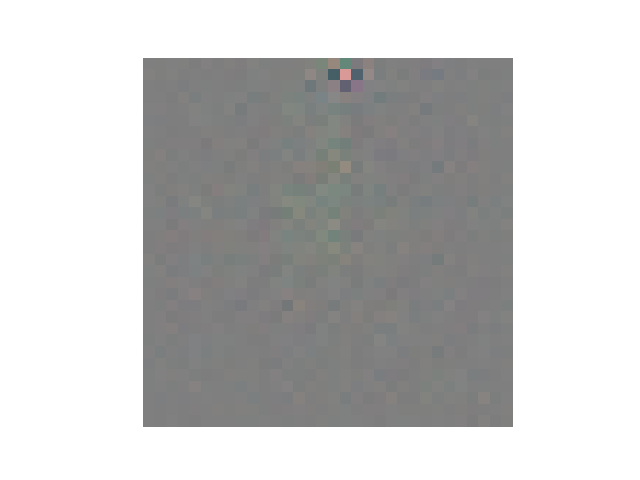}}
		\subcaption{pattern E}
	\end{minipage}
	\begin{minipage}[b]{.3\linewidth}
		\centering
		\centerline{\includegraphics[width=\linewidth]{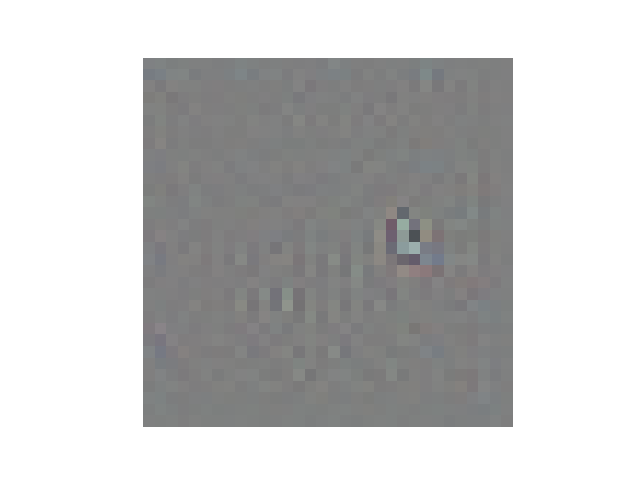}}
		\subcaption{pattern F}
	\end{minipage}
	\caption{Pattern estimated by RE for the 1SC attacks for the wide DNN architecture.}
	\label{fig:bd_pattern_est}
	\vspace{-0.00in}
\end{figure}

In Figure \ref{fig:bd_pattern_est}, we show the pattern estimated by RE for the six 1SC attacks for the wide DNN architecture. We accurately recover most of the true backdoor patterns shown in Figure \ref{fig:bd_pattern} (with, e.g., $2\times10^{-5}$ squared error per pixel for pattern A). For pattern E, we recover the topmost pixel, which is the actual backdoor pattern learned by the classifier. Example backdoor training images with the estimated backdoor pattern being ``removed'' are shown in the supplementary material -- these images are easily classified to the source classes by the classifier trained on the poisoned training set.

\subsubsection{Retraining}

In Table \ref{tab:ASR_ACC_retrain_RE}, we show the ASR and the clean test ACC of the classifiers retrained on the sanitized training set, for RE following removal of the suspicious training images. Retraining results regarding SS, AC, and CI are reported in the supplementary material. With the protection of our RE, {\it the maximum ASR of the 18 attacks is merely 4.9\%}. Also, we observe no degradation in the clean test ACC of the retrained classifiers due to the low FPR of RE in training set cleansing.

\begin{table}[t]
	\begin{center}
		\caption{Attack success rate (ASR) and clean test accuracy (ACC), jointly represented by ASR/ACC, of the retrained classifiers for the 18 attacks when RE is applied, for both wide and compact DNN architectures.}
		\vspace{-0.1in}
		\resizebox{0.48\textwidth}{!}{
			\begin{tabular}{ |c|c|c|c|c|c|c|c| }
				\hline 
				\multicolumn{2}{|c|}{pattern} & A & B & C & D & E & F\\
				\hline
				\multirow{3}{*}{\thead{wide\\DNN}}
				& \multicolumn{1}{|c|}{1SC} & \multicolumn{1}{|c|}{4.9/91.9} & \multicolumn{1}{|c|}{2.7/92.1} & \multicolumn{1}{|c|}{2.2/91.8} & \multicolumn{1}{|c|}{1.1/92.7} & \multicolumn{1}{|c|}{3.5/92.6} & \multicolumn{1}{|c|}{0.8/91.6}\\
				\cline{2-8}
				& \multicolumn{1}{|c|}{3SC} & \multicolumn{1}{|c|}{0.8/92.3} & \multicolumn{1}{|c|}{0.1/92.9} & \multicolumn{1}{|c|}{0/92.0} & \multicolumn{1}{|c|}{0.7/92.3} & \multicolumn{1}{|c|}{1.6/92.7} & \multicolumn{1}{|c|}{0.3/91.6}\\
				\cline{2-8}
				& \multicolumn{1}{|c|}{9SC} & \multicolumn{1}{|c|}{0.6/92.5} & \multicolumn{1}{|c|}{0.2/92.9} & \multicolumn{1}{|c|}{0.3/92.9} & \multicolumn{1}{|c|}{1.4/91.7} & \multicolumn{1}{|c|}{0.8/92.1} & \multicolumn{1}{|c|}{0.8/91.9}\\
				\hline
				\multirow{3}{*}{\thead{com-\\pact\\DNN}}
				& \multicolumn{1}{|c|}{1SC} & \multicolumn{1}{|c|}{1.9/90.2} & \multicolumn{1}{|c|}{0/90.2} & \multicolumn{1}{|c|}{2.6/90.4} & \multicolumn{1}{|c|}{0.3/90.0} & \multicolumn{1}{|c|}{3.1/89.9} & \multicolumn{1}{|c|}{0.6/90.5}\\
				\cline{2-8}
				& \multicolumn{1}{|c|}{3SC} & \multicolumn{1}{|c|}{1.1/90.1} & \multicolumn{1}{|c|}{0.4/90.3} & \multicolumn{1}{|c|}{0.1/90.5} & \multicolumn{1}{|c|}{0.9/90.5} & \multicolumn{1}{|c|}{1.8/90.3} & \multicolumn{1}{|c|}{0.5/89.7}\\
				\cline{2-8}
				& \multicolumn{1}{|c|}{9SC} & \multicolumn{1}{|c|}{0.9/89.9} & \multicolumn{1}{|c|}{0.9/90.0} & \multicolumn{1}{|c|}{0.7/90.2} & \multicolumn{1}{|c|}{2.2/90.4} & \multicolumn{1}{|c|}{0.7/90.1} & \multicolumn{1}{|c|}{0.9/90.6}\\
				\hline
			\end{tabular}\label{tab:ASR_ACC_retrain_RE}}
	\end{center}
	\vspace{-0.15in}
\end{table}

\vspace{-0.025in}
\section{Conclusions}

In this paper, we proposed an unsupervised defense deployed before/during the training phase of DNN classifiers against imperceptible backdoors. Our defense is the first to jointly detect any backdoor attacks and remove the backdoor training images, if there are any, by subtracting the reverse-engineered backdoor pattern. Our defense outperforms other defenses deployed before/during classifier training.

\bibliography{Biblio}
\bibliographystyle{IEEEbib}

\vfill\pagebreak

\begin{center}
	\textbf{\large Supplementary Materials}
\end{center}

\section{Backdoor Patterns and Example Poisoned Images}

In the main paper, we evaluate the performance of the proposed reverse engineering (RE) defense using attacks involving six backdoor patterns. Here we provide details of the generation process of each backdoor pattern. Note that pattern generation may involve some randomness (e.g. in selecting the pixels to be perturbed for the local patterns). But once a backdoor pattern is generated for an attack, it is {\it commonly} embedded in all backdoor training images.

Pattern A is a ``chessboard'' pattern which is also considered by \cite{Post-TNNLS}. For each pair of neighboring pixels, one and only one pixel is perturbed positively. Here, we set the perturbation size to 2/255 for all pixels being perturbed.

Pattern B is considered by \cite{Haoti}. A pixel $(i, j)$ is perturbed positively if and only if $i$, $j$ are both even numbers\footnote{Pixel indices start from 0.}. The perturbation size is 3/255 for all pixels being perturbed.

Pattern C, D, and F, i.e. the ``cross'', the ``square'', and the ``L'', are crafted (with a little modification) based on the backdoor patterns used by \cite{SS}. However, perturbation for pattern C and pattern F is applied to all the three channels (i.e. colors) with perturbation size 70/255. For pattern D, we perturb only the first channel with perturbation size 80/255. The location of the ``cross'', the ``square'', and the ``L'' are randomly selected.

Pattern E is also considered by \cite{Post-TNNLS}, where four pixels are perturbed in one of the three channels. The location of the pixel to be perturbed, the channel to be perturbed, and the sign of the perturbation (either positive or negative) are all randomly selected. The absolute perturbation size is also randomly selected from the set $\{80/255, \ldots, 96/255\}$.

In Figure \ref{fig:bd_images_example}, for each backdoor pattern A-F, we show three example images embedded with the backdoor pattern and their originally clean, backdoor-free images.

\begin{figure}[t]
	\centering
	\begin{minipage}[b]{.45\linewidth}
		\centering
		\centerline{\includegraphics[width=\linewidth]{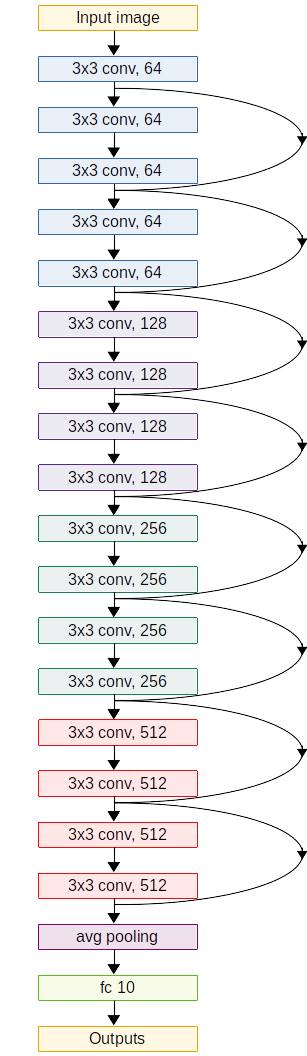}}
		\subcaption{wide}
	\end{minipage}
	\begin{minipage}[b]{.45\linewidth}
		\centering
		\centerline{\includegraphics[width=\linewidth]{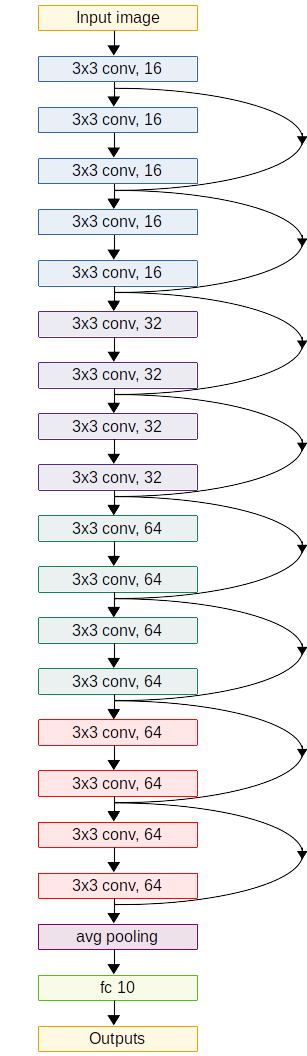}}
		\subcaption{compact}
	\end{minipage}
	\caption{The wide ResNet-18 architecture (left) and the compact ResNet-18 architecture (right) used in the main paper to evaluate attack effectiveness against defenseless DNNs and to evaluate the performance of the proposed RE defense.}
	\label{fig:ResNet18_architectures}
\end{figure}

\begin{figure*}
	\centering
	\begin{minipage}[b]{.3\linewidth}
		\centering
		{\includegraphics[width=.55\linewidth]{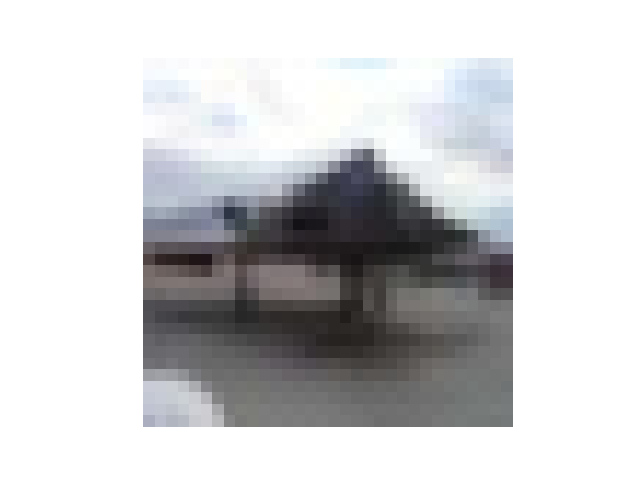}}%
		{\includegraphics[width=.55\linewidth]{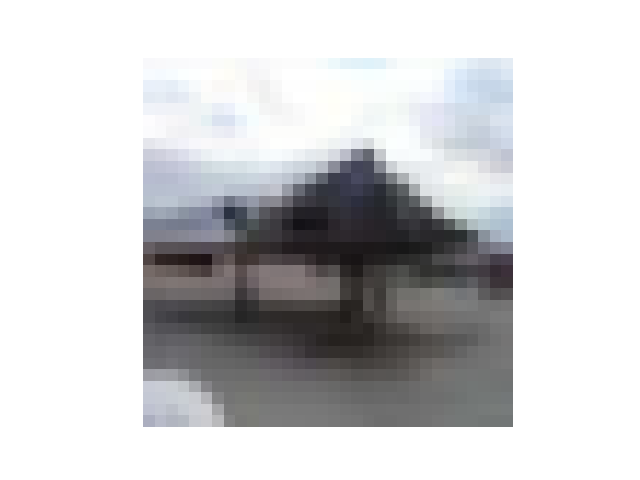}}
		{\includegraphics[width=.55\linewidth]{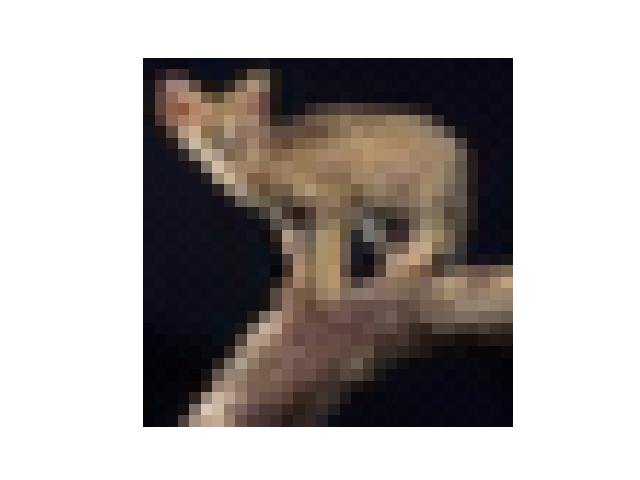}}%
		{\includegraphics[width=.55\linewidth]{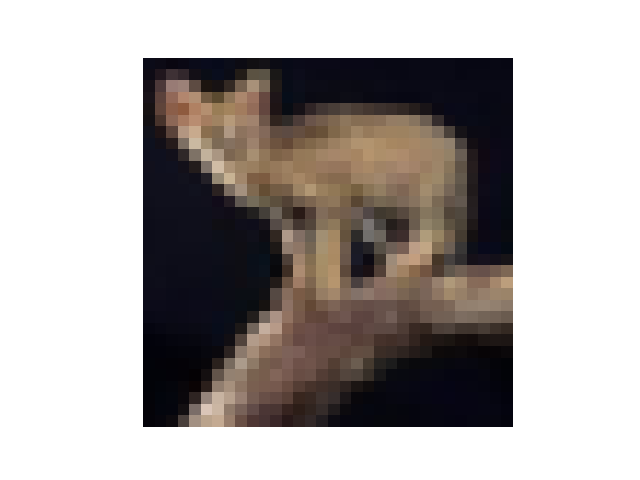}}
		{\includegraphics[width=.55\linewidth]{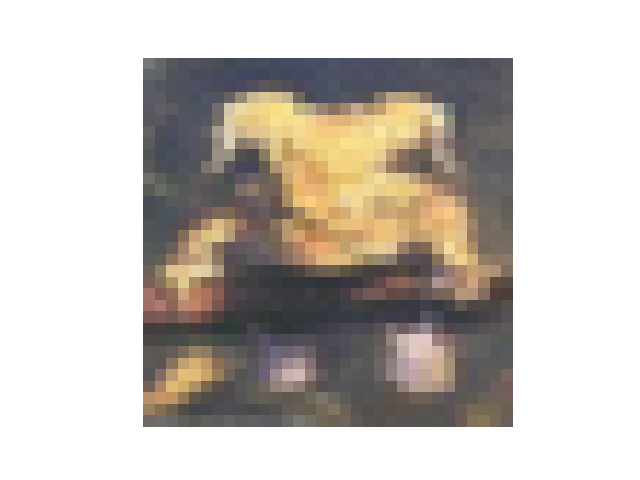}}%
		{\includegraphics[width=.55\linewidth]{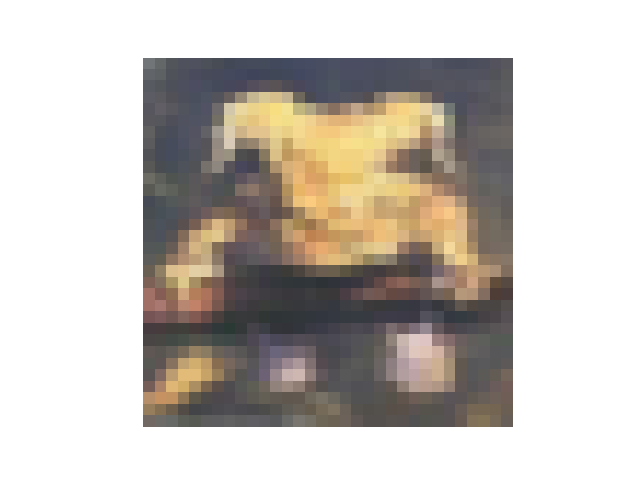}}
		\subcaption{pattern A}
	\end{minipage}
	\begin{minipage}[b]{.3\linewidth}
		\centering
		{\includegraphics[width=.55\linewidth]{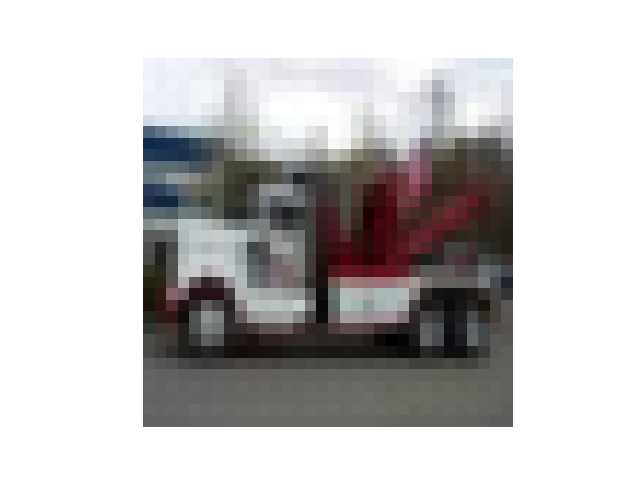}}%
		{\includegraphics[width=.55\linewidth]{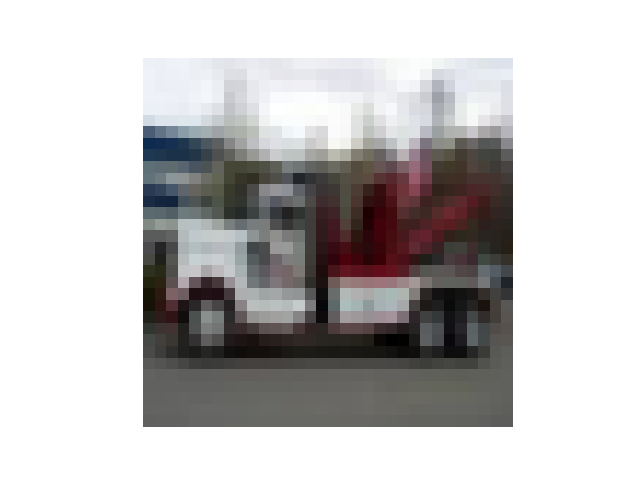}}
		{\includegraphics[width=.55\linewidth]{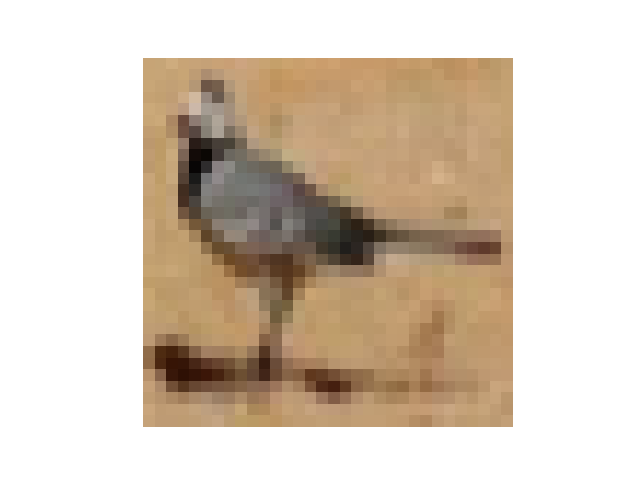}}%
		{\includegraphics[width=.55\linewidth]{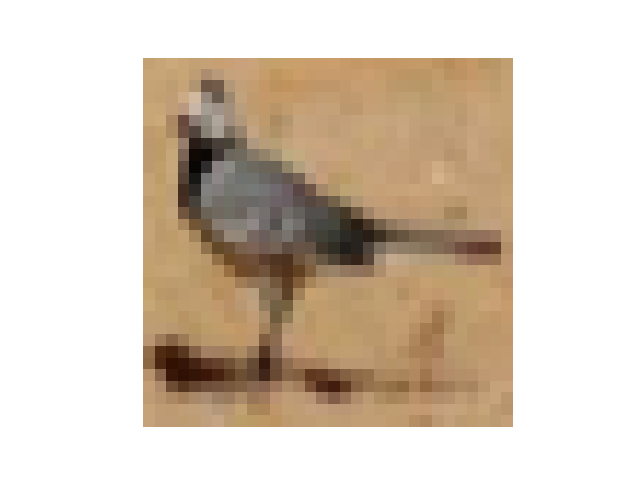}}
		{\includegraphics[width=.55\linewidth]{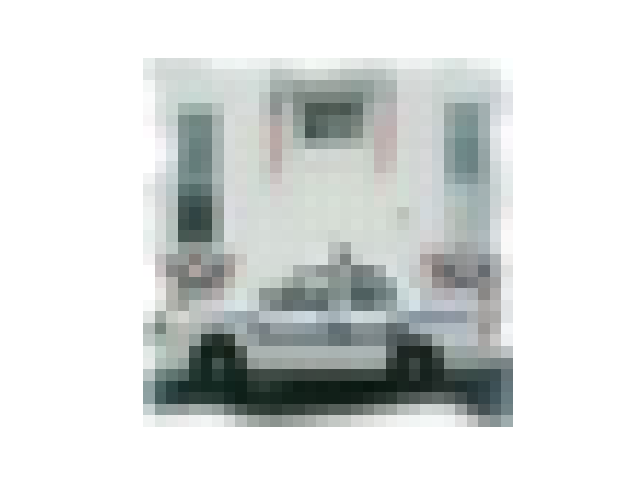}}%
		{\includegraphics[width=.55\linewidth]{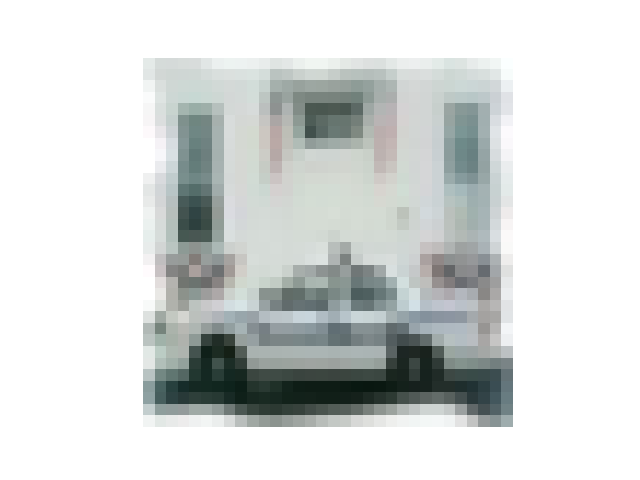}}
		\subcaption{pattern B}
	\end{minipage}
	\begin{minipage}[b]{.3\linewidth}
		\centering
		{\includegraphics[width=.55\linewidth]{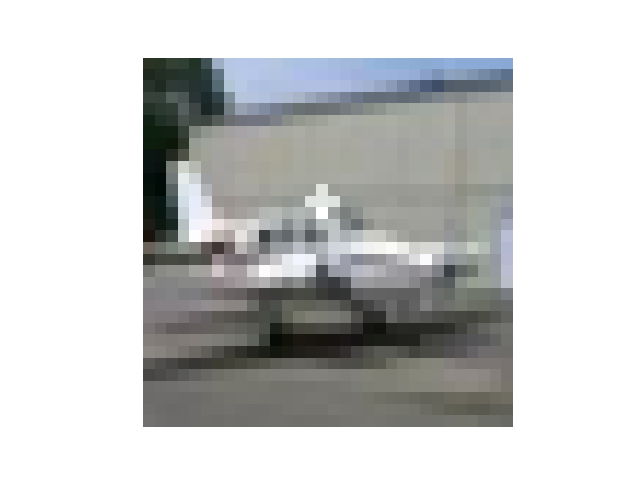}}%
		{\includegraphics[width=.55\linewidth]{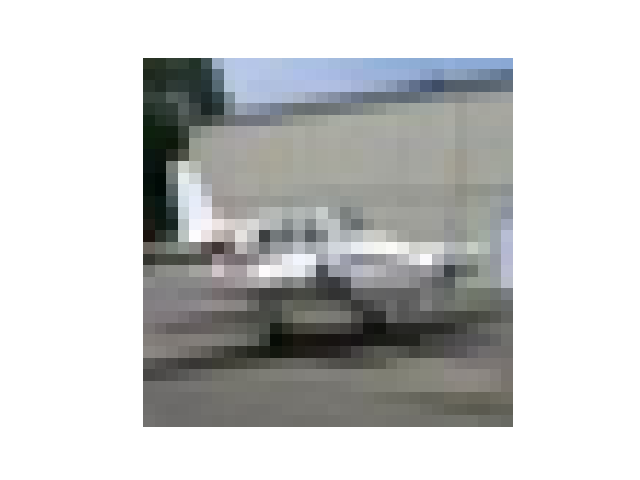}}
		{\includegraphics[width=.55\linewidth]{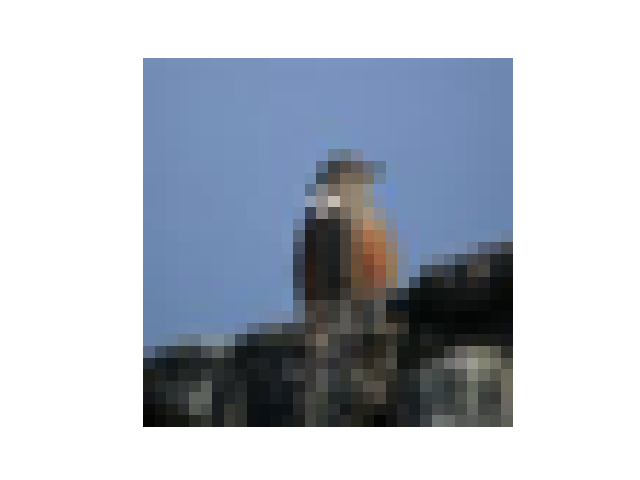}}%
		{\includegraphics[width=.55\linewidth]{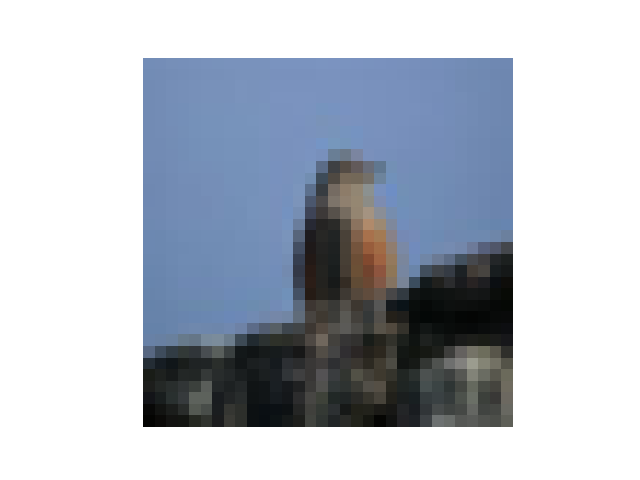}}
		{\includegraphics[width=.55\linewidth]{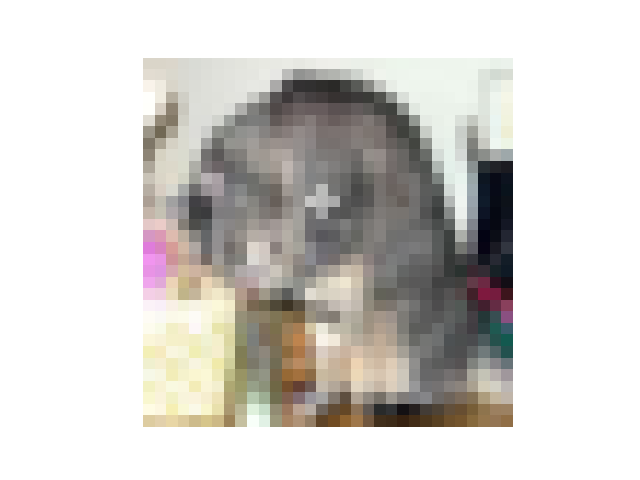}}%
		{\includegraphics[width=.55\linewidth]{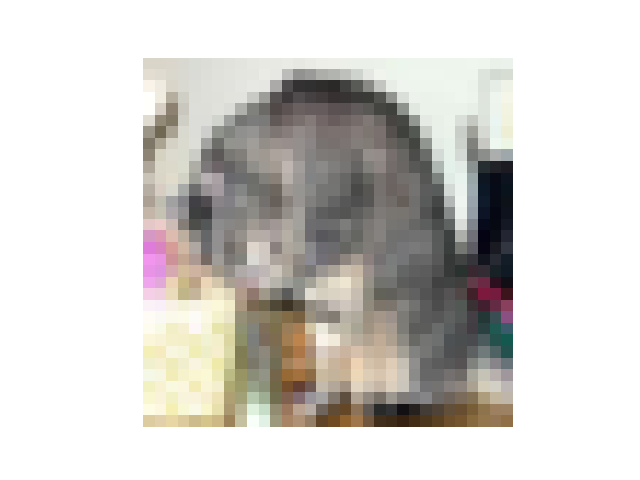}}
		\subcaption{pattern C}
	\end{minipage}
	\begin{minipage}[b]{.3\linewidth}
		\centering
		{\includegraphics[width=.55\linewidth]{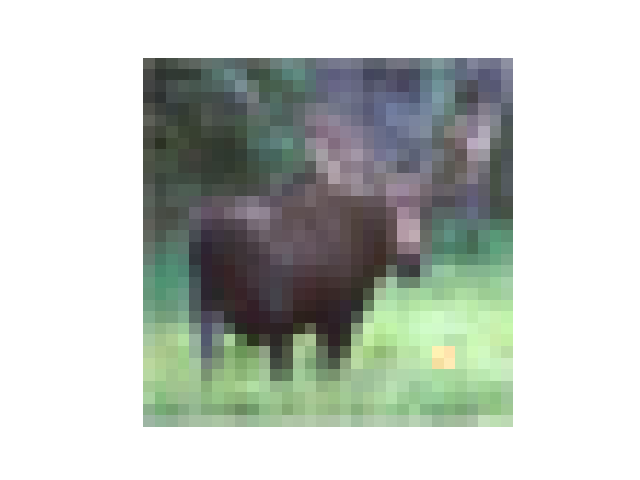}}%
		{\includegraphics[width=.55\linewidth]{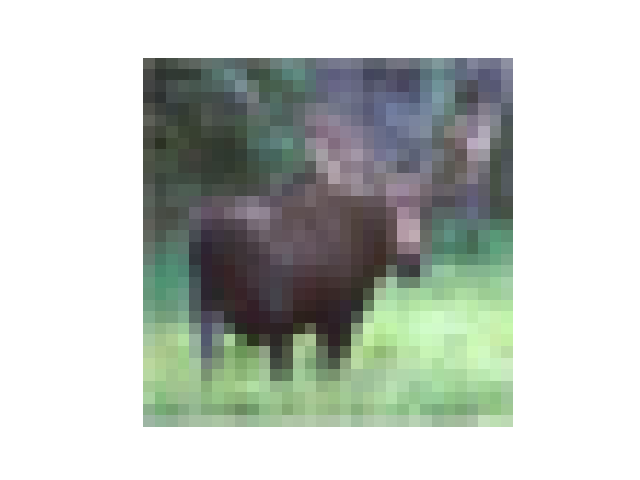}}
		{\includegraphics[width=.55\linewidth]{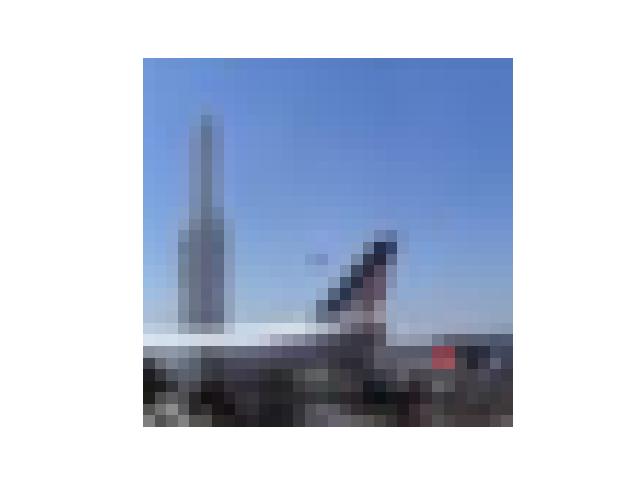}}%
		{\includegraphics[width=.55\linewidth]{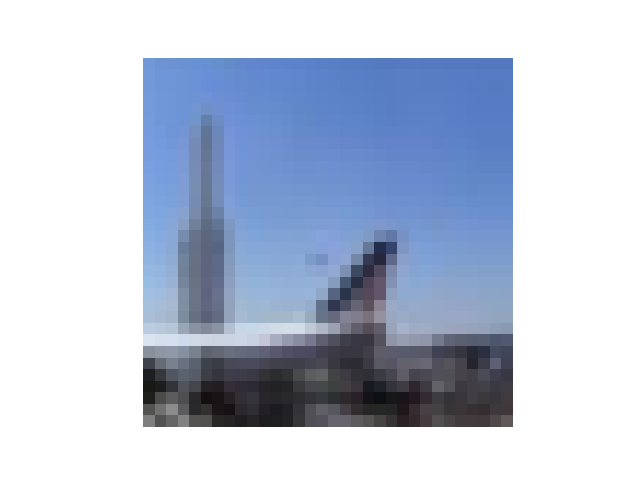}}
		{\includegraphics[width=.55\linewidth]{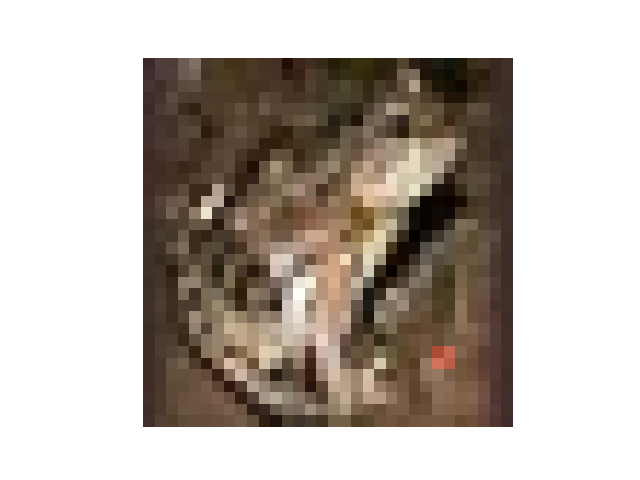}}%
		{\includegraphics[width=.55\linewidth]{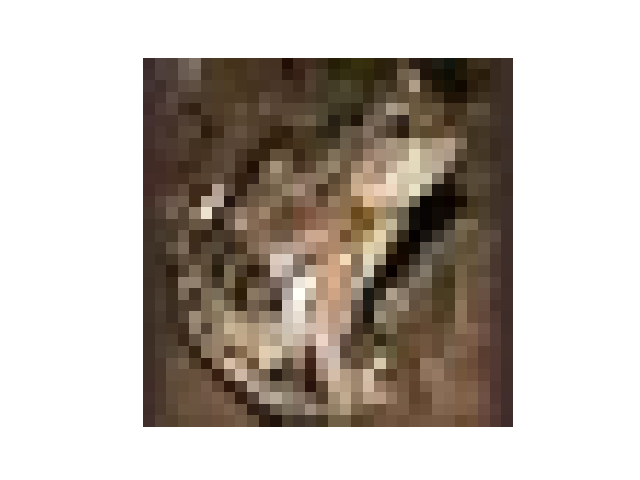}}
		\subcaption{pattern D}
	\end{minipage}
	\begin{minipage}[b]{.3\linewidth}
		\centering
		{\includegraphics[width=.55\linewidth]{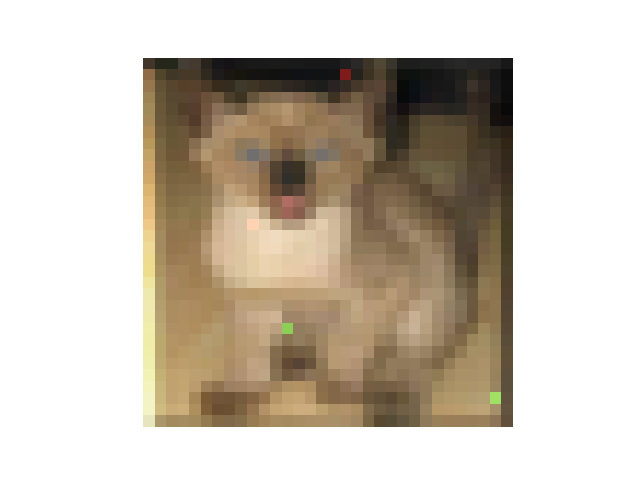}}%
		{\includegraphics[width=.55\linewidth]{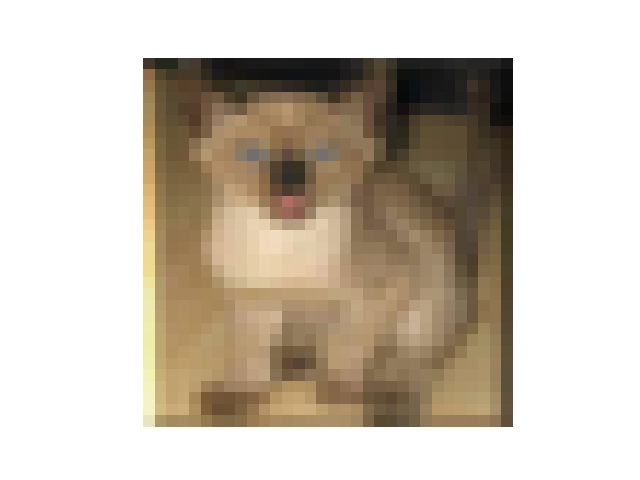}}
		{\includegraphics[width=.55\linewidth]{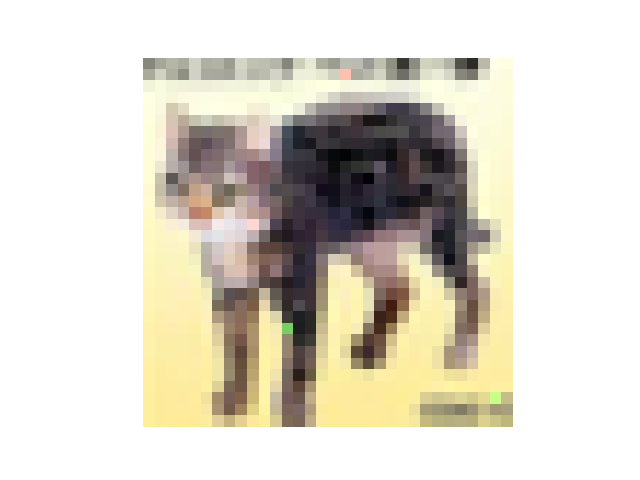}}%
		{\includegraphics[width=.55\linewidth]{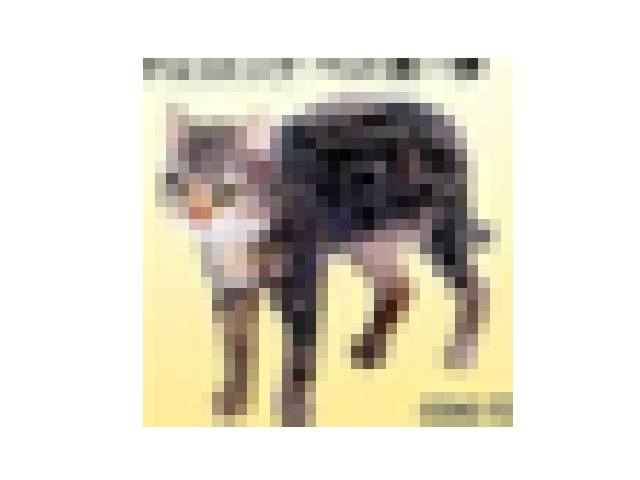}}
		{\includegraphics[width=.55\linewidth]{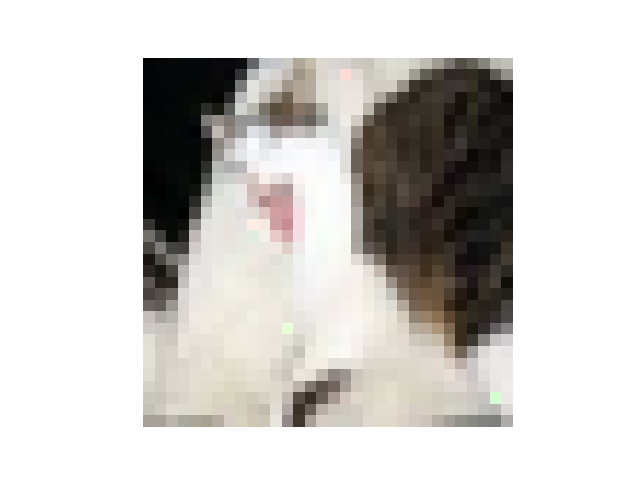}}%
		{\includegraphics[width=.55\linewidth]{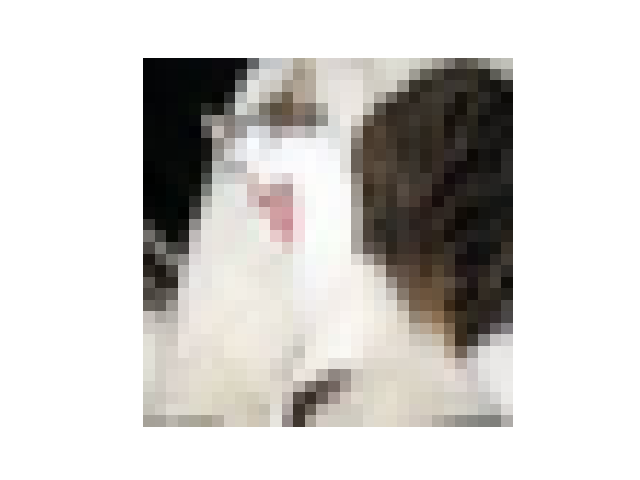}}
		\subcaption{pattern E}
	\end{minipage}
	\begin{minipage}[b]{.3\linewidth}
		\centering
		{\includegraphics[width=.55\linewidth]{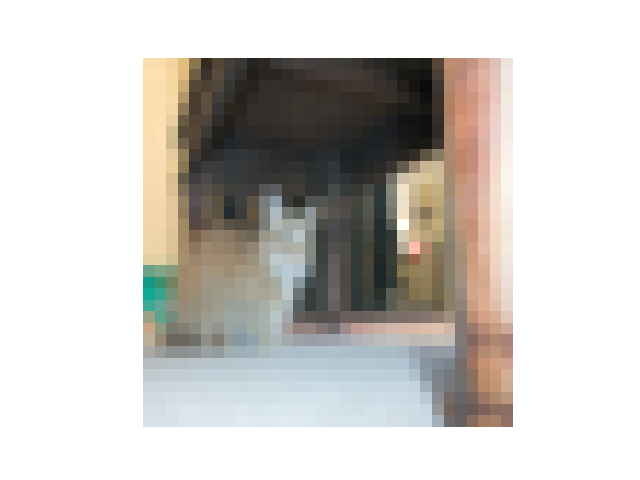}}%
		{\includegraphics[width=.55\linewidth]{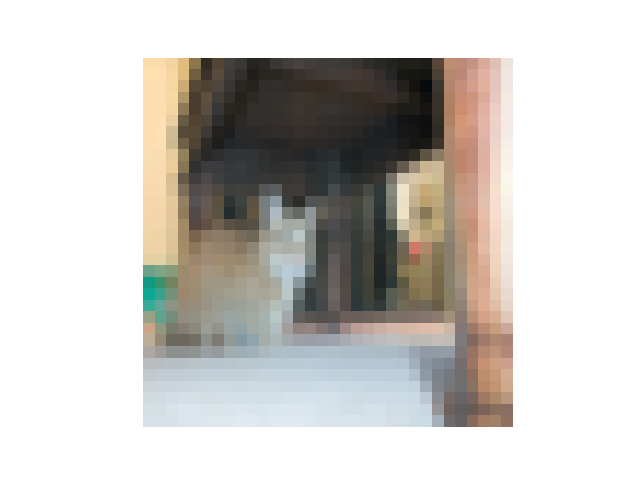}}
		{\includegraphics[width=.55\linewidth]{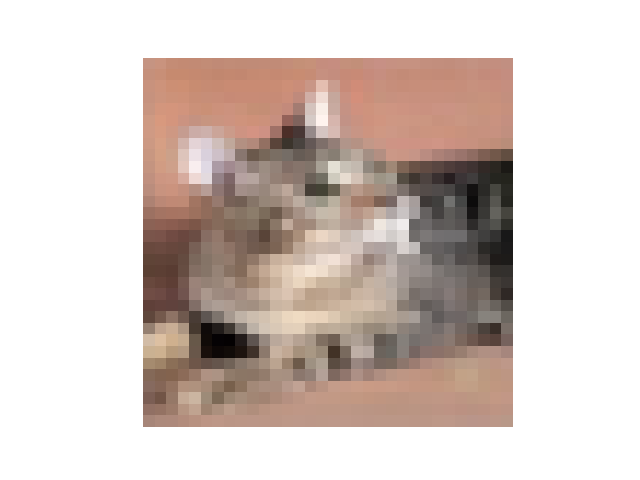}}%
		{\includegraphics[width=.55\linewidth]{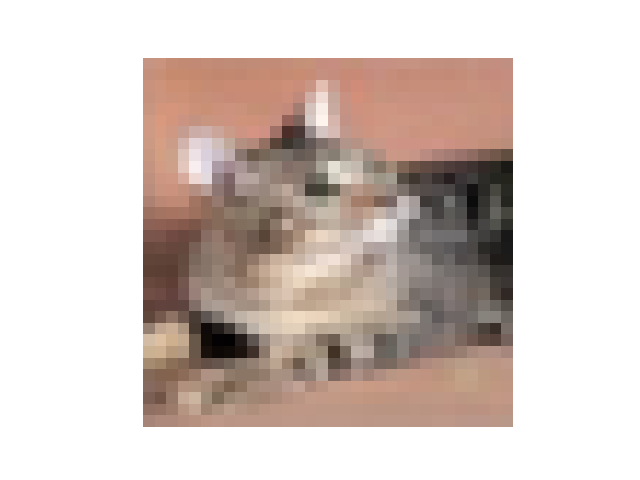}}
		{\includegraphics[width=.55\linewidth]{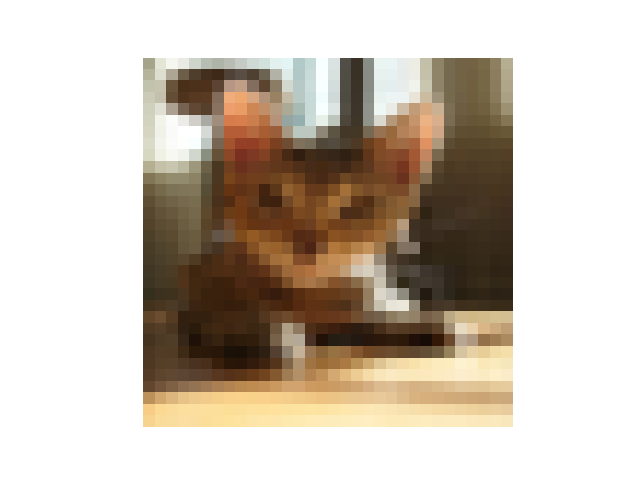}}%
		{\includegraphics[width=.55\linewidth]{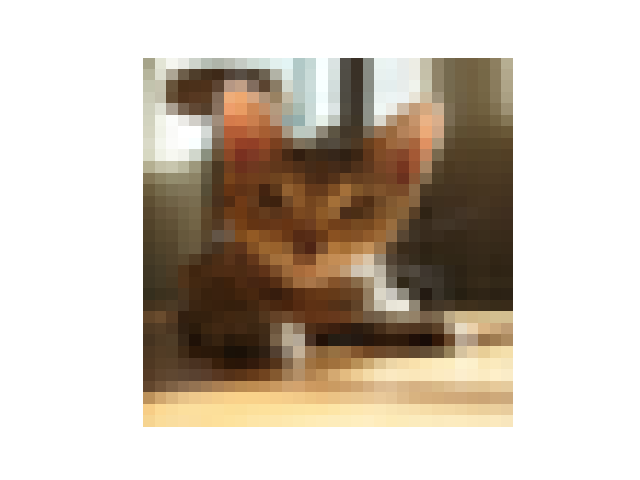}}
		\subcaption{pattern F}
	\end{minipage}
	\caption{Examples of poisoned images with backdoor pattern A-F embedded and the originally clean images. For each pair of images, the image with the backdoor pattern embedded is on the left, and the backdoor-free image is on the right.}
	\label{fig:bd_images_example}
\end{figure*}

\section{Choice of Source Class(es) for Each Attack in the Baseline Experiment}

In the baseline experiments evaluating the effectiveness of the proposed RE defense, we create three attacks for each of the six backdoor patterns. The three attacks involve one source class (1SC), three source classes (3SC), and nine source classes (9SC), respectively. The source class(es) for each attack are arbitrarily selected by the authors. Here, we provide the detailed choices.

We enumerate the 10 classes of CIFAR-10, i.e. `plane', `car', `bird', `cat', `deer', `dog', `frog', `horse', `ship', `truck', as classes 1 to 10 for convenience. For all three attacks associated with each backdoor pattern, we choose the same target class. The choices of the source class(es) ${\mathcal S}^{\ast}$ and the target class $t^{\ast}$ are summarized in Table \ref{tab:choice_SC}.

\begin{table}[t]
	\begin{center}
		\caption{Choices of the source class(es) ${\mathcal S}^{\ast}$ and the target class $t^{\ast}$ for the 18 attacks (1SC, 3SC, and 9SC attack for pattern A-F) created in the baseline experiment (for evaluating the performance of the proposed RE defense) in the main paper.}
		\resizebox{0.38\textwidth}{!}{
			\begin{tabular}{ |c|c|c|c|c| }
				\hline 
				& $t^{\ast}$ & ${\mathcal S}^{\ast}$ of 1SC & ${\mathcal S}^{\ast}$ of 3SC & ${\mathcal S}^{\ast}$ of 9SC\\
				\hline
				A & 10 & 2 & 2, 5, 8 & except 10\\
				\hline
				B & 8 & 2 & 2, 3, 10 & except 8\\
				\hline
				C & 10 & 2 & 5, 7, 8 & except 10\\
				\hline
				D & 4 & 9 & 8, 9, 10 & except 4\\
				\hline
				E & 7 & 4 & 2, 4, 6 & except 7\\
				\hline
				F & 9 & 4 & 4, 5, 6 & except 9\\
				\hline
			\end{tabular}\label{tab:choice_SC}}
	\end{center}
\end{table}

\section{Details of the Defenseless DNNs Used to Evaluate Attack Effectiveness}

\subsection{DNN Architectures}

In the main paper, our main experiments on CIFAR-10 involve two DNN architectures from the ResNet family \cite{ResNet}. A standard ResNet architecture is featured by four groups of residual blocks. A (``non-bottleneck'') residual block consists of two identical, consecutive convolutional layers (i.e. two sets of identical convolutional filters) in parallel with an identity map (see Figure 2 of \cite{ResNet}). For example, for an 18-layer ResNet, there are 2 residual blocks in each of the four groups of residual blocks. In our experiments, we consider a wide architecture where there are 64, 128, 256, 512 filters for each convolutional layer for the four groups of residual blocks, respectively, and a compact architecture with 16, 32, 64, 64 filters instead (see Figure \ref{fig:ResNet18_architectures}). Conceptually, the wide DNN architecture has a stronger representation power than the compact DNN architecture, but it also contains more parameters and requires more training resources and training time than the compact DNN architecture to achieve a high test accuracy. For a dataset containing a large number of classes, e.g. ImageNet, using a wide DNN structure can achieve a clearly higher test accuracy than using a compact DNN architecture. But for a dataset containing relatively small number of classes, e.g. CIFAR-10, MNIST, SVHN, using a compact DNN architecture achieves sufficiently high test accuracy with much less training resources/time required than using a wide DNN architecture.

\subsection{Training Configurations}

In the baseline experiments of the main paper, the training of the defenseless DNNs is performed on the poisoned CIFAR-10 training set. CIFAR-10 contains 60k color images with size $32\times32$ evenly distributed in 10 classes. The training set contains 50k images (5k per class) and the test set contains the remaining 10k images. Note that in our experiments, the backdoor training images are crafted using clean training images of the CIFAR-10 training set due to the limited number of training images. In practice, an attacker may not have access to the training set. In other words, the originally clean images of the backdoor training images are collected by the attacker and are not typically present in the training set. Hence in our experiments, once a training image from the CIFAR-10 training set is used to create a backdoor training image, its clean, backdoor-free version will no longer appear in the training set. Theoretically, our training set cleansing problem can be much easier without this setting. Suppose $f(\cdot;\Theta^{\ast})$ is the DNN trained on the poisoned training set where the backdoor pattern is ${\bf v}^{\ast}$. If for any backdoor training image ${\bf x}$, its clean version $\tilde{\bf x}$ also exists in the training set, we will have $p_s(\tilde{\bf x};\Theta)$ very close to 1 for some $s\in{\mathcal S}^{\ast}$ where ${\mathcal S}^{\ast}$ is the set of source classes. Then ${\bf v}^{\ast}$ will be associated with a high peak of objective of (9) in the main paper.

For both DNN architectures, our training is performed for 100 epochs, with batch size 32 and learning rate 0.001. Adam optimizer is used with decay rate 0.9 and 0.999 for the first and second moment, respectively. Training data augmentation is used, including random crop, random horizontal flip, and random rotation.

\section{Additional Experiments on CIFAR-10 with More Recent DNN Architectures}

In the main paper, we compared our proposed RE defense with other existing backdoor defenses deployed before/during the classifier's training phase, including the spectral signature (SS) defense \cite{SS}, the activation clustering (AC) defense \cite{AC}, and the cluster impurity (CI) defense \cite{CI}, on 18 attacks on the CIFAR-10 dataset. We considered a wide ResNet architecture and a compact ResNet architecture that could possibly be chosen by the defender. Here, we replicate the baseline experiment in the main paper for two more recent DNN architectures -- MobileNetV2\footnote{Link of implementation: \url{https://github.com/kuangliu/pytorch-cifar/blob/master/models/mobilenetv2.py}}. \cite{MobileNetV2} and DenseNet\footnote{Link of implementation: \url{https://github.com/gpleiss/efficient_densenet_pytorch/blob/master/models/densenet.py}} \cite{DenseNet}. The MobileNet family is designed to fulfill resource restrictions, such that it can be deployed on mobile devices. The DenseNet family exploits ``feature reuse'' by passing the feature map of each layer to all\footnote{In practice, a DenseNet DNN is usually implemented as several consecutive ``densely connected'' blocks. In each block, the feature map of a layer is passed to all the subsequent layers in the same block.} subsequent layers, forming a ``dense'' connection between layers. We emphasize again that the DNN architecture for training the classifier on the possibly poisoned training set, as part of the defense for SS, AC, CI, and our RE, is determined by the defender. Theoretically, this DNN architecture can even be different from the architecture of the final classifier retrained after the training set cleansing step. Hence, an effective defense should perform uniformly well no matter what DNN architecture is chosen, if an initial training on the possibly poisoned training set is required.

\begin{table}[t]
	\begin{center}
		\caption{Attack success rate (ASR) and the accuracy (ACC) on clean test set (jointly represented by ASR/ACC) for the 6 attacks for MobileNetV2 and DenseNet; and the test ACC of the clean benchmark DNNs.}
		\resizebox{0.43\textwidth}{!}{
			\begin{tabular}{ |c|c|c|c|c| }
				\hline 
				& \multicolumn{2}{|c|}{MobileNetV2} & \multicolumn{2}{|c|}{DenseNet}\\
				\hline
				pattern & 3SC & Clean & 3SC & Clean\\
				\hline
				A & 92.1/99.9 & n.a./93.8 & 91.5/99.4 & n.a./94.0\\
				\hline
				B & 91.5/98.1 & n.a./93.1 & 91.6/98.7 & n.a./94.4\\
				\hline
				C & 91.5/99.7 & n.a./92.9 & 92.2/99.7 & n.a./93.8\\
				\hline
				D & 91.6/95.1 & n.a./93.2 & 92.2/94.7 & n.a./94.2\\
				\hline
				E & 91.3/97.7 & n.a./93.2 & 91.5/94.7 & n.a./94.8\\
				\hline
				F & 91.9/94.4 & n.a./93.1 & 91.7/94.7 & n.a./94.2\\
				\hline
			\end{tabular}\label{tab:ASR_ACC_MobileNetV2_DenseNet}}
	\end{center}
\end{table}

For simplicity, this experiment involves only attacks with 3 source classes, i.e. the 6 3SC attacks in the main paper. Similar to the experiments in the main paper, we first show, for both the MobileNetV2 and DenseNet architectures, that the attacks are effective when there is no defense. Then we apply our RE defense and its competitors (i.e., SS, AC, and CI) to the backdoor-poisoned training set for each attack. We report the detection and training set cleansing results for all these defenses; but we do not perform retraining on the sanitized training set for concision.

In Table \ref{tab:ASR_ACC_MobileNetV2_DenseNet}, we show the attack success rate (ASR) and clean test accuracy (ACC) of the 6 attacks when there is no defense. Again, for each attack, we also train a classifier (for each of the two DNN architectures) on the clean, poisoning-free training set. For both DNN architectures, all six attacks are successful with high ASR and small degradation in ACC.

\begin{table}[t]
	\caption{Detection results of (a) RE defense, in comparison with (b) AC and (c) CI, on the 6 poisoned training sets and the clean training sets for both MobileNetV2 and DenseNet architectures. Symbols $\otimes$, $\oslash$, and $\odot$ represent: attack is not detected (or falsely detected for clean training set), attack is detected but the target class is incorrectly inferred, and attack is detected and the target class is correctly inferred (or no attack is detected for clean training set), respectively.}\label{tab:detection_MobileNetV2_DenseNet}
	\centering
	\begin{subtable}[t]{1\columnwidth}
		\centering
		\resizebox{0.7\textwidth}{!}{
			\begin{tabular}{ |c|c|c|c|c|c|c|c| }
				\hline 
				\multicolumn{2}{|c|}{pattern} & A & B & C & D & E & F\\
				\hline
				\multirow{2}{*}{\thead{MobileNetV2}} 
				& \multicolumn{1}{|c|}{clean} & \multicolumn{1}{|c|}{$\odot$} & \multicolumn{1}{|c|}{$\odot$} & \multicolumn{1}{|c|}{$\odot$} & \multicolumn{1}{|c|}{$\odot$} & \multicolumn{1}{|c|}{$\odot$} & \multicolumn{1}{|c|}{$\odot$}\\
				\cline{2-8}
				& \multicolumn{1}{|c|}{3SC} & \multicolumn{1}{|c|}{$\odot$} & \multicolumn{1}{|c|}{$\odot$} & \multicolumn{1}{|c|}{$\odot$} & \multicolumn{1}{|c|}{$\odot$} & \multicolumn{1}{|c|}{$\odot$} & \multicolumn{1}{|c|}{$\odot$}\\
				\hline
				\multirow{2}{*}{\thead{DenseNet}} 
				& \multicolumn{1}{|c|}{clean} & \multicolumn{1}{|c|}{$\odot$} & \multicolumn{1}{|c|}{$\odot$} & \multicolumn{1}{|c|}{$\odot$} & \multicolumn{1}{|c|}{$\odot$} & \multicolumn{1}{|c|}{$\odot$} & \multicolumn{1}{|c|}{$\odot$}\\
				\cline{2-8}
				& \multicolumn{1}{|c|}{3SC} & \multicolumn{1}{|c|}{$\odot$} & \multicolumn{1}{|c|}{$\odot$} & \multicolumn{1}{|c|}{$\odot$} & \multicolumn{1}{|c|}{$\odot$} & \multicolumn{1}{|c|}{$\odot$} & \multicolumn{1}{|c|}{$\odot$}\\
				\hline
		\end{tabular}}
		\subcaption{RE detection}\label{subtab:detection_MobileNetV2_DenseNet_RE}
	\end{subtable}
	\begin{subtable}[t]{0.7\columnwidth}
		\centering
		\resizebox{\textwidth}{!}{
			\begin{tabular}{ |c|c|c|c|c|c|c|c| }
				\hline 
				\multicolumn{2}{|c|}{pattern} & A & B & C & D & E & F\\
				\hline
				\multirow{2}{*}{\thead{MobileNetV2}} 
				& \multicolumn{1}{|c|}{clean} & \multicolumn{1}{|c|}{$\odot$} & \multicolumn{1}{|c|}{$\odot$} & \multicolumn{1}{|c|}{$\oslash$} & \multicolumn{1}{|c|}{$\odot$} & \multicolumn{1}{|c|}{$\oslash$} & \multicolumn{1}{|c|}{$\odot$}\\
				\cline{2-8}
				& \multicolumn{1}{|c|}{3SC} & \multicolumn{1}{|c|}{$\odot$} & \multicolumn{1}{|c|}{$\odot$} & \multicolumn{1}{|c|}{$\odot$} & \multicolumn{1}{|c|}{$\odot$} & \multicolumn{1}{|c|}{$\odot$} & \multicolumn{1}{|c|}{$\odot$}\\
				\hline
				\multirow{2}{*}{\thead{DenseNet}} 
				& \multicolumn{1}{|c|}{clean} & \multicolumn{1}{|c|}{$\odot$} & \multicolumn{1}{|c|}{$\odot$} & \multicolumn{1}{|c|}{$\odot$} & \multicolumn{1}{|c|}{$\odot$} & \multicolumn{1}{|c|}{$\odot$} & \multicolumn{1}{|c|}{$\odot$}\\
				\cline{2-8}
				& \multicolumn{1}{|c|}{3SC} & \multicolumn{1}{|c|}{$\odot$} & \multicolumn{1}{|c|}{$\odot$} & \multicolumn{1}{|c|}{$\oslash$} & \multicolumn{1}{|c|}{$\oslash$} & \multicolumn{1}{|c|}{$\oslash$} & \multicolumn{1}{|c|}{$\oslash$}\\
				\hline
		\end{tabular}}
		\subcaption{AC detection}\label{subtab:detection_MobileNetV2_DenseNet_AC}
	\end{subtable}
	\begin{subtable}[t]{0.7\columnwidth}
		\centering
		\resizebox{\textwidth}{!}{
			\begin{tabular}{ |c|c|c|c|c|c|c|c| }
				\hline 
				\multicolumn{2}{|c|}{pattern} & A & B & C & D & E & F\\
				\hline
				\multirow{2}{*}{\thead{MobileNetV2}} 
				& \multicolumn{1}{|c|}{clean} & \multicolumn{1}{|c|}{$\odot$} & \multicolumn{1}{|c|}{$\odot$} & \multicolumn{1}{|c|}{$\odot$} & \multicolumn{1}{|c|}{$\odot$} & \multicolumn{1}{|c|}{$\otimes$} & \multicolumn{1}{|c|}{$\odot$}\\
				\cline{2-8}
				& \multicolumn{1}{|c|}{3SC} & \multicolumn{1}{|c|}{$\otimes$} & \multicolumn{1}{|c|}{$\otimes$} & \multicolumn{1}{|c|}{$\otimes$} & \multicolumn{1}{|c|}{$\otimes$} & \multicolumn{1}{|c|}{$\otimes$} & \multicolumn{1}{|c|}{$\otimes$}\\
				\hline
				\multirow{2}{*}{\thead{DenseNet}} 
				& \multicolumn{1}{|c|}{clean} & \multicolumn{1}{|c|}{$\odot$} & \multicolumn{1}{|c|}{$\odot$} & \multicolumn{1}{|c|}{$\odot$} & \multicolumn{1}{|c|}{$\odot$} & \multicolumn{1}{|c|}{$\odot$} & \multicolumn{1}{|c|}{$\odot$}\\
				\cline{2-8}
				& \multicolumn{1}{|c|}{3SC} & \multicolumn{1}{|c|}{$\otimes$} & \multicolumn{1}{|c|}{$\otimes$} & \multicolumn{1}{|c|}{$\otimes$} & \multicolumn{1}{|c|}{$\otimes$} & \multicolumn{1}{|c|}{$\otimes$} & \multicolumn{1}{|c|}{$\otimes$}\\
				\hline
		\end{tabular}}
		\subcaption{CI detection}\label{subtab:detection_MobileNetV2_DenseNet_CI}
	\end{subtable}
\end{table}

In Table \ref{tab:detection_MobileNetV2_DenseNet}, we show the detection performance of AC, CI, and RE. Again, SS does not propose any detection method; hence it is not included in the table. We also set the detection threshold for AC and CI as 0.23 and 0.36, respectively, for MobileNetV2, 0.24 and 0.46, respectively, for DenseNet. These thresholds are the minimum for AC and CI to achieve zero false detections on clean training sets. For RE, we use the same settings as described in the implementation detail section of the main paper. Our RE outperforms both AC and CI by successfully detecting all the attacks, with no false detections on clean training sets. Note that CI fails to detect any attacks because there is, again, only one component estimated by Gaussian mixture modeling (with the number of components selected by Bayesian information criterion) for the true target class, for all the attacks.

\begin{table}[t]
	\caption{Training set cleansing true positive rate (TPR) and false positive rate (FPR) of SS, AC, and our RE (represented in TPR/FPR form), for the 6 3SC attacks, for (a) the MobileNetV2 architecture, and (b) the DenseNet architecture.}\label{tab:cleansing_MobileNetV2_DenseNet}
	\begin{subtable}[t]{1\columnwidth}
		\centering
		\resizebox{\textwidth}{!}{
			\begin{tabular}{ |c|c|c|c|c|c|c| }
				\hline 
				& A & B & C & D & E & F\\
				\hline
				SS & 44.5/15.0 & 1.0/20.3 & 7.7/19.4 & 77.8/11.0 & 45.8/14.9 & 2.7/20.0\\
				\hline
				AC & 84.2/35.6 & 34.2/12.6 & 48.8/38.6 & 88.3/{\bf 9.5} & 84.3/30.9 & 52.7/22.4\\
				\hline
				RE & {\bf 93.0}/{\bf 9.1} & {\bf 99.2}/{\bf 9.62} & {\bf 99.7}/15.3 & {\bf 96.8}/{\bf 4.9} & {\bf 90.8}/{\bf 1.34} & {\bf 98.8}/13.6\\
				\hline
		\end{tabular}}
		\subcaption{MobileNetV2 architecture}\label{subtab:cleansing_MobileNetV2}
	\end{subtable}
	\begin{subtable}[t]{1\columnwidth}
		\centering
		\resizebox{\textwidth}{!}{
			\begin{tabular}{ |c|c|c|c|c|c|c| }
				\hline 
				& A & B & C & D & E & F\\
				\hline
				SS & 88.5/{\bf 9.7} & {\bf 95.8}/{\bf 8.9} & 74.8/11.2 & 70.3/11.9 & 43.2/15.2 & 14.7/18.6\\
				\hline
				AC & 85.7/{\bf 0} & 78.2/{\bf 0} & {\bf 97.3}/37.8 & {\bf 93.0}/23.5 & 70.7/26.5 & 22.2/49.2\\
				\hline
				RE & 88.2/{\bf 0} & {\bf 97.7}/{\bf 0.1} & 88.0/17.2 & {\bf 96.5}/{\bf 3.6} & {\bf 91.1}/11.4 & {\bf 91.5}/15.8\\
				\hline
		\end{tabular}}
		\subcaption{DenseNet architecture}\label{subtab:cleansing_DenseNet}
	\end{subtable}
\end{table}

In Table \ref{tab:cleansing_MobileNetV2_DenseNet}, we show the training set cleansing true positive rate (TPR) and false positive rate (FPR) of SS, AC, and our RE, against the six 3SC attacks, for the MobilNetV2 and the DensNet architectures. Similar to the main paper, we do not include the results for CI here because it estimated one component for the true target class for all the six attacks, such that no backdoor training images will be identified. For both DNN architectures, our RE generally outperforms AC and CI, with very large gains on several poisoned training sets.

\section{Robustness Analysis of the Proposed RE Defense}

Our proposed RE defense requires setting the target group misclassification fraction $\pi$ for the pattern estimation algorithm (line 3-7 of Algorithm 1 of the main paper) and the Lagrange multiplier $\lambda$ for the surrogate objective function (Eq. (9) of the main paper). In both the main paper and the supplementary material, not including the current section, we use the default $\pi=0.95$ and $\lambda=1$ for our RE defense and achieved generally good performance in all the experiments. In this section, we discuss how the choices of $\pi$ and $\lambda$ will possibly affect the detection and the training set cleansing performance of RE. We also experimentally show the empirical range of choices for $\pi$ and $\lambda$ that ensures good performance of RE. We find that both $\pi$ and $\lambda$ can be easily and comprehensively determined with sufficient liberty.

\subsection{Reasonable Choices of $\pi$ and $\lambda$}

In principle, $\pi$ should be relatively large, such that {\it only} for the backdoor class pairs $(s, t^{\ast})$ with $s\in{\mathcal S}^{\ast}$, $\pi$ fraction of group misclassification from class $s$ to $t^{\ast}$ can be achieved with a relatively small perturbation. If $\pi$ is chosen too small, for non-backdoor class pairs, there could be a small perturbation (with norm similar to the true backdoor pattern) that induces a (small) $\pi$ fraction of images from the source class being misclassified to the target class. Then the detection will likely fail. Moreover, for a backdoor class pair, a pattern that only induces a small fraction of misclassifications from $s$ to $t^{\ast}$ will not likely be similar to the true backdoor pattern. ``Removing'' such a pattern from the backdoor training images labeled to class $t^{\ast}$ will likely not induce misclassifications for these images to classes other than $t^{\ast}$, which causes a low true positive rate in training set cleansing. However, if $\pi$ is chosen too large (e.g. 1.0), it may be hard for all class pairs, including the true backdoor class pair, to achieve the $\pi$ fraction of misclassifications from the source class to the target class. For a true backdoor class pair, while our iterative pattern estimation algorithm ``polishes'' the pattern to achieve the $\pi$ fraction of misclassifications from $s$ to $t^{\ast}$, the pattern estimation will also induce more clean images from $t^{\ast}$ to be misclassified when the pattern is ``removed'', which causes a high false positive rate. Hence, intuitively, $\pi$ should be chosen large, but not too close to 1.

Our default choice of $\lambda=1$ is straightforwardly based on the design of RE. By setting $\lambda=1$, for pattern estimation for all $(s, t)$ pairs, a balance is kept between: 1) searching a pattern to induce a high fraction of group misclassification from class $s$ to $t$; and 2) searching a pattern to cause misclassification when it is ``removed'' from the training images labeled to class $t$. If $\lambda$ is chosen too large, for a backdoor class pair $(s, t^{\ast})$ with $s\in{\mathcal S}^{\ast}$, the estimated pattern will likely be effective in inducing a high misclassification from $s$ to $t^{\ast}$, but will not induce misclassification when ``removed'' from the backdoor training images labeled to $t^{\ast}$. For example, for attacks involving the ``chessboard'' pattern (i.e. pattern A) of the main paper, if we let $\lambda$ go to infinity, for a backdoor class pair, the estimated pattern will be several ``patches'' of the chessboard pattern (see Figure \ref{subfig:be_est_AD_A}). ``Removing'' these patches from the backdoor training images will not remove the entire backdoor pattern and will possibly not induce misclassification to their source class(es). If $\lambda$ is chosen too small, for a backdoor class pair $(s, t^{\ast})$ with $s\in{\mathcal S}^{\ast}$, pattern estimation will focus on inducing misclassification to classes other than $t^{\ast}$ for all images labeled to $t^{\ast}$ (including both backdoor backdoor training images and clean training images), such that $\pi$ fraction misclassification will not be easily achieved with a small perturbation, causing a failure in detection. Moreover, the estimated pattern will possibly not be similar to the true backdoor pattern, and will cause a high fraction of misclassifications from class $t^{\ast}$ to ``not'' $t^{\ast}$, resulting in a high false positive rate for training set cleansing. Hence, intuitively, any $\lambda$ close to 1 should be a good choice.

\begin{figure}
	\centering
	\centerline{\includegraphics[width=\linewidth]{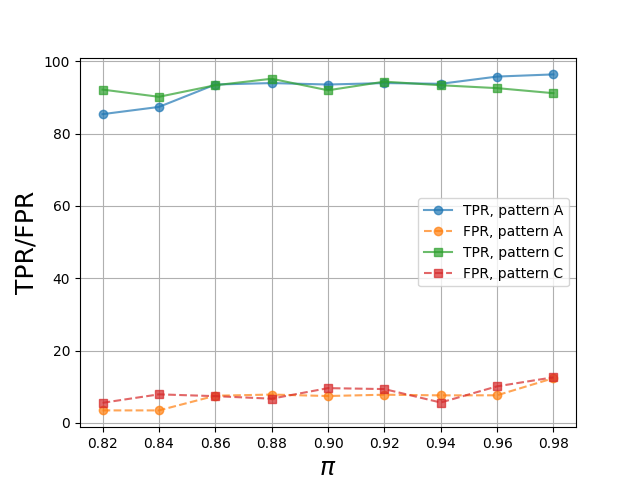}}
	\caption{True positive rate (TPR) and false positive rate (FPR) of training data cleansing of RE against the attack with pattern A and the attack with pattern C, respectively, for a wide range of $\pi$.}
	\label{fig:robustness_pi}
\end{figure}

\begin{figure}
	\centering
	\centerline{\includegraphics[width=\linewidth]{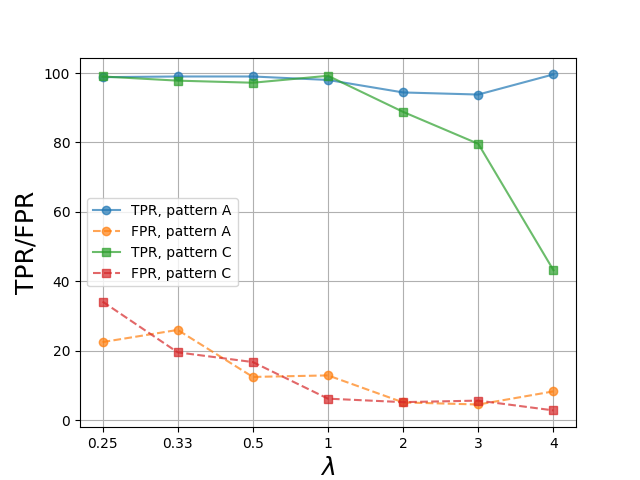}}
	\caption{True positive rate (TPR) and false positive rate (FPR) of training data cleansing of RE against the attack with pattern A and the attack with pattern C, respectively, for a wide range of $\lambda$}
	\label{fig:robustness_lambda}
\end{figure}

\subsection{Experimental Evaluation of RE Robustness}

In the following experiment, we test the detection and training set cleansing performance of the RE defense for a range of $\pi$ and a range of $\lambda$, respectively. We consider two 3-source-class (3SC) attacks, with one attack involving pattern A (a global pattern), and the other involving pattern C (a local pattern). We use the wide ResNet DNN architecture and the same training configurations as in the main paper to train the classifier (for defense purpose) on the poisoned training set. We also use the same configurations as in the main paper for RE, except that one of $\pi$ and $\lambda$ varies. Although we have strong intuitions (based on previous discussions) on the reasonable range of $\pi$ and $\lambda$, we experimentally evaluate RE for a much wider range of $\pi$ and $\lambda$, hoping to see performance drops of RE for either detection or training set cleansing. Hence, we consider $\pi$ for a range from 0.98 to as low as 0.82, and $\lambda$ for a range from 1/4 to 4.

\subsubsection{Detection}

Both attacks are successfully detected for $\pi\in\{0.82, 0.84, \ldots, 0.96, 0.98\}$ with the target class correctly inferred. The maximum order statistic p-values (over the range of $\pi$) for the attack with pattern A and the attack with pattern C are ``underflow'' (less than $10^{-323}$) and 0.018, respectively -- clearly below the detection confidence threshold $\theta=0.05$. For $\lambda\in\{1/4, 1/3, 1/2, 1, 2, 3, 4\}$, the maximum order statistic p-values for the attack with pattern A and the attack with pattern C are $10^{-16}$ and $4\times10^{-4}$, respectively -- far less than $\theta=0.05$. Also, the target class is correctly inferred.

\subsubsection{Training Set Cleansing}

In Figure \ref{fig:robustness_pi}, we show the training set cleansing performance of RE against the two attacks for $\pi\in\{0.82, 0.84, \ldots, 0.96, 0.98\}$. For both attacks, RE achieves high true positive rate (TPR) and low false positive rate (FPR)\footnote{TPR and FPR are defined in the main paper.} for a large range of $\pi$. For the attack with pattern A, there is a slight drop in TPR when $\pi$ is small (below 0.86), which matches with our previous discussion. Also, we observe a slight increase in FPR for both attacks when $\pi$ is too close to 1. However, this experiment still empirically validates a large range of reasonable choices for $\pi$ (e.g. from 0.86 to 0.96).

In Figure \ref{fig:robustness_lambda}, we show the training set cleansing performance of RE against the two attacks for $\lambda\in\{1/4, 1/3, 1/2, 1, 2, 3, 4\}$. For both attacks, the FPR increases as $\lambda$ decreases. For the attack with pattern C, the TPR decreases as $\lambda$ increases. Although for the attack with pattern A, the TPR stays high even for $\lambda=4$, we expect it to decrease as $\lambda$ further increases. Again, the observations match with our previous discussion -- $\lambda$ should be close to 1 (not exceeding the interval $[1/2, 2]$ if necessary) to ensure a high TPR and a low FPR.

\section{More on Related Works}

In the main paper, we reviewed three existing backdoor defenses deployed before/during the classifier's training phase; and a defense deployed post-training which inspired the current work. Here, we give a broader review of backdoor defenses.

\subsection{Defense Scenarios}

As introduced in the introduction section of the main paper, backdoor defense can be ``deployed either before/during training or post-training''. Defense developments for these two different scenarios are actually different problems, with the following differences:

\begin{itemize}
	\item In the before/during training scenario, the defender is usually the training authority who provide users with a trained classifier. In the post-training scenario, the defender is the user who receives/downloads the classifier from the training authority.
	\item In the before/during training scenario, the defender has access to the training set and has the resources and capability to train classifiers; but none of the training images are guaranteed to be clean (with no backdoor pattern embedded). In the post-training scenario, the defender has no access to the training set or resources to train any classifiers; but the defender may independently possess a (typically small) clean dataset.
	\item In the before/during training scenario, the defender aims to detect whether the training set is poisoned by backdoor training images. In the post-training scenario, the defender aims to detect whether the received classifier was trained on a poisoned training set.
	\item In the before/during training scenario, the defender aims to remove the backdoor training images from the training set, if there are any. In the post-training scenario, the defender discard the classifier if it is deemed to contain a backdoor.
\end{itemize}

Typical backdoor defenses deployed post-training include Neural Cleanse \cite{NC}, TABOR \cite{Tabor}, DeepInspect \cite{DeepInspect}, the anomaly detection method proposed by \cite{Post-TNNLS}, Fine-Pruning \cite{FP}, STRIP \cite{STRIP}, etc. All of these post-training defenses require an independent dataset that is guaranteed to be clean\footnote{Such a dataset can also be obtained through model inversion with GANS \cite{DeepInspect}.}.

In addition to SS, AC, and CI that we have reviewed in the main paper, there are other backdoor defenses deployed before/during training. The SCAn approach proposed by \cite{SCAn} improves the clustering approach of AC (k-means with $K=2$). For each putative target class, SCAn fits the internal layer activations with a single multivariate Gaussian model and a two-component mixture Gaussian mixture model. Then a likelihood ratio test is used to evaluate which model fits the activations better. However, SCAn requires not only the possibly poisoned training set, but also an independent clean dataset for estimating the distribution of the activations of the clean training images. Note that such an independent clean dataset is a strong assumption for the before/during training defense scenario, and is not assumed by any of SS, AC, CI, and our RE. SentiNet proposed by \cite{SentiNet} focuses on detecting perceptible backdoors. It examines the training images to locate the backdoor pattern which results in a prediction to the target class. Again, SentiNet requires an independent clean dataset. We recognize the contributions of these works. However, we find it is unfair to compare them with our work -- we only compare our RE with SS, AC, and CI, with no strong assumptions.

\subsection{Comparison with Reverse Engineering Approach for Post-Training Backdoor Defense}

As mentioned in the related work section of the main paper, our RE defense is inspired by a post-training defense which reverse engineers a pattern for each class pair using an independent clean data set \cite{Post-TNNLS}. Although this defense is a post-training defense, which is designed for a different scenario compared with our RE deployed before/during training, its objective function is a special case of RE's objective function for pattern estimation (Eq. (9) of the main paper) by letting $\lambda\rightarrow\infty$. Previously in the robustness analysis section of this supplementary material, we have discussed the effect of choosing an overly large $\lambda$ and tested RE with large $\lambda$'s experimentally. Here, we test the training set cleansing performance if the pattern estimation uses the objective function proposed by \cite{Post-TNNLS} (also an extreme case of RE where $\lambda\rightarrow\infty$).

\begin{figure}
	\centering
	\begin{minipage}[b]{.3\linewidth}
		\centering
		\centerline{\includegraphics[width=\linewidth]{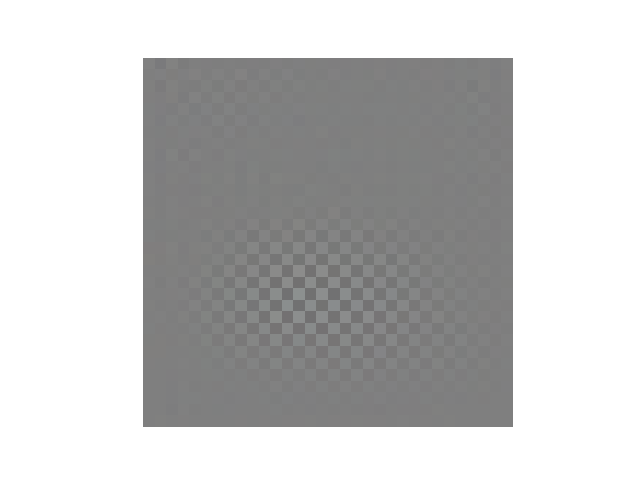}}
		\subcaption{pattern A}\label{subfig:be_est_AD_A}
	\end{minipage}
	\begin{minipage}[b]{.3\linewidth}
		\centering
		\centerline{\includegraphics[width=\linewidth]{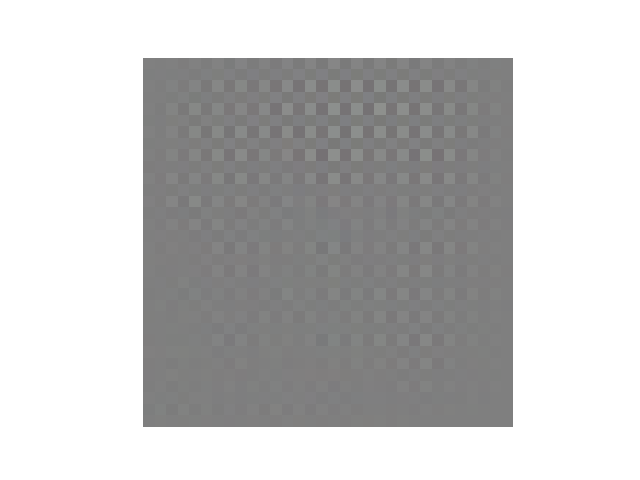}}
		\subcaption{pattern B}
	\end{minipage}
	\begin{minipage}[b]{.3\linewidth}
		\centering
		\centerline{\includegraphics[width=\linewidth]{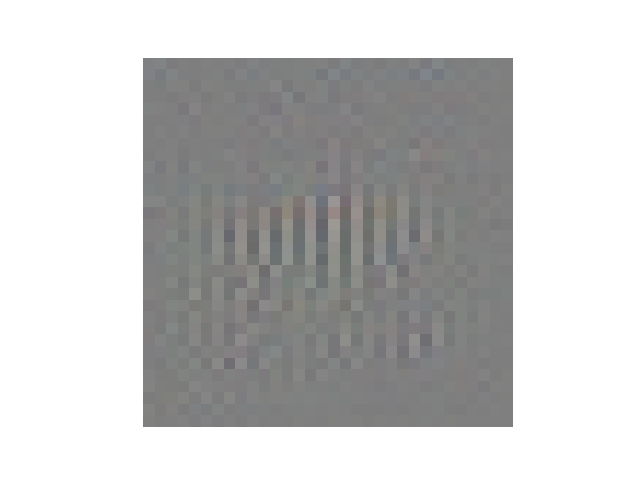}}
		\subcaption{pattern C}
	\end{minipage}
	\begin{minipage}[b]{.3\linewidth}
		\centering
		\centerline{\includegraphics[width=\linewidth]{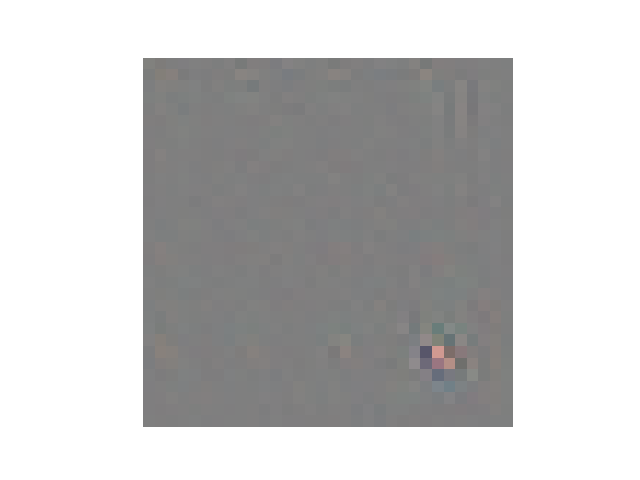}}
		\subcaption{pattern D}
	\end{minipage}
	\begin{minipage}[b]{.3\linewidth}
		\centering
		\centerline{\includegraphics[width=\linewidth]{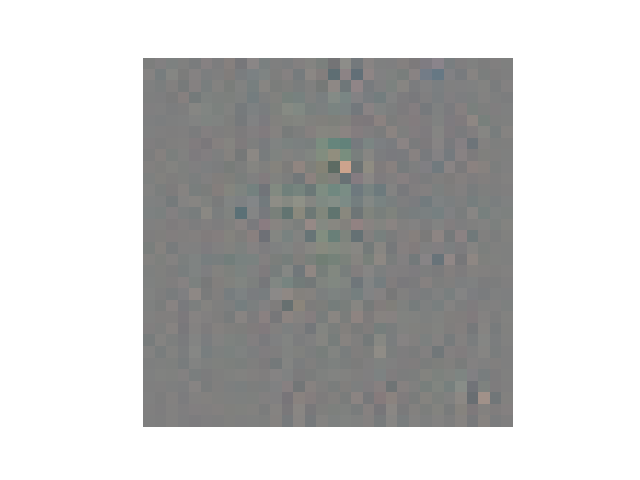}}
		\subcaption{pattern E}
	\end{minipage}
	\begin{minipage}[b]{.3\linewidth}
		\centering
		\centerline{\includegraphics[width=\linewidth]{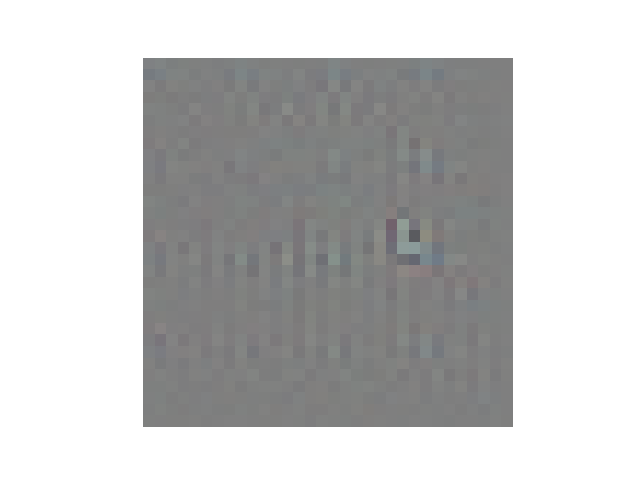}}
		\subcaption{pattern F}
	\end{minipage}
	\caption{Pattern estimated using the objective function proposed by \cite{Post-TNNLS} for the six 1SC attacks with backdoor pattern A-F.}
	\label{fig:AD_bd_pattern_est}
\end{figure}

For each of the six 1SC attacks in the main paper, we first estimate a pattern for the true backdoor class pair with the objective function proposed by \cite{Post-TNNLS}. The same configuration as for the baseline experiment in the main paper is used for pattern estimation here, with the wide ResNet architecture being used. The estimated patterns are shown in Figure \ref{fig:AD_bd_pattern_est}. Then training set cleansing is performed by manually ``removing'' the estimated pattern from all the training images labeled to the true backdoor target class, for each attack. The TPR and the FPR of training set cleansing for the six attacks are shown in Table \ref{tab:AD_cleansing}. Only for the attack with backdoor pattern F, the pattern estimated using the objective function of \cite{Post-TNNLS} achieves good performance. For the other attacks, not surprisingly, the TPR and FPR are both low. This can be explained by the estimated patterns shown in Figure \ref{fig:AD_bd_pattern_est}, as well. For the attack with pattern F, the ``l-shape'' and its location are both accurately estimated. Hence ``removing'' the ``l'' from the backdoor training images will likely cause misclassifications to the source class. However, for other patterns, although the estimated pattern can induce a high fraction of group misclassification from the source class to the target class for the backdoor class pair, the estimation is far from accurate. For example, for attacks with pattern A, only a patch of the ``chessboard'' is estimated; for attack with pattern E, the pixel being learned as the true backdoor pattern is successfully estimated, but the location is wrong.

\begin{table}[t]
	\begin{center}
		\caption{Training set cleansing true positive rate (TPR) and false positive rate (FPR) using the pattern estimated with the objective function of \cite{Post-TNNLS}}
		\resizebox{0.4\textwidth}{!}{
			\begin{tabular}{ |c|c|c|c|c|c|c| }
				\hline 
				& A & B & C & D & E & F\\
				\hline
				TPR & 0 & 63.0 & 3.8 & 43.8 & 58.4 & 96.0\\
				\hline
				FPR & 3.2 & 0.7 & 0.7 & 0.1 & 0.4 & 0.8\\
				\hline
			\end{tabular}\label{tab:AD_cleansing}}
	\end{center}
\end{table}

\section{More Insights and Discussions Regarding the Performance Evaluation Results}

In the main paper, we show that the proposed RE defense performs well in both attack detection and training set cleansing for a variety of attacks, regardless of the DNN architecture used by the defender. We compared our RE with existing backdoor defenses deployed before/during the classifier's training phase, including SS, AC, and CI. While these earlier defenses are shown to be effective in limited cases, they are generally outperformed by our RE defense in both attack detection and training set cleansing. Here, we dig deep to provide more insights related to the experiment results in the main paper -- we study when SS, AC, and CI will fail and why.

\subsection{Limitations of SS and AC}

\subsubsection{Limitation 1:} {\it Internal layer activations of backdoor training images are not always separable from internal layer activations of clean training images labeled to the target class when projected onto low-dimensional feature spaces.}

In the related work section of the main paper, we introduced the key ideas of SS and AC. Both methods first train a classifier using the possibly poisoned training set. Then for each putative target class, the training images are fed into the trained classifier, and the internal layer (e.g. the penultimate layer) activation for each image is extracted and projected onto a low-dimensional (1-dimensional for SS and 10-dimensional for AC) space. If there is an attack, for the true target class, the internal layer activations of the backdoor patterns are expected to be separable from the internal layer activations of the clean images on the low-dimensional feature space. But will this separation always happen?

\begin{figure*}
	\centering
	\begin{minipage}[b]{.3\linewidth}
		\centering
		\includegraphics[width=\linewidth]{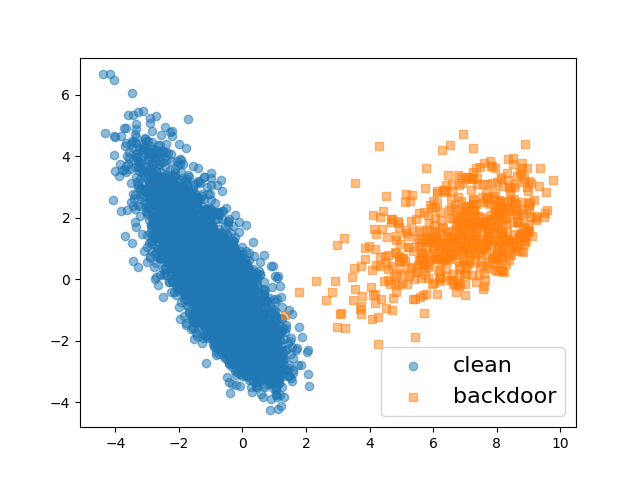}
		\subcaption{pattern A, ResNet, wide}
	\end{minipage}
	\begin{minipage}[b]{.3\linewidth}
		\centering
		\includegraphics[width=\linewidth]{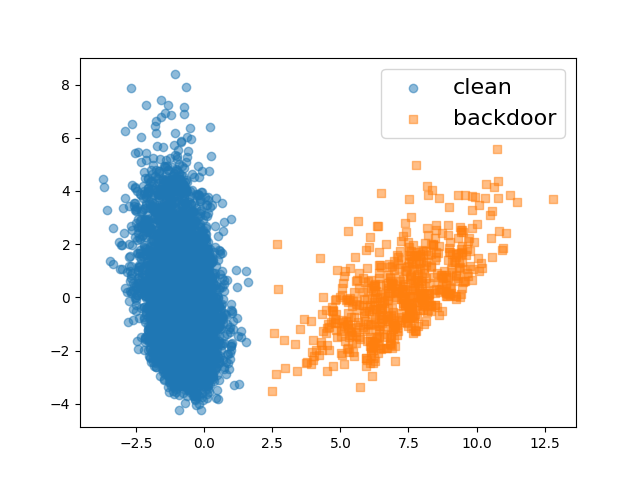}
		\subcaption{pattern B, ResNet, wide}
	\end{minipage}
	\begin{minipage}[b]{.3\linewidth}
		\centering
		\includegraphics[width=\linewidth]{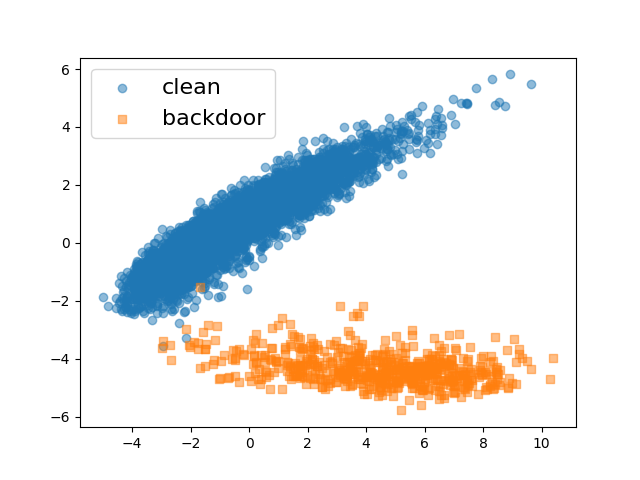}
		\subcaption{pattern C, ResNet, wide}
	\end{minipage}
	\begin{minipage}[b]{.3\linewidth}
		\centering
		\includegraphics[width=\linewidth]{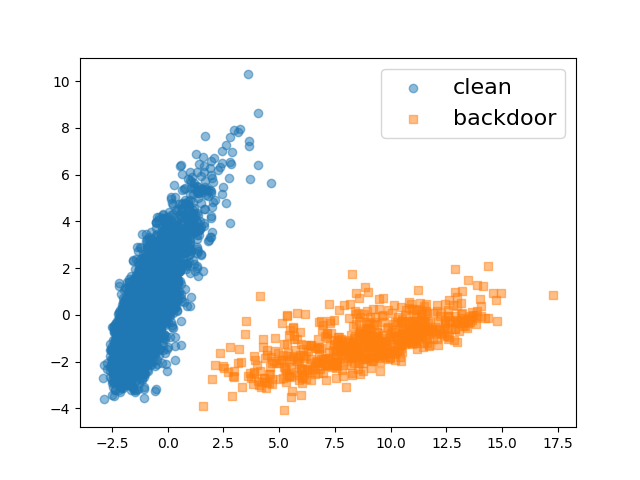}
		\subcaption{pattern D, ResNet, wide}
	\end{minipage}
	\begin{minipage}[b]{.3\linewidth}
		\centering
		\includegraphics[width=\linewidth]{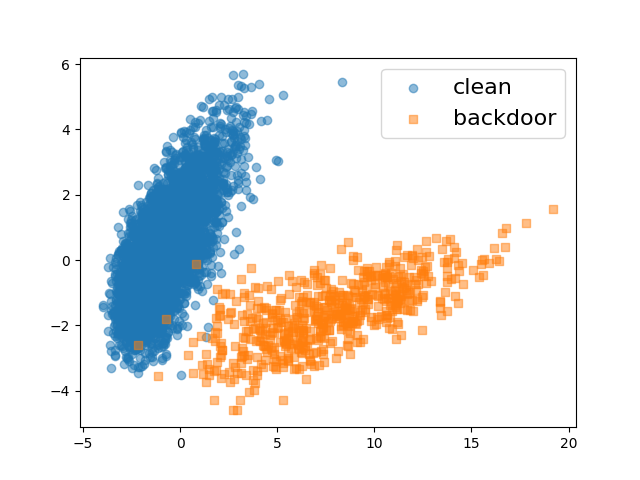}
		\subcaption{pattern E, ResNet, wide}
	\end{minipage}
	\begin{minipage}[b]{.3\linewidth}
		\centering
		\includegraphics[width=\linewidth]{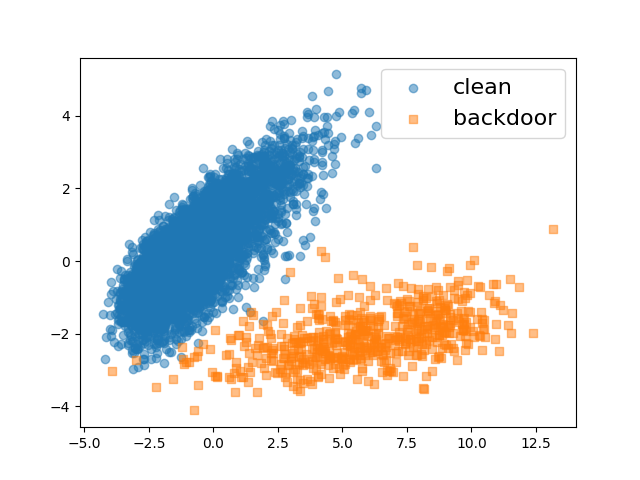}
		\subcaption{pattern F, ResNet, wide}
	\end{minipage}
	\caption{Distribution of the internal layer activations of the backdoor training images (orange squares) and the internal layer activations of the clean training images labeled to the true target class (blue circles) in 2D, for the 3SC attacks for pattern A-F, when the wide ResNet architecture is used by the defender.}
	\label{fig:nonseparable_wide}
\end{figure*}

\begin{figure*}
	\centering
	\begin{minipage}[b]{.3\linewidth}
		\centering
		\includegraphics[width=\linewidth]{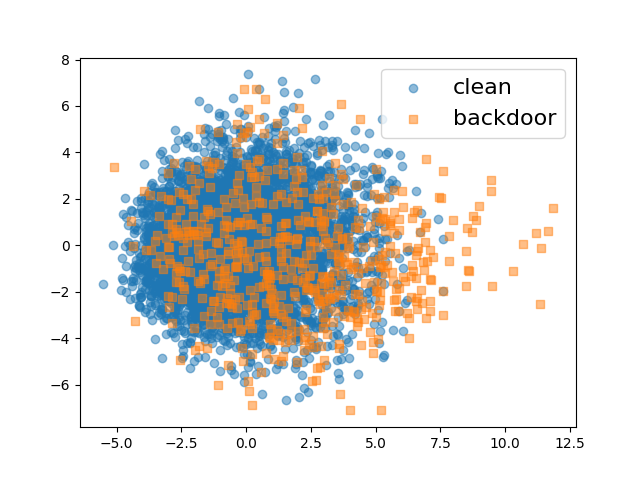}
		\subcaption{pattern A, ResNet, compact}
	\end{minipage}
	\begin{minipage}[b]{.3\linewidth}
		\centering
		\includegraphics[width=\linewidth]{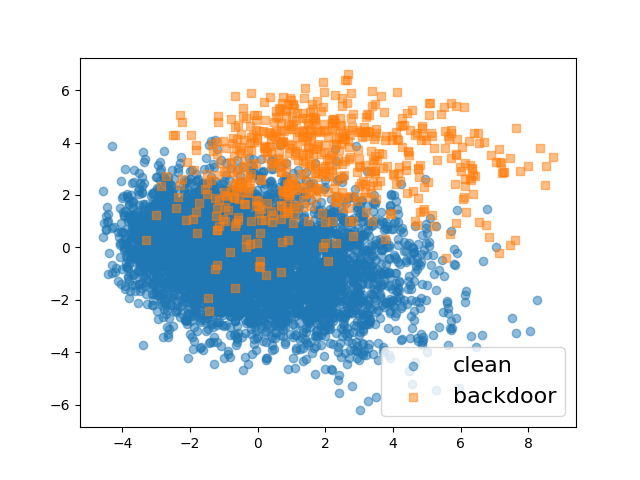}
		\subcaption{pattern B, ResNet, compact}
	\end{minipage}
	\begin{minipage}[b]{.3\linewidth}
		\centering
		\includegraphics[width=\linewidth]{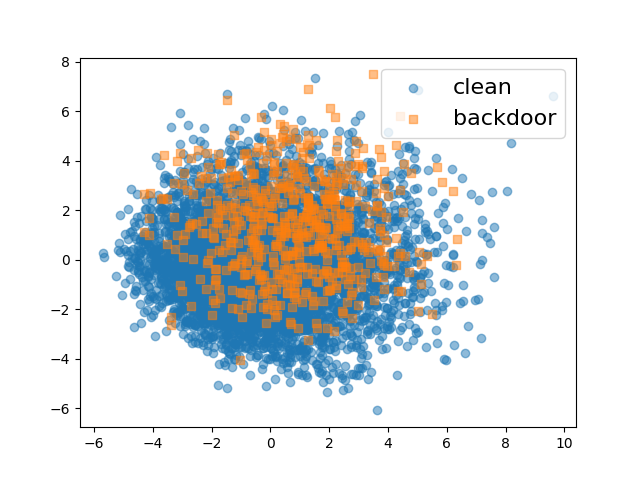}
		\subcaption{pattern C, ResNet, compact}
	\end{minipage}
	\begin{minipage}[b]{.3\linewidth}
		\centering
		\includegraphics[width=\linewidth]{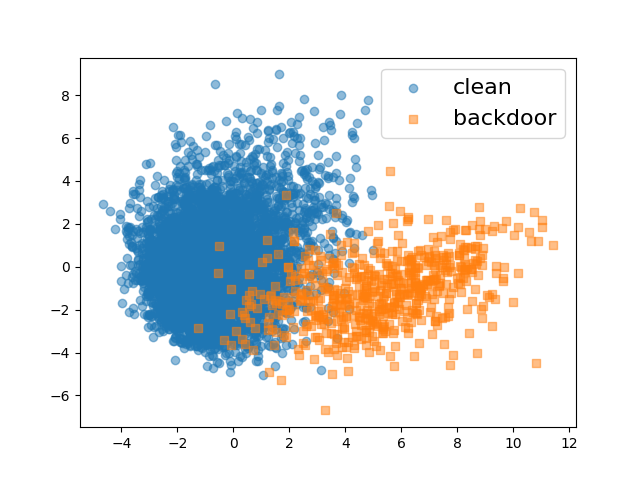}
		\subcaption{pattern D, ResNet, compact}\label{subfig:exception}
	\end{minipage}
	\begin{minipage}[b]{.3\linewidth}
		\centering
		\includegraphics[width=\linewidth]{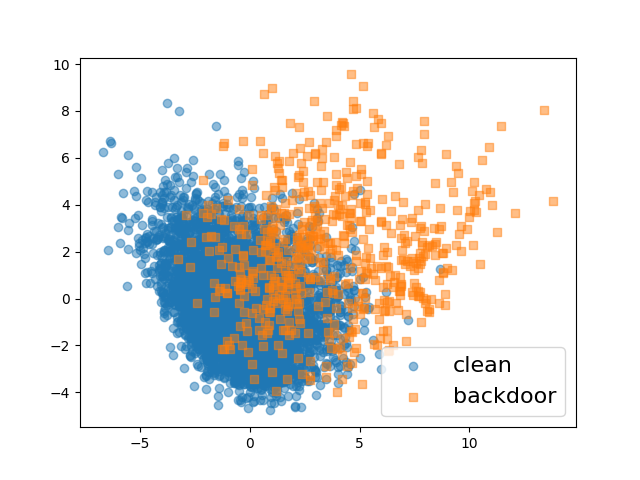}
		\subcaption{pattern E, ResNet, compact}
	\end{minipage}
	\begin{minipage}[b]{.3\linewidth}
		\centering
		\includegraphics[width=\linewidth]{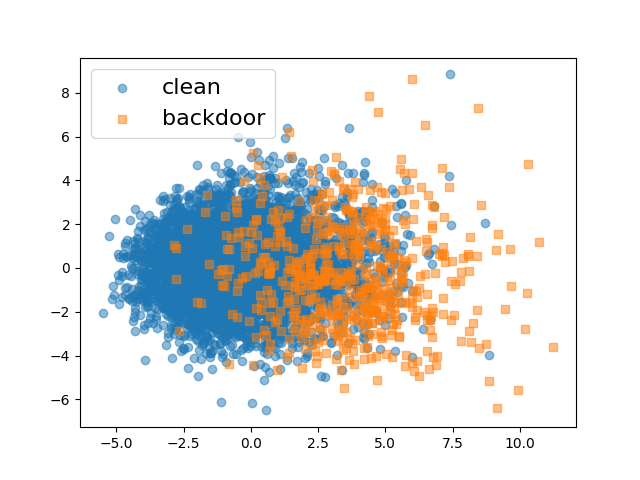}
		\subcaption{pattern F, ResNet, compact}
	\end{minipage}
	\caption{Distribution of the internal layer activations of the backdoor training images (orange squares) and the internal layer activations of the clean training images labeled to the true target class (blue circles) in 2D, for the 3SC attacks for pattern A-F, when the compact ResNet architecture is used by the defender.}
	\label{fig:nonseparable_compact}
\end{figure*}

\begin{figure*}
	\centering
	\begin{minipage}[b]{.3\linewidth}
		\centering
		\includegraphics[width=\linewidth]{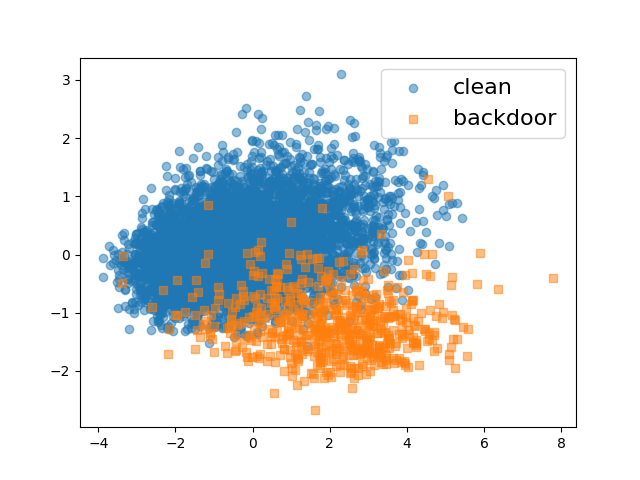}
		\subcaption{pattern A, MobileNetV2}
	\end{minipage}
	\begin{minipage}[b]{.3\linewidth}
		\centering
		\includegraphics[width=\linewidth]{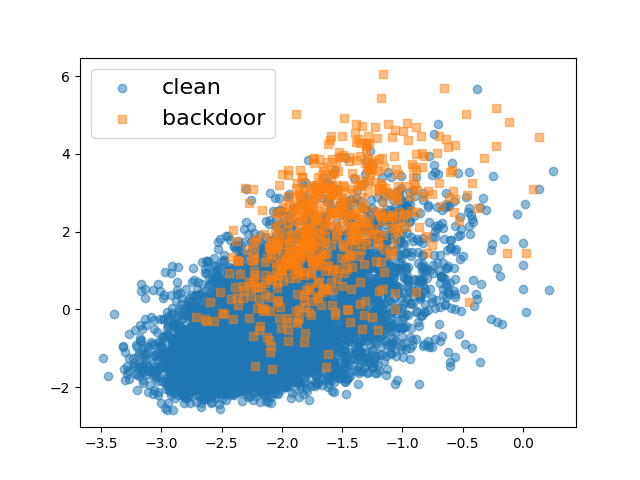}
		\subcaption{pattern B, MobileNetV2}
	\end{minipage}
	\begin{minipage}[b]{.3\linewidth}
		\centering
		\includegraphics[width=\linewidth]{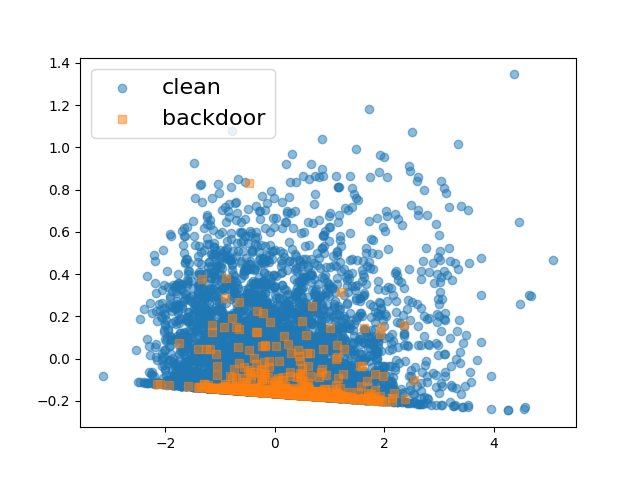}
		\subcaption{pattern C, MobileNetV2}
	\end{minipage}
	\begin{minipage}[b]{.3\linewidth}
		\centering
		\includegraphics[width=\linewidth]{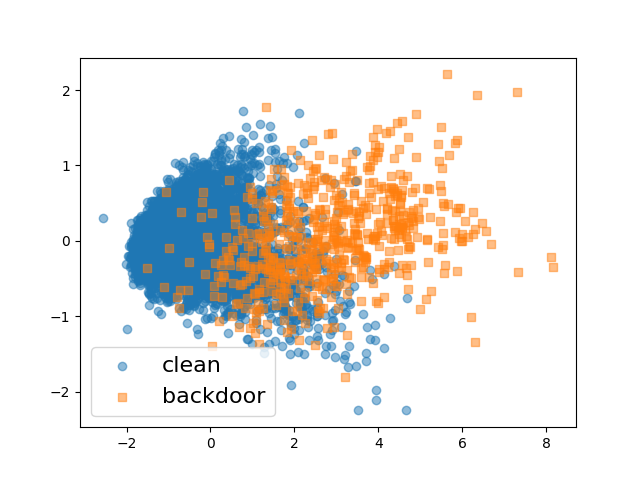}
		\subcaption{pattern D, MobileNetV2}
	\end{minipage}
	\begin{minipage}[b]{.3\linewidth}
		\centering
		\includegraphics[width=\linewidth]{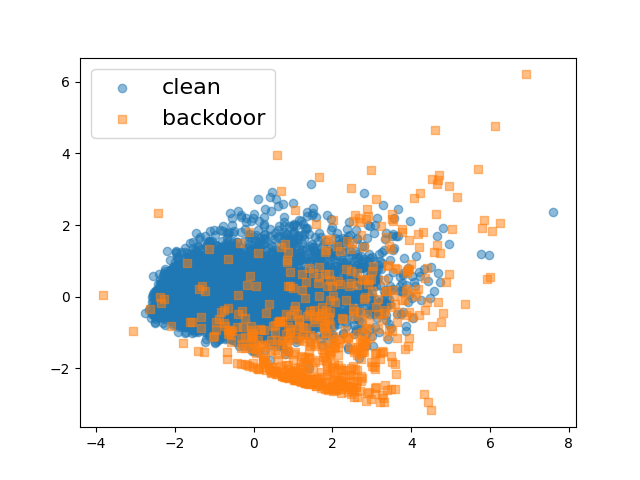}
		\subcaption{pattern E, MobileNetV2}
	\end{minipage}
	\begin{minipage}[b]{.3\linewidth}
		\centering
		\includegraphics[width=\linewidth]{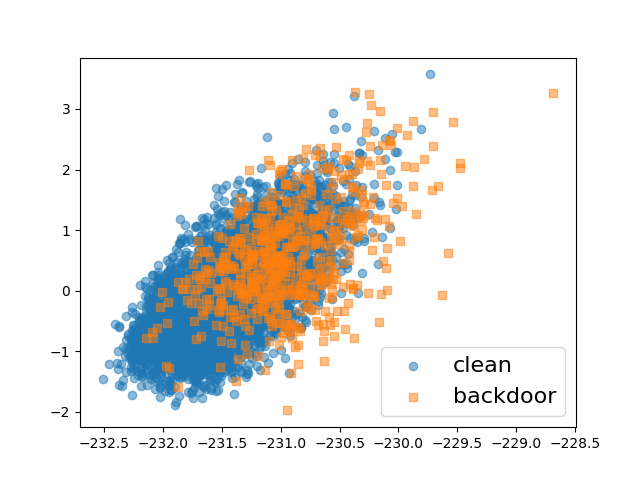}
		\subcaption{pattern F, MobileNetV2}
	\end{minipage}
	\caption{Distribution of the internal layer activations of the backdoor training images (orange squares) and the internal layer activations of the clean training images labeled to the true target class (blue circles) in 2D, for the 3SC attacks for pattern A-F, when the MobileNetV2 architecture is used by the defender.}
	\label{fig:nonseparable_MobileNet}
\end{figure*}

Here, we visualize the distribution of the penultimate layer activations of the backdoor training images and the penultimate layer activations of the clean training images labeled to the true target class in 2D (i.e. the first two principal components), for the 3-source-class (3SC) attacks for the six backdoor patterns (i.e. pattern A-F of the main paper), for the wide ResNet architecture (in Figure \ref{fig:nonseparable_wide}), the compact ResNet architecture (in Figure \ref{fig:nonseparable_compact}), and the MobileNetV2 architecture (in Figure \ref{fig:nonseparable_MobileNet}). Clear separation between the activations of the backdoor training images and the clean training images can be observed (in 2D) only if the wide ResNet architecture is used by the defender. In this case, SS and AC are relatively effective in training set cleansing, as shown in Table 3(a) of the main paper. However, if the compact ResNet architecture or the MobileNet architecture is used by the defender, the penultimate layer activations of the backdoor training images are not clearly separable from the penultimate layer activations of the clean training images in 2D for most of the attacks (except for, e.g. pattern D for the compact ResNet architecture, see Figure \ref{subfig:exception}). Correspondingly, SS works poorly in most of these cases (except for, e.g. pattern D for the compact ResNet architecture, see Table 3(b) of the main paper). For AC, although non-separability in 2D is not equivalent to non-separability in higher dimensional spaces (e.g. the 10-dimensional space considered by AC \cite{AC}), we can still infer from the high false positive rates (FPRs) of AC in Table 3(b) of the main paper that the activations of the backdoor training images fail to form an independent cluster (using k-means with $k=2$) for most of the attacks, if the compact DNN architecture is used by the defender.

\subsubsection{Limitation 2:} {\it Detection approaches proposed by AC are not practical/reliable for general backdoor attacks.}

As detailed in the related work section of the main paper, for each putative target class, AC applies k-means clustering with $k=2$ to the penultimate layer activations (projected onto a low-dimensional space) of all training images labeled to this class. As the first backdoor defense deployed before/during the training phase that addresses backdoor detection, three detection approaches are proposed by AC \cite{AC}:

1) For each putative target class, two classifiers are trained, with each classifier trained using the entire training set with the training images corresponding to one of the two clusters (obtained by k-means with $k=2$ for this class) being excluded. Then for each classifier, the excluded training images are fed into the trained classifier. If the training images being excluded contain mainly the backdoor training images, and the training images remaining contain few backdoor training images, the backdoor pattern will not be learned, and the excluded training images will likely be classified to source class(es) -- different from the target class they are labeled to. However, this method requires the backdoor training images forming an independent cluster with high purity, which is not true in many cases as we have just shown. Moreover, this method requires training $2K$ ($K$ the number of classes) classifiers, which is too time-consuming in practice.

2) An attack is claimed to be detected if, for any class, one of the two components obtained by 2-means has a mass larger than a threshold. Again, this method relies on the activations of the backdoor training images forming an independent cluster. Moreover, this method requires that the proportion of the backdoor training images embedded in the true target class be small. Based on the results reported by AC \cite{AC}, for MNIST, such proportion cannot exceed 30\% (see Table 3 of the AC paper). Although a higher poisoning rate can be more noticeable, an attacker can choose to defeat this detection method by using more backdoor training images. Also, this detection method is supervised (while our RE is not).

3) In the main paper, we introduced the third detection method proposed by AC. This method evaluates the Silhouette score for the 2-means clustering results for each putative target class. If the Silhouette score for any class is larger than a threshold, an attack is claimed to be detected, with the class corresponding to the largest Silhouette score inferred as the target class. Since this method does not make any strong assumptions about the attack (e.g. an upper bound on the poisoning rate for method 2) and is practically feasible to test with (not like method 1 with very high complexity), we use this method as the detection method for AC in our main paper.

However, the detection approach based on the Silhouette score is very sensitive to the dataset and the DNN architecture. As reported by \cite{AC}, the maximum Silhouette score for classifiers trained on the clean MNIST dataset is 0.11, while the maximum detection threshold to detect 75\% attacks is 0.15. However, for CIFAR-10, the maximum Silhouette scores for the (six) classifiers trained on the clean training set are 0.52, 0.29, 0.23, and 0.24 (all greater than 0.15), for the wide ResNet, the compact ResNet, MobileNetV2, and DensNet architectures, respectively. Although AC achieves relatively good performance in training set cleansing if the true target class is assumed to be known (due to the clear separation between backdoor training images and clean training images in terms of internal layer activation) when the wide ResNet architecture is used, more than half of the attacks are not detected with the AC's detection approach based on the Silhouette score. Since for the wide ResNet architecture, the maximum Silhouette scores (over the ten classes) for each of the six classifiers trained on the clean training set are 0.42, 0.51, 0.36, 0.51, 0.35, and 0.37, respectively, using a smaller threshold may help AC detect more attacks, but will also cause more false detections.

\subsubsection{Limitation 3:} {\it AC's clustering approach, k-means with $K=2$, may fail due to the ``multi-modality'' of specific classes.}

In fact, the multi-modality issue has been discussed in \cite{AC}. The authors combined several classes of MNIST into a ``super-class'' and set it as the target class of an attack. In this case, k-means with $K=2$ successfully separated the activations of backdoor training images from the activations of the clean training images in low-dimensional spaces.

However, we find that the ``multi-modality'' of the target class becomes an issue for k-means clustering with $K=2$ if the activations of the backdoor training images are not far apart from the activations of the clean training images labeled to the target class (see Figure \ref{fig:nonseparable_compact} and Figure \ref{fig:nonseparable_MobileNet}). In such a case, since the number of backdoor training images is usually much smaller than the number of clean training images, k-means with $K=2$ will likely form two large clusters based on the ``multi-modality'' of the target class, without considering the distribution of the activations of the backdoor training images. This explains why the FPR is high (even close to 50\%) for AC against many attacks (see Table 3(b) of the main paper). 

\subsection{Limitations of CI}

CI realized limitations of AC regarding its naive clustering approach (i.e. k-means with $K=2$). Hence it proposes a more sophisticated clustering approach based on Gaussian mixture modeling. For each putative target class, the internal layer activation is modeled as several Gaussian components with full covariance matrix, with the number of components selected by Bayesian information criterion (BIC). By allowing more clusters/components, and also exploring the full covariance information in high dimensional spaces, CI can separate the activations of backdoor training images into one or several independent clusters with high purity.

However, the main limitation of CI is that it requires the internal layer activations to have relatively small dimension. Otherwise, BIC will always choose one component even for the true target class, since the number of parameters of the Gaussian mixture model is much larger than the number of training images labeled to the target class. This explains why CI achieves relatively good performance (though not as good as our RE defense) in both detection and training set cleansing when the compact ResNet architecture is used, but fails when the wide ResNet architecture is used.

\section{Example Backdoor Training Images with the Estimated Backdoor Pattern Removed}

\begin{figure}
	\centering
	\begin{minipage}[b]{\linewidth}
		\centering
		\includegraphics[width=0.42\linewidth]{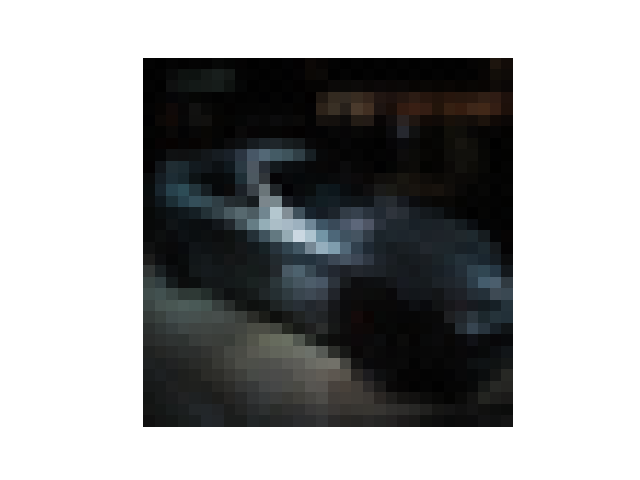}
		\includegraphics[width=0.42\linewidth]{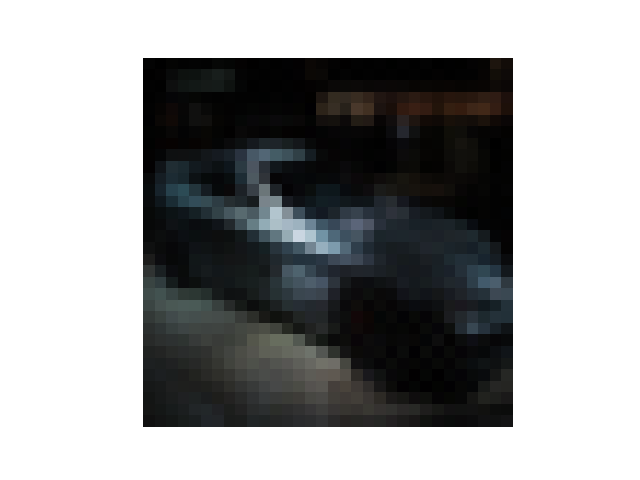}
		\subcaption{pattern A}
	\end{minipage}
	\begin{minipage}[b]{\linewidth}
		\centering
		\includegraphics[width=0.45\linewidth]{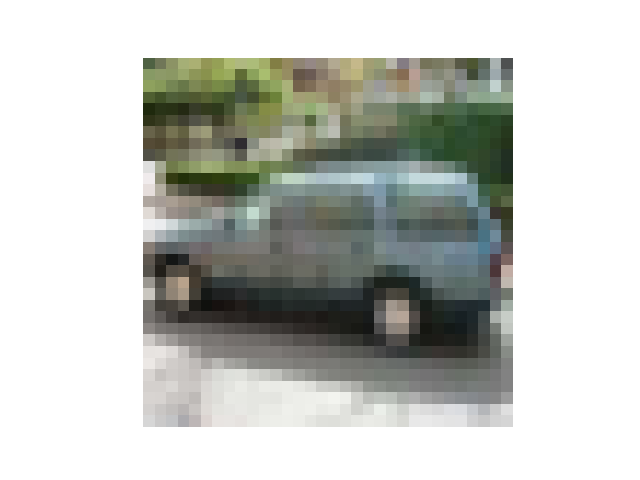}
		\includegraphics[width=0.45\linewidth]{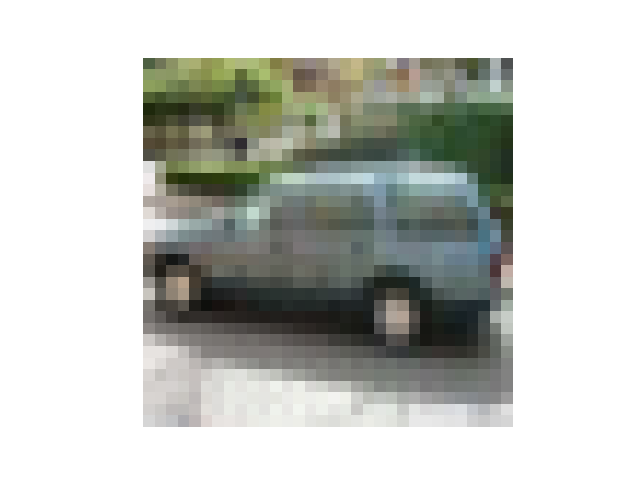}
		\subcaption{pattern B}
	\end{minipage}
	\begin{minipage}[b]{\linewidth}
		\centering
		\includegraphics[width=0.45\linewidth]{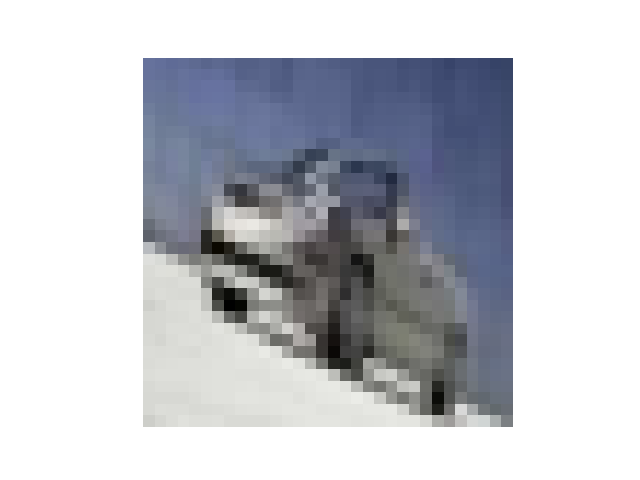}
		\includegraphics[width=0.45\linewidth]{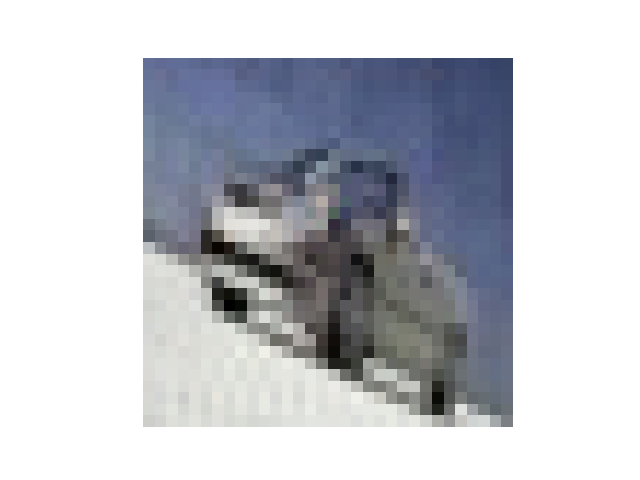}
		\subcaption{pattern C}
	\end{minipage}
	\begin{minipage}[b]{\linewidth}
		\centering
		\includegraphics[width=0.45\linewidth]{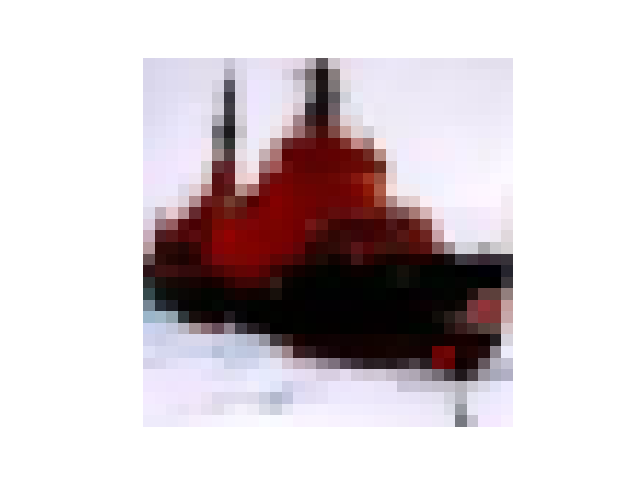}
		\includegraphics[width=0.45\linewidth]{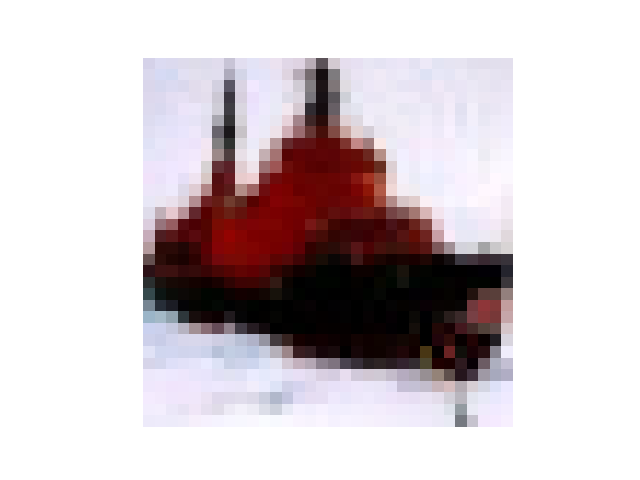}
		\subcaption{pattern D}
	\end{minipage}
	\begin{minipage}[b]{\linewidth}
		\centering
		\includegraphics[width=0.45\linewidth]{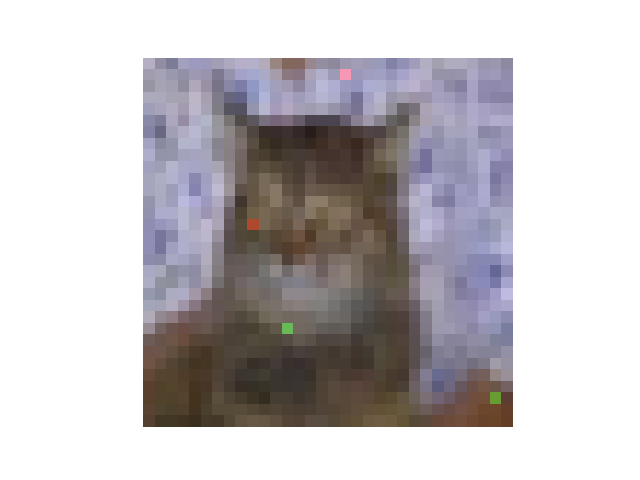}
		\includegraphics[width=0.45\linewidth]{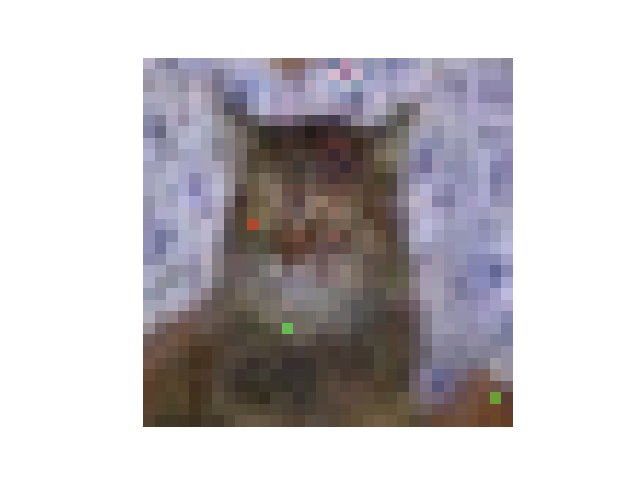}
		\subcaption{pattern E}
	\end{minipage}
	\begin{minipage}[b]{\linewidth}
		\centering
		\includegraphics[width=0.45\linewidth]{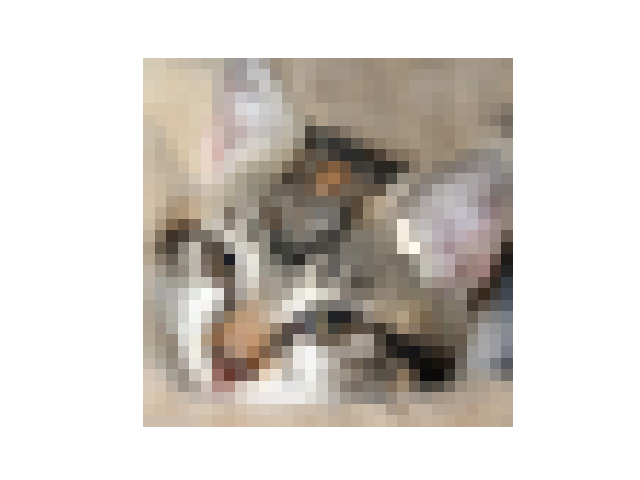}
		\includegraphics[width=0.45\linewidth]{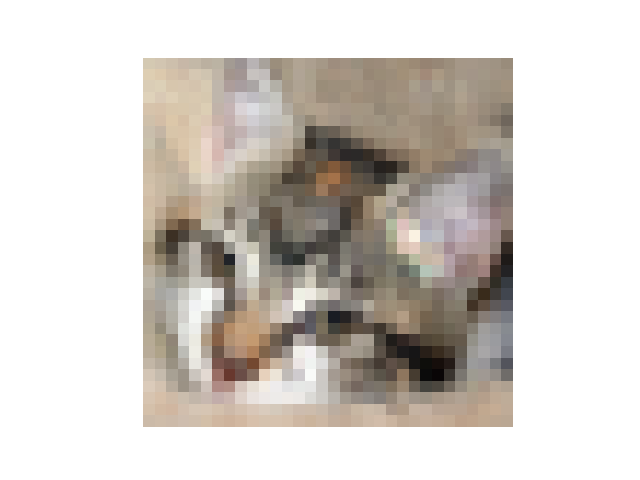}
		\subcaption{pattern F}
	\end{minipage}
	\caption{Example backdoor training images with the backdoor pattern embedded (left) and the backdoor training images with the estimated backdoor pattern removed (right), for 1SC attacks for pattern A-F.}
	\label{fig:bd_images_cleaned}
\end{figure}

In the main paper, we showed that the estimated backdoor patterns are highly similar to the true backdoor patterns used in the attacks. For pattern E, although we did not recover all the four pixels being perturbed, we realized that it is the topmost pixel being learned by the classifier as the effective backdoor pattern\footnote{By embedding this estimated pattern into the clean test images from the source class of the 1SC attack for pattern E, we achieve 99.0\% ASR on defenseless DNN with the wide ResNet architecture.}. In Figure \ref{fig:bd_images_cleaned}, we show, for each 1SC attack with pattern A-F, an example of backdoor training images with the estimated backdoor pattern removed.

\begin{table}[t]
	\begin{center}
		\caption{Attack success rate (ASR) and clean test accuracy (ACC), jointly represented by ASR/ACC, of the retrained classifiers for the 18 attacks when SS is applied (with the assumption that the existence of attack and the true target class are known to the defender), for both wide and compact DNN architectures.}
		\resizebox{0.48\textwidth}{!}{
			\begin{tabular}{ |c|c|c|c|c|c|c|c| }
				\hline 
				\multicolumn{2}{|c|}{pattern} & A & B & C & D & E & F\\
				\hline
				\multirow{3}{*}{\thead{wide\\DNN}}
				& \multicolumn{1}{|c|}{1SC} & \multicolumn{1}{|c|}{1.7/91.7} & \multicolumn{1}{|c|}{0/92.7} & \multicolumn{1}{|c|}{98.3/91.8} & \multicolumn{1}{|c|}{0.4/92.6} & \multicolumn{1}{|c|}{88.1/91.9} & \multicolumn{1}{|c|}{73.6/92.0}\\
				\cline{2-8}
				& \multicolumn{1}{|c|}{3SC} & \multicolumn{1}{|c|}{0.8/91.8} & \multicolumn{1}{|c|}{0.1/91.8} & \multicolumn{1}{|c|}{99.2/92.2} & \multicolumn{1}{|c|}{0.4/91.6} & \multicolumn{1}{|c|}{1.2/91.9} & \multicolumn{1}{|c|}{53.3/92.6}\\
				\cline{2-8}
				& \multicolumn{1}{|c|}{9SC} & \multicolumn{1}{|c|}{0.4/92.4} & \multicolumn{1}{|c|}{0.3/91.4} & \multicolumn{1}{|c|}{10.6/91.8} & \multicolumn{1}{|c|}{0.7/92.4} & \multicolumn{1}{|c|}{0.6/92.3} & \multicolumn{1}{|c|}{0.6/92.6}\\
				\hline
				\multirow{3}{*}{\thead{com-\\pact\\DNN}}
				& \multicolumn{1}{|c|}{1SC} & \multicolumn{1}{|c|}{94.8/90.3} & \multicolumn{1}{|c|}{0.2/90.3} & \multicolumn{1}{|c|}{97.6/90.5} & \multicolumn{1}{|c|}{15.0/90.4} & \multicolumn{1}{|c|}{93.4/90.1} & \multicolumn{1}{|c|}{79.5/89.9}\\
				\cline{2-8}
				& \multicolumn{1}{|c|}{3SC} & \multicolumn{1}{|c|}{98.3/90.2} & \multicolumn{1}{|c|}{96.9/90.5} & \multicolumn{1}{|c|}{97.4/90.8} & \multicolumn{1}{|c|}{1.3/90.2} & \multicolumn{1}{|c|}{79.9/90.2} & \multicolumn{1}{|c|}{86.4/90.4}\\
				\cline{2-8}
				& \multicolumn{1}{|c|}{9SC} & \multicolumn{1}{|c|}{98.9/90.4} & \multicolumn{1}{|c|}{98.2/90.1} & \multicolumn{1}{|c|}{96.7/89.8} & \multicolumn{1}{|c|}{3.1/89.6} & \multicolumn{1}{|c|}{13.6/88.9} & \multicolumn{1}{|c|}{42.4/90.1}\\
				\hline
			\end{tabular}\label{tab:ASR_ACC_retrain_SS}}
	\end{center}
\end{table}

\begin{table}[t]
	\begin{center}
		\caption{Attack success rate (ASR) and clean test accuracy (ACC), jointly represented by ASR/ACC, of the retrained classifiers for the 18 attacks when AC is applied (with the assumption that the existence of attack and the true target class are known to the defender), for both wide and compact DNN architectures.}
		\resizebox{0.48\textwidth}{!}{
			\begin{tabular}{ |c|c|c|c|c|c|c|c| }
				\hline 
				\multicolumn{2}{|c|}{pattern} & A & B & C & D & E & F\\
				\hline
				\multirow{3}{*}{\thead{wide\\DNN}}
				& \multicolumn{1}{|c|}{1SC} & \multicolumn{1}{|c|}{26.7/92.0} & \multicolumn{1}{|c|}{0/91.8} & \multicolumn{1}{|c|}{92.7/92.6} & \multicolumn{1}{|c|}{0.3/91.9} & \multicolumn{1}{|c|}{5.5/91.7} & \multicolumn{1}{|c|}{7.0/92.0}\\
				\cline{2-8}
				& \multicolumn{1}{|c|}{3SC} & \multicolumn{1}{|c|}{0.6/92.6} & \multicolumn{1}{|c|}{0.2/91.7} & \multicolumn{1}{|c|}{93.5/91.8} & \multicolumn{1}{|c|}{0.7/91.7} & \multicolumn{1}{|c|}{2.2/92.6} & \multicolumn{1}{|c|}{10.2/91.9}\\
				\cline{2-8}
				& \multicolumn{1}{|c|}{9SC} & \multicolumn{1}{|c|}{0.3/92.1} & \multicolumn{1}{|c|}{0.3/92.1} & \multicolumn{1}{|c|}{55.4/92.7} & \multicolumn{1}{|c|}{1.4/92.4} & \multicolumn{1}{|c|}{0.8/91.6} & \multicolumn{1}{|c|}{4.0/92.5}\\
				\hline
				\multirow{3}{*}{\thead{com-\\pact\\DNN}}
				& \multicolumn{1}{|c|}{1SC} & \multicolumn{1}{|c|}{99.3/90.3} & \multicolumn{1}{|c|}{0.3/90.4} & \multicolumn{1}{|c|}{80.6/90.3} & \multicolumn{1}{|c|}{0.4/90.2} & \multicolumn{1}{|c|}{32.5/90.1} & \multicolumn{1}{|c|}{0.7/89.9}\\
				\cline{2-8}
				& \multicolumn{1}{|c|}{3SC} & \multicolumn{1}{|c|}{91.1/90.6} & \multicolumn{1}{|c|}{73.8/89.9} & \multicolumn{1}{|c|}{97.2/90.0} & \multicolumn{1}{|c|}{1.1/90.4} & \multicolumn{1}{|c|}{2.6/90.1} & \multicolumn{1}{|c|}{1.1/90.4}\\
				\cline{2-8}
				& \multicolumn{1}{|c|}{9SC} & \multicolumn{1}{|c|}{98.2/90.9} & \multicolumn{1}{|c|}{45.7/90.1} & \multicolumn{1}{|c|}{94.1/90.0} & \multicolumn{1}{|c|}{2.8/89.7} & \multicolumn{1}{|c|}{0.8/90.0} & \multicolumn{1}{|c|}{1.4/90.4}\\
				\hline
			\end{tabular}\label{tab:ASR_ACC_retrain_AC}}
	\end{center}
\end{table}

\begin{table}[t]
	\begin{center}
		\caption{Attack success rate (ASR) and clean test accuracy (ACC), jointly represented by ASR/ACC, of the retrained classifiers for the 18 attacks when CI is applied with the compact DNN architectures.}
		\resizebox{0.48\textwidth}{!}{
			\begin{tabular}{ |c|c|c|c|c|c|c| }
				\hline 
				& A & B & C & D & E & F\\
				\hline
				1SC & 98.5/90.6 & 0.1/90.9 & 3.2/90.5 & n.a/n.a & 3.0/89.9 & 0.4/90.0\\
				\hline
				3SC & 2.2/90.6 & 0.5/90.1 & 0.3/90.2 & 0.9/90.1 & 0.9/89.3 & 0.5/90.0\\
				\hline
				9SC & 0.6/90.3 & 0.8/90.3 & 0.8/89.9 & 2.3/90.5 & 0.6/89.6 & 0.9/90.1\\
				\hline
			\end{tabular}\label{tab:ASR_ACC_retrain_CI}}
	\end{center}
\end{table}

\section{Retraining Results of SS, AC, and CI for the Baseline Experiment}

Supposing that there are suspicious training images being identified and removed by a backdoor defense deployed before/during the classifier's training phase, a retraining will be performed on the sanitized training set. In the main paper, we show that with the protection of RE, the effective ASR of imperceptible backdoor attacks on CIFAR-10 can be reduced to as low as 4.9\%, with negligible degradation in the clean test ACC of the retrained classifier. In Table \ref{tab:ASR_ACC_retrain_SS}, Table \ref{tab:ASR_ACC_retrain_AC}, and table \ref{tab:ASR_ACC_retrain_CI}, we show the ASR and clean test ACC of the retrained classifiers for the 18 attacks for SS, AC, and CI, respectively. Again, results for SS and AC are based on the assumption that the existence of the backdoor attack and the true target class are known to the defender. In practice, if an attack escapes detection, there will be no training set cleansing or retraining. Even with our ``assistance'', SS and AC fail to defeat a significant number of attacks respectively -- the ASR of the retrained classifiers after training set cleansing for these attacks are as high as if there is no defense. For CI, we only show the retraining results when there are training images being removed -- there is no retraining if the wide ResNet architecture is used by the defender. Still, CI cannot defeat all the 18 attacks even if the wide ResNet architecture is in use. Clearly, our RE outperforms SS, AC, and CI.

The ASC and the clean test ACC of the retrained classifiers generally match the TPR and FPR of training set cleansing reported in Table 3 of the main paper. Usually, when the TPR of training set cleansing is high, it is likely that the ASC of the retrained classifier is low. While this relation is not always true, especially for TPR not sufficiently large, our RE achieves very large TPR ($\leq90\%$) in training set cleansing against almost all the attacks. With only tens (or even a few)  backdoor training images remaining, it is not surprising that the backdoor attack will fail (as shown in Table 4 of the main paper).

\section{Performance Evaluation of RE against Attacks with Various Strengths}

It has been shown by \cite{Post-TNNLS} that an imperceptible backdoor pattern can be better learned if the perturbation size increases. However, the attack may be more easily noticeable to humans or detected by simple data examination techniques. Although the ASRs of the attacks considered in the main paper are close to 100\%, which means that further increasing the perturbation size will only harm the evasiveness of the attacks, we evaluate the performance of RE against imperceptible backdoors with large perturbation size. Note that both attack detection and training set cleansing of RE relies on searching a pattern with small norm for any backdoor class pair $(s, t^{\ast})$ with $s\in{\mathcal S}^{\ast}$, such that the estimated pattern: 1) induces high misclassification from $s$ to $t^{\ast}$, and 2) changes the class decision of the backdoor training images to classes other than $t^{\ast}$ when it is ``removed''. Hence, we expect that RE will fail when the backdoor pattern is perceptible (i.e. having an overly large size/norm).

\begin{figure}[t]
	\centering
	\begin{minipage}[b]{.45\linewidth}
		\centering
		\centerline{\includegraphics[width=\linewidth]{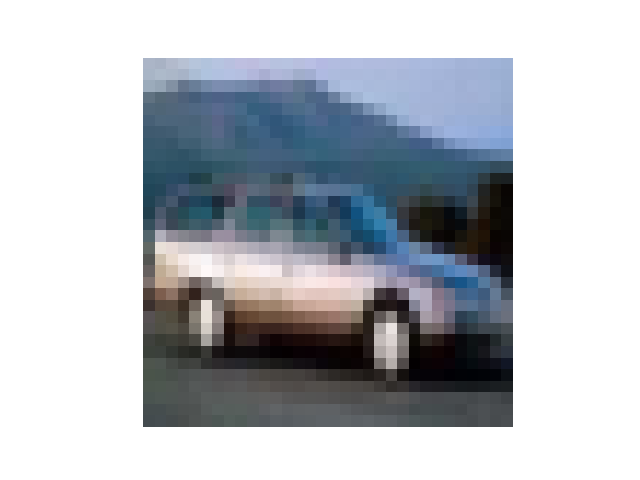}}
		\subcaption{1/255}
	\end{minipage}
	\begin{minipage}[b]{.45\linewidth}
		\centering
		\centerline{\includegraphics[width=\linewidth]{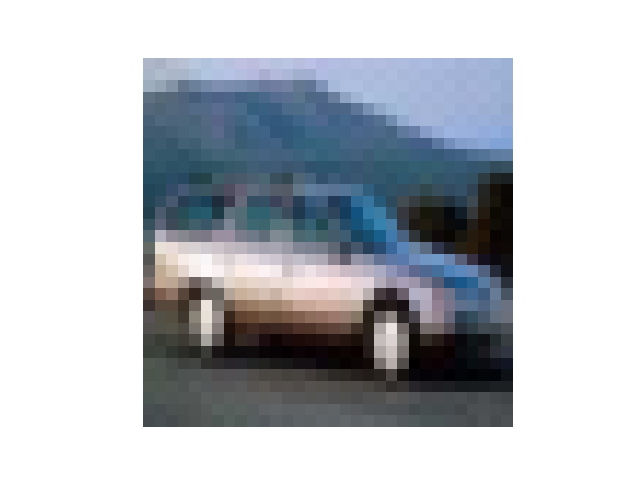}}
		\subcaption{3/255}
	\end{minipage}
	\begin{minipage}[b]{.45\linewidth}
		\centering
		\centerline{\includegraphics[width=\linewidth]{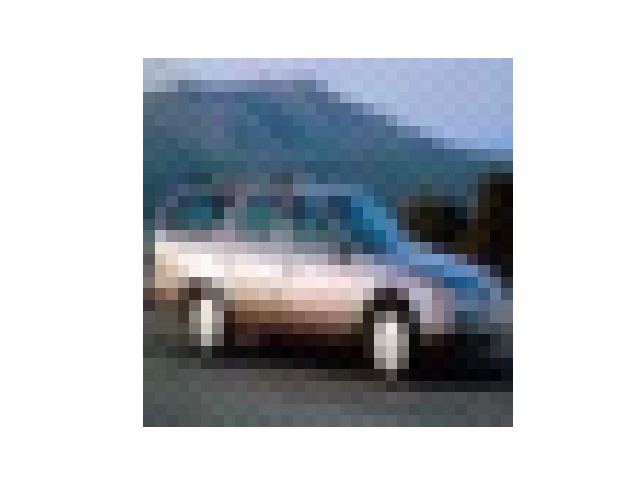}}
		\subcaption{5/255}
	\end{minipage}
	\begin{minipage}[b]{.45\linewidth}
		\centering
		\centerline{\includegraphics[width=\linewidth]{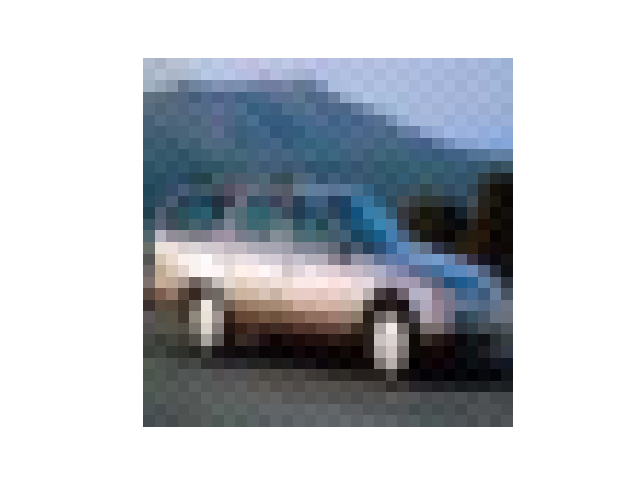}}
		\subcaption{7/255}\label{subfig:attack_strength_bd_image_7}
	\end{minipage}
	\begin{minipage}[b]{.45\linewidth}
		\centering
		\centerline{\includegraphics[width=\linewidth]{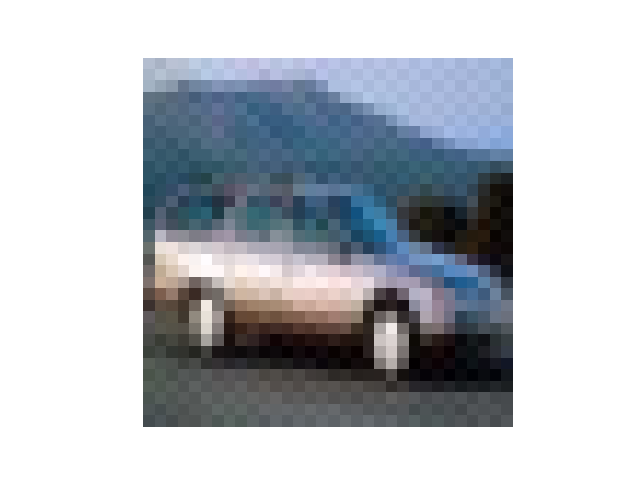}}
		\subcaption{9/255}\label{subfig:attack_strength_bd_image_9}
	\end{minipage}
	\begin{minipage}[b]{.45\linewidth}
		\centering
		\centerline{\includegraphics[width=\linewidth]{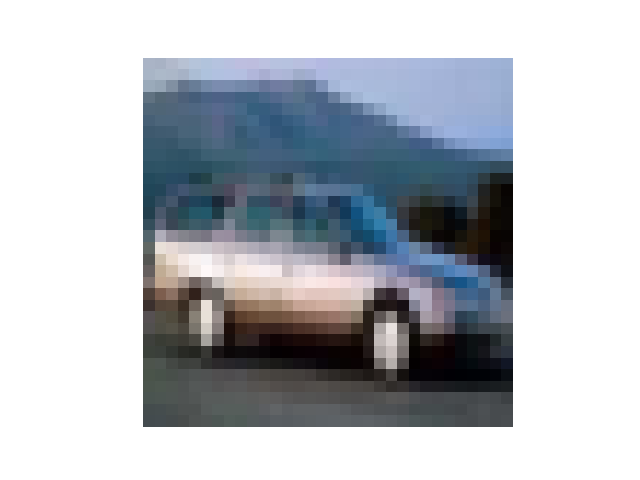}}
		\subcaption{Clean}
	\end{minipage}
	\caption{Example backdoor training images with the chessboard pattern with maximum perturbation size 1/255, 3/255, 5/255, 7/255, and 9/255 embedded; and the clean version of the backdoor training images.}
	\label{fig:attack_strength_bd_image}
\end{figure}

Here, we consider attacks with 3 source classes (3SC) and the ``chessboard'' backdoor pattern (pattern A of the main paper). We create 5 attacks with maximum perturbation size (of the chessboard pattern) 1/255, 3/255, 5/255, 7/255, 9/255. Other attack configurations, including the choices of the source classes and the target class, are the same as in the experiments in the main paper. Example backdoor training images with each of the five patterns embedded are shown in Figure \ref{fig:attack_strength_bd_image}. Note that patterns with perturbation size 7/255 and 9/255 are no longer imperceptible when embedded in the clean images (see the clear mesh grid in Figure \ref{subfig:attack_strength_bd_image_7} and \ref{subfig:attack_strength_bd_image_9}). The ASR and clean test ACC of the five attacks when there is no defense are shown in Table \ref{tab:attack_strength_ASR_ACC}. The DNN architecture used in this section is the wide ResNet architecture.

\begin{table}[t]
	\begin{center}
		\caption{Attack success rate (ASR) and the accuracy (ACC) of the five 3SC attacks with the chessboard pattern with maximum perturbation size 1/255, 3/255, 5/255, 7/255, and 9/255, respectively.}
		\resizebox{0.4\textwidth}{!}{
			\begin{tabular}{ |c|c|c|c|c|c| }
				\hline 
				& 1/255 & 3/255 & 5/255 & 7/255 & 9/255\\
				\hline
				ASR & 93.5 & 99.6 & 100 & 99.5 & 100\\
				\hline
				ACC & 92.2 & 91.9 & 91.4 & 91.6 & 91.8\\
				\hline
			\end{tabular}\label{tab:attack_strength_ASR_ACC}}
	\end{center}
\end{table}

\begin{table}[t]
	\begin{center}
		\caption{Training set cleansing true positive rate (TPR) and false positive rate (FPR) of RE for the five 3SC attacks with the chessboard pattern with maximum perturbation size 1/255, 3/255, 5/255, 7/255, and 9/255, respectively.}
		\resizebox{0.4\textwidth}{!}{
			\begin{tabular}{ |c|c|c|c|c|c| }
				\hline 
				& 1/255 & 3/255 & 5/255 & 7/255 & 9/255\\
				\hline
				TPR & 95.8 & 97.2 & 89.0 & 82.5 & 71.3\\
				\hline
				FPR & 3.1 & 8.4 & 7.7 & 12.2 & 2.4\\
				\hline
			\end{tabular}\label{tab:attack_strength_cleansing}}
	\end{center}
\end{table}

We use the same configurations as in the main paper for our RE defense. Although our defense is not designed for attacks with perceptible perturbations (with large norm), all five attacks are successfully detected by RE with order statistic p-values ``underflow'' (less than $10^{-323}$), with also the target class correctly inferred. In Table \ref{tab:attack_strength_cleansing}, we show the TPR and FPR of training set cleansing for the five attacks. As we expected, there is a slight drop in TPR when the attack becomes perceptible.

\section{Performance Evaluation of RE against More Sophisticated Backdoor Attacks}

%We evaluate our RE defense against a more recent backdoor attack proposed by \cite{Haoti}, which has a more sophisticated design than a typical backdoor attack. For a prescribed source and target class pair $(s^{\ast}, t^{\ast})$, the attacker aims to find an imperceptible backdoor pattern with constrained norm, 
%DJM -- key thing you have to say is that this attack assumes knowledge of the classifier
%DJM or a surrogate, which allows them to craft more evasive backdoors -- you don't mention this
%DJM -- which is easier to be learned ??  what does that mean ?  It can be learned with fewer
%DJM poisoned images, because the attacker is exploiting knowledge of the classifier....
%which is easier to be learned than most of patterns (e.g. pattern A-F in the main paper) during classifier training when embedded in backdoor training images. With this imperceptible backdoor pattern, fewer backdoor training images will be required to poison the training set to have the classifier learn to classify to class $t^{\ast}$ when the same pattern is embedded in clean test images from class $s^{\ast}$.

% Zhen -- re-wrote as the following
We evaluate our RE defense against a more recent backdoor attack proposed by \cite{Haoti}, which leverages the knowledge of the training set and the training configurations (including the DNN model to be used by the learner) to produce an imperceptible backdoor pattern. This pattern is adaptive to the source class and the target class chosen by the attacker, such that fewer backdoor training images will be required to poison the training set than using, e.g., any of patterns A-F in the main paper, to achieve a successful attack.

To launch such an attack, the attacker first train a surrogate classifier $f(\cdot;\tilde{\Theta})$ on the clean training set ${\mathcal D}_{\rm train}$, using the same DNN architecture and training configurations that will be used by the learner. The attacker initializes a pattern ${\bf v}={\bf 0}$ and updates it iteratively. In each iteration, all the training image labeled to the source class $s^{\ast}$ are {\it sequentially} visited. In each visit, an increment to the pattern ${\bf v}$ is estimated by
\begin{equation}\label{eq:opt_Haoti}
\begin{aligned}
& \underset{\bf u}{\text{minimize}}
& & ||{\bf u}||_2 \\
& \text{subject to}
& & f([{\bf x}+{\bf v}+{\bf u}]_c;\tilde{\Theta}) = t^{\ast}
\end{aligned}
\end{equation}
where $t^{\ast}$ is the target class specified by the attacker; ${\bf x}$ is a training image labeled to the source class $s^{\ast}$ that is currently being visited; $[\cdot]_c$ is the clipping operation that is defined in the main paper. The above optimization problem is solved by the ``DeepFool'' algorithm proposed by \cite{DeepFool}; and the optimal solution is denoted by $\Delta{\bf v}$. Each increment obtained is added to the pattern ${\bf v}$ as ${\bf v}\leftarrow{\bf v} + \Delta{\bf v}$ and the updated ${\bf v}$ is projected to a $l_{\rm 2}$ ball with a small radius $\xi$, such that the pattern remains {\it imperceptible}.

At convergence, the obtained adaptive backdoor pattern ``pushes'' the training images labeled to the source class $s^{\ast}$ to the decision boundary between class $s^{\ast}$ and class $t^{\ast}$ specified by the surrogate classifier $f(\cdot;\tilde{\Theta})$ (by solving \ref{eq:opt_Haoti} for these training images). Due to the norm constraint on the adaptive backdoor pattern which is satisfied by projection onto the $l_{\rm 2}$ ball with radius $\xi$, test images labeled to class $s^{\ast}$ will not be guaranteed to be classified to class $t^{\ast}$ by any classifier trained on the clean training set when the backdoor pattern is embedded. Hence the attacker still needs to use this adaptive backdoor pattern to create backdoor training images to poison the training set. During training, the backdoor mapping will be easier to be learned, since the backdoor training images with this adaptive backdoor pattern are closer to the clean training images labeled to the target class than backdoor training images with non-adaptive backdoor patterns (like pattern A-F in the main paper).

%As our main purpose here is to evaluate our defense, we maximize the capability of the attacker by considering a ``full knowledge'' scenario where the attacker has perfect knowledge of the training set and the training configurations (including the DNN model to be used) (see \cite{Haoti} for more details). The attack is devised by first training a surrogate classifier on the clean training set. Then a ``universal'' additive perturbation that ``pushes'' all the training images from class $s^{\ast}$ to the decision boundary between class $s^{\ast}$ and class $t^{\ast}$ (by applying the perturbation to these images) is obtained using the method proposed in \cite{DeepFool_Univ}. However, different from \cite{DeepFool_Univ} which aims to devise a test-time evasion (TTE) attack, the perturbation obtained here will not induce high group misclassification from $s^{\ast}$ to $t^{\ast}$, but will slightly increase the posterior of class $t^{\ast}$, when applied to the clean training images from class $s^{\ast}$. In such a way, the pattern will be easier to be learned when embedded in backdoor training images as the backdoor pattern {\it and} it will have small norm; hence it is imperceptible.
%DJM above paragraph is not clear at all -- how is backdoor mapping learned if perturbation only
%slightly increases P[t |x], rather than actually inducing a misclassification ??

\begin{figure}[t]
	\centering
	\begin{minipage}[b]{.45\linewidth}
		\centering
		\centerline{\includegraphics[width=\linewidth]{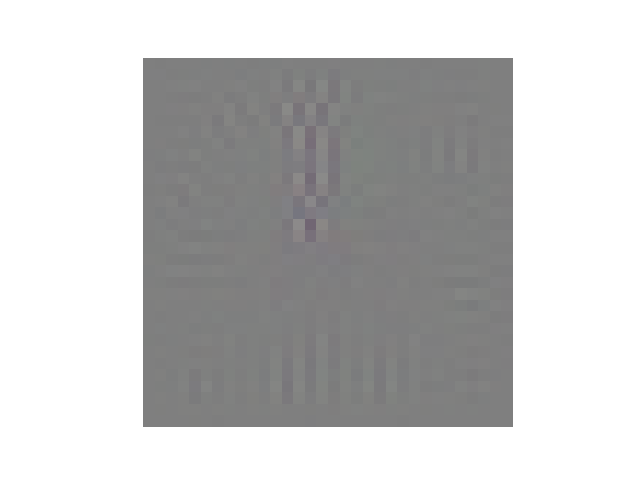}}
		\subcaption{true pattern}
	\end{minipage}
	\begin{minipage}[b]{.45\linewidth}
		\centering
		\centerline{\includegraphics[width=\linewidth]{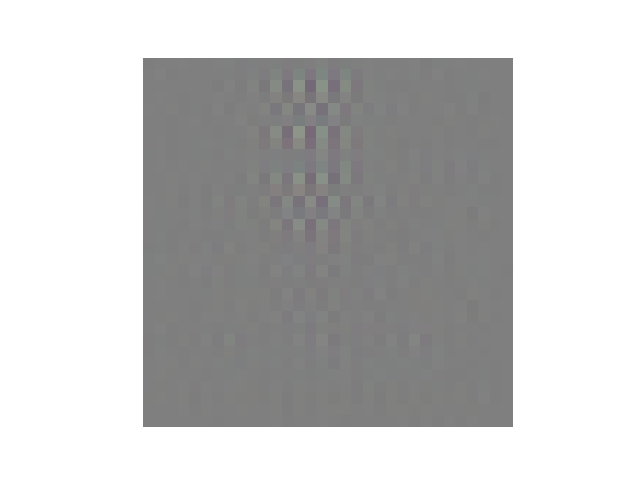}}
		\subcaption{estimated pattern}
	\end{minipage}
	\caption{Left: imperceptible backdoor pattern obtained using the method in \cite{Haoti}. Right: pattern estimated for the detected backdoor class pair by our RE.}
	\label{fig:bd_pattern_advanced_attack}
\end{figure}

Here, we create such an attack on CIFAR-10. We arbitrarily choose the source class and the target class as `frog' and `deer', respectively. We separately train a surrogate classifier (for obtaining the adaptive backdoor pattern) and benchmark classifier (for estimating a benchmark ``attack success rate (ASR)'')\footnote{Although test images labeled to class $s^{\ast}$ will not be guaranteed to be classified to class $t^{\ast}$ by a classifier trained on the clean training set when the backdoor pattern is embedded, there can still be some test images labeled to class $s^{\ast}$ being classified to class $t^{\ast}$ when the backdoor pattern is embedded. This benchmark classifier is trained to estimate a benchmark ``ASR'', i.e. the percentage of test images from class $s^{\ast}$ being classified to class $t^{\ast}$ by a clean classifier when the backdoor pattern is embedded, when there is no backdoor attack. Later on, we will apply our defense to this attack and retrain a classifier. We will expect the ASR of the retrained classifier to be not clearly greater than this benchmark ``ASR''.}, on the clean CIFAR-10 training set, with the wide ResNet architecture and the same training configurations as in the main paper. 
%DJM -- removed "both" in above sentence
We apply the attack crafting approach in \cite{Haoti} on the surrogate classifier and the clean CIFAR-10 training set, with norm constraint $||{\bf v}^{\ast}||_2=0.5$. The obtained pattern is shown on the left of Figure \ref{fig:bd_pattern_advanced_attack}. When embedding this pattern into the clean training images from the source class, the average posterior of the target class is 0.206. When embedding this pattern to the clean test images (of CIFAR-10) from the source class, the misclassification rate to the target class is 20.5\% and 10.1\% for the surrogate classifier and the benchmark classifier, respectively. The difference between the two misclassification rates is related to the ``transferability'' of TTE attacks \cite{Papernot3}. Here, since the classifiers trained by the learner/defender are independent of the surrogate classifier trained by the attacker, we use the misclassification rate 10.1\% (of the benchmark classifier which is also trained independently from the surrogate classifier) as the benchmark ``ASR''.

Again, we first show the effectiveness of the attack when there is no defense. With 500 backdoor training images inserted in the training set, the ASR and the clean test accuracy (ACC) of the classifier trained using the wide ResNet DNN architecture and the same training configurations as in the main paper are 92.2\% and 92.2\% respectively -- the attack is successful. Then we apply our RE defense with the same configurations specified in the main paper. Our RE successfully detects the attack with order statistic p-value $1.2\times10^{-13}$ ($\ll\theta=0.5$); and both the source class and the target class are successfully inferred. The TPR and FPR of training set cleansing are 96.4\% and 6.5\%, respectively. The estimated pattern, with squared error per pixel $6.1\times10^{-5}$ is shown on the right of Figure \ref{fig:bd_pattern_advanced_attack}. After removing the detected training images, we retrain a classifier using the same configuration as training the defenseless DNN. The ASR and ACC are 92.0\% and 6.7\%, respectively -- the ASR is even lower than the benchmark ``ASR''.

\section{Experiments on More Datasets}

In this section, we evaluate our RE defense on more datasets, including MNIST \cite{MNIST}, Fashion-MNIST (F-MNIST) \cite{FashionMNIST}, GTSRB \cite{GTSRB}, and CIFAR-100 \cite{CIFAR10}. 

\subsection{Datasets and Training Configurations}

MNIST and F-MNIST both contain $28\times28$ gray scale images from 10 classes. GTSRB contains colored images of street signs from 43 classes. We resize all the images in GTSRB to $32\times32$. CIFAR-100 contains $32\times32$ colored images from 100 classes. The details of each dataset are summarized in Table \ref{tab:datasets_details}.

\begin{table}[t]
	\begin{center}
		\caption{Details of MNIST, Fashion-MNIST (F-MNIST), GTSRB, and CIFAR-100 datasets.}
		\resizebox{0.45\textwidth}{!}{
			\begin{tabular}{ |c|c|c|c|c| }
				\hline 
				& MNIST & F-MNIST & GTSRB & CIFAR-100\\
				\hline
				image size & $28\times28$ & $28\times28$ & $32\times32$ & $32\times32$\\
				\hline
				colored & no & no & yes & yes\\
				\hline
				no. class & 10 & 10 & 43 & 100\\
				\hline
				training size & 60,000 & 60,000 & 26,640 & 50,000\\
				\hline
				test size & 10,000 & 10,000 & 12,630 & 10,000\\
				\hline
			\end{tabular}\label{tab:datasets_details}}
	\end{center}
\end{table}

\begin{table}[t]
	\begin{center}
		\caption{Training configurations for each of MNIST, F-MNIST, GTSRB, and CIFAR-100; and the resulting (clean) test accuracy (ACC) when there is no attack.}
		\resizebox{0.45\textwidth}{!}{
			\begin{tabular}{ |c|c|c|c|c| }
				\hline 
				& MNIST & F-MNIST & GTSRB & CIFAR-100\\
				\hline
				{\thead{DNN\\architecture}} & LeNet-5 & VGG-9 & {\thead{DNN\\in \cite{NC}}} & ResNet-34\\
				\hline
				epoch & 60 & 60 & 100 & 120\\
				\hline
				{\thead{batch\\size}} & 256 & 256 & 64 & 32\\
				\hline
				{\thead{learning\\rate}} & $10^{-2}$ & $10^{-2}$ & $10^{-3}$ & $10^{-3}$\\
				\hline
				{\thead{data\\augmentation}} & no & no & no & yes\\
				\hline
				ACC & 99.1\% & 90.8\% & 98.7\% & 72.8\%\\
				\hline
			\end{tabular}\label{tab:datasets_training}}
	\end{center}
\end{table}

We use LeNet-5 \cite{MNIST} DNN architecture for training on MNIST. Training is performed for 60 epochs, with batch size 256 and learning rate $10^{-2}$, without data augmentation, and achieves 99.1\% test accuracy when there is no attack. For F-MNIST, we use a ``VGG-9'' DNN architecture, which is customized by removing the last two convolutional layers of VGG-11 \cite{VGG} and using 16, 32, 64, 64, 128, 128 filters in the six remaining convolutional layers, respectively. Training is performed for 60 epochs with batch size 256 and learning rate $10^{-2}$, without data augmentation, and achieves 90.8\% test accuracy when there is no attack. For GTSRB, we use the same DNN architecture used in \cite{NC}. Training is performed for 100 epochs with batch size 64 and learning rate $10^{-3}$, without data augmentation, and achieved 98.7\% test accuracy when there is no attack. For CIFAR-100, we use the standard ResNet-34 architecture \cite{ResNet}. Training is performed for 120 epochs with batch size 32 and learning rate $10^{-3}$ (with 90\% decay starting from each of epoch 60, 100, 110), and achieved 72.8\% test accuracy when there is no attack. Data augmentation is used for training on CIFAR-100, including random cropping and random horizontal flipping.

We use the cross entropy loss function for the training for all the four datasets. For MNIST and F-MNIST, we use stochastic gradient descent (SGD) optimizer with momentum 0.9. For GTSRB and CIFAR-100, we use Adam optimizer with decay rate 0.9 and 0.999 for the first and second moment, respectively. The training configurations are summarized in Table \ref{tab:datasets_training}.

\subsection{Attacks Crafting}

\begin{table}[t]
	\begin{center}
		\caption{Source class(es), target class, and the number of backdoor training images per source class for the attacks on MNIST, F-MNIST, GTSRB, and CIFAR-100.}
		\resizebox{0.45\textwidth}{!}{
			\begin{tabular}{ |c|c|c|c|c| }
				\hline 
				& MNIST & F-MNIST & GTSRB & CIFAR-100\\
				\hline
				{\thead{source\\class(es)}} & 5, 7, 10 & 1 & {\thead{14, ..., 18\\25, ..., 31}} & 1, ..., 20\\
				\hline
				{\thead{target\\class}} & 9 & 3 & 9 & 21\\
				\hline
				{\thead{backdoor\\training\\images per\\source class}} & 500 & 900 & 30 & 30\\
				\hline
			\end{tabular}\label{tab:datasets_attack_configurations}}
	\end{center}
\end{table}

\begin{figure}[t]
	\centering
	\begin{minipage}[b]{\linewidth}
		\centering
		\includegraphics[width=0.32\linewidth]{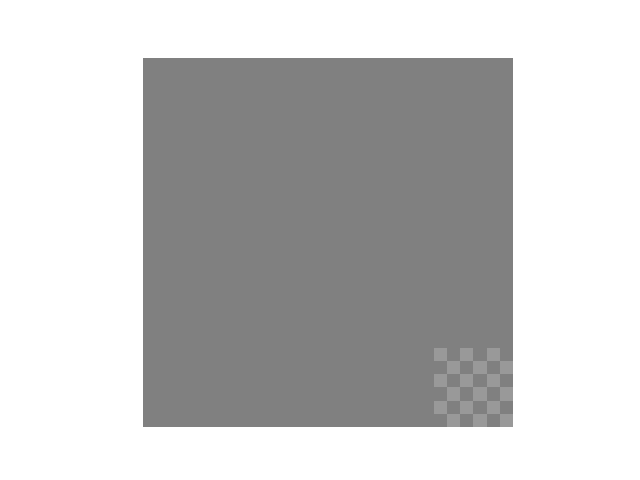}
		\includegraphics[width=0.32\linewidth]{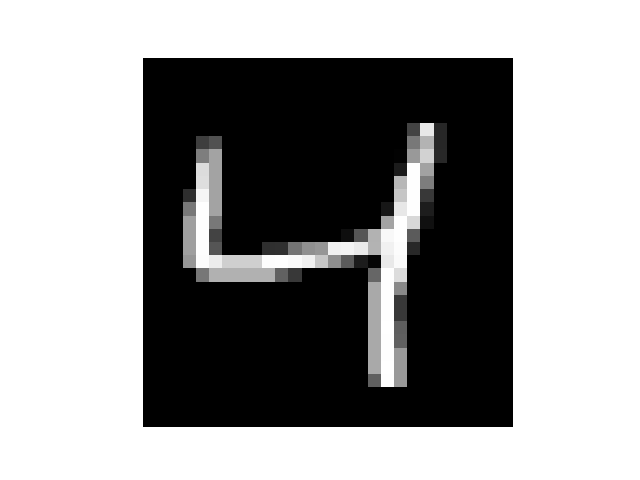}
		\includegraphics[width=0.32\linewidth]{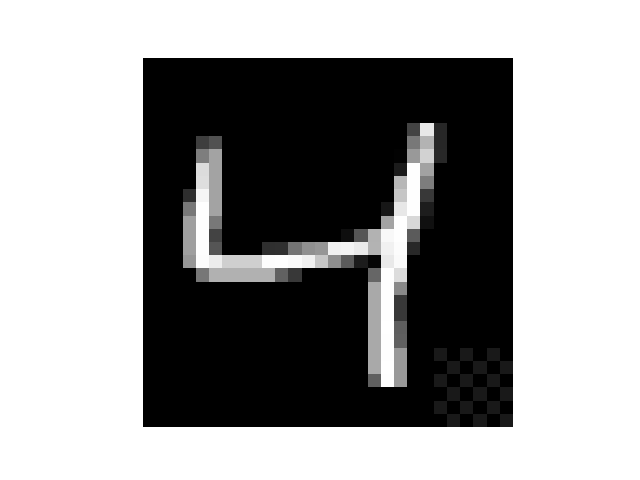}
		\subcaption{MNIST}
	\end{minipage}
	\begin{minipage}[b]{\linewidth}
		\centering
		\includegraphics[width=0.32\linewidth]{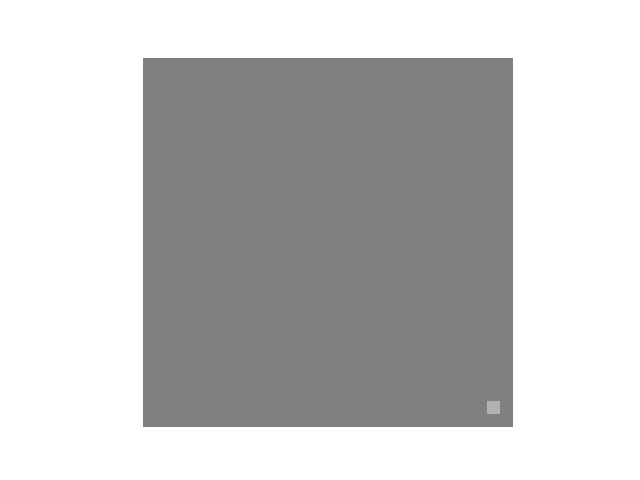}
		\includegraphics[width=0.32\linewidth]{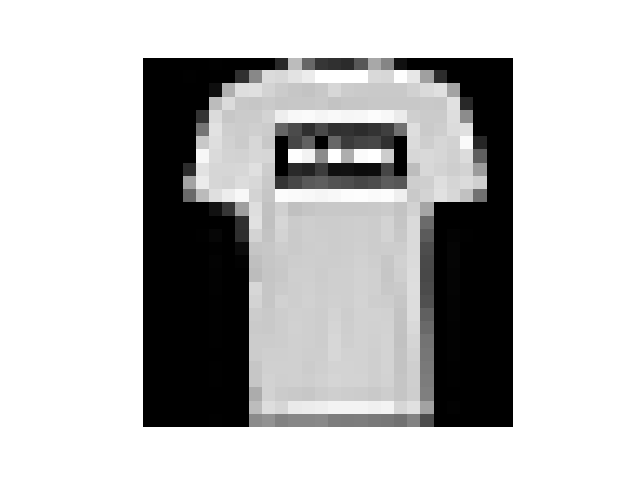}
		\includegraphics[width=0.32\linewidth]{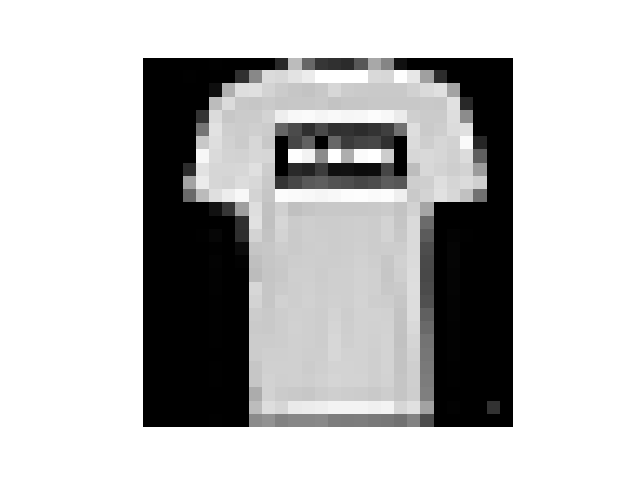}
		\subcaption{F-MNIST}
	\end{minipage}
	\begin{minipage}[b]{\linewidth}
		\centering
		\includegraphics[width=0.32\linewidth]{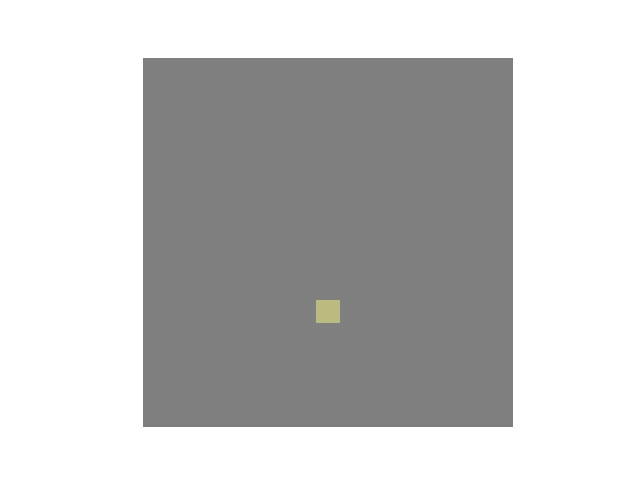}
		\includegraphics[width=0.32\linewidth]{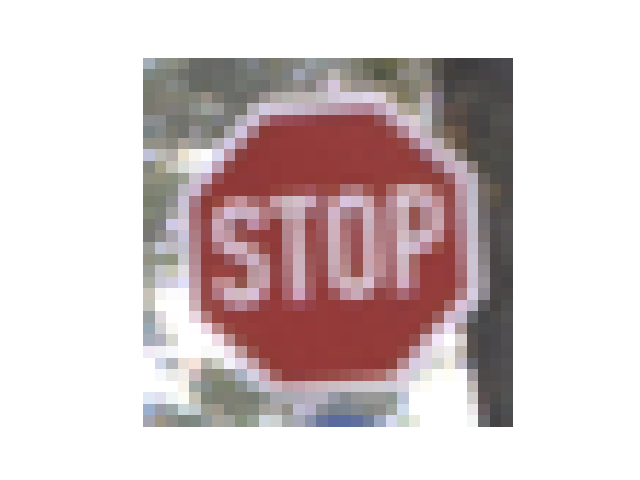}
		\includegraphics[width=0.32\linewidth]{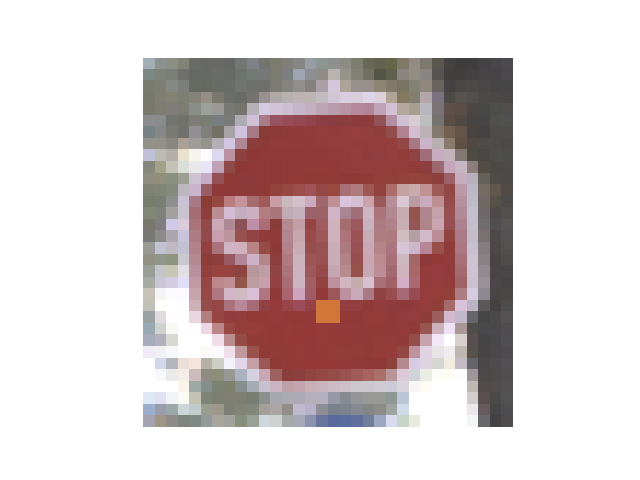}
		\subcaption{GTSRB}
	\end{minipage}
	\begin{minipage}[b]{\linewidth}
		\centering
		\includegraphics[width=0.32\linewidth]{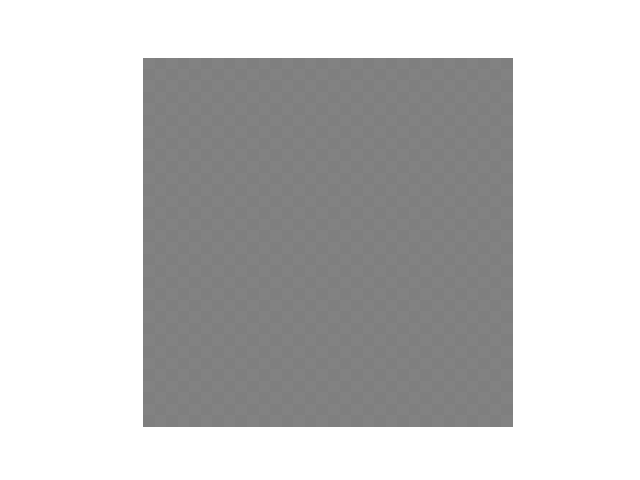}
		\includegraphics[width=0.32\linewidth]{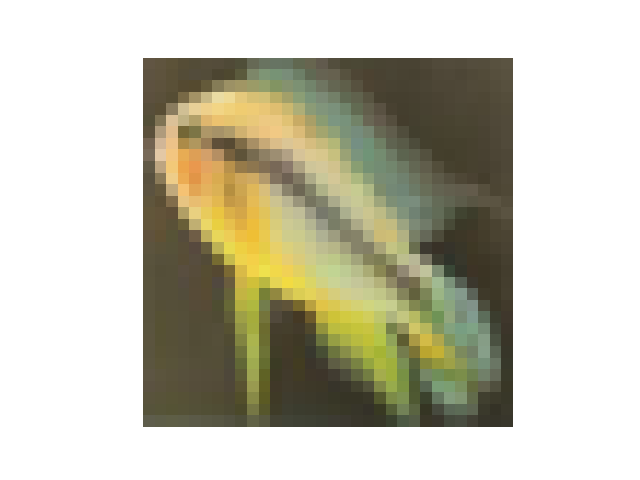}
		\includegraphics[width=0.32\linewidth]{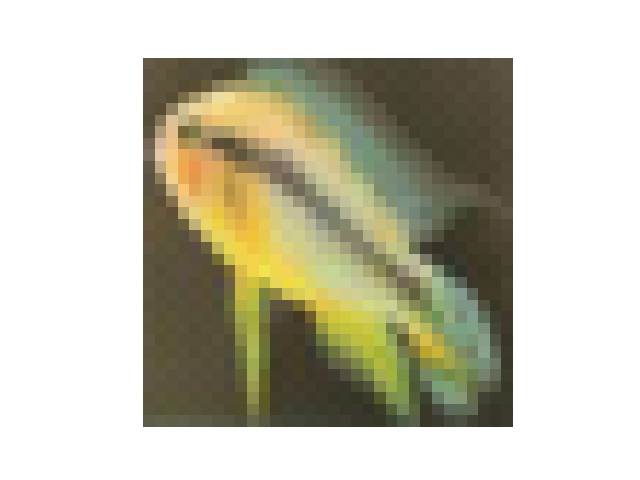}
		\subcaption{CIFAR-100}
	\end{minipage}
	\caption{Backdoor pattern (with 0.5 offset) (left), example backdoor training image (right), and originally clean image (middle), for MNIST, F-MNIST, GTSRB, and CIFAR-100.}
	\label{fig:datasets_bd_patterns_images}
\end{figure}

\begin{table}[t]
	\begin{center}
		\caption{Attack success rate (ASR) and clean test accuracy (ACC) of the classifiers trained on the poisoned training set of MNIST, F-MNIST, GTSRB, and CIFAR-100, when there is no defense.}
		\resizebox{0.45\textwidth}{!}{
			\begin{tabular}{ |c|c|c|c|c| }
				\hline 
				& MNIST & F-MNIST & GTSRB & CIFAR-100\\
				\hline
				ASR & 96.3 & 91.7 & 87.8 & 98.0\\
				\hline
				ACC & 98.9 & 91.0 & 98.1 & 72.4\\
				\hline
			\end{tabular}\label{tab:datasets_ASR_ACC}}
	\end{center}
\end{table}

We create one attack for each of the four datasets. The source class(es), target class, and the number of backdoor training images per source class are summarized in Table \ref{tab:datasets_attack_configurations}. For MNIST, F-MNIST, and CIFAR-100, the class indices used here (starting from 1) follow the label assignment specified by ``torchvision.dataset''\footnote{\url{https://pytorch.org/docs/stable/torchvision/datasets.html}}. For GTSRB, the class labels are specified online\footnote{\url{http://benchmark.ini.rub.de/?section=gtsrb\&subsection=news}}. Especially for the GTSRB dataset of trafic signs, we set the source classes to be those traffic signs related to ``using cautious'', e.g. a stop sign, a yield sign, etc., and the target class is chosen as a speed limit sign of 120km/h. If an autonomous vehicle with a traffic sign recognizer/classifier undergoes such an attack, during testing, it will likely recognize a, e.g. stop sign with the backdoor pattern (that will be shown next), as a speed limit sign of 120km/h; and it will speed up instead stop, which may cause a catastrophe.

In Figure \ref{fig:datasets_bd_patterns_images}, we show the backdoor patterns (with 0.5 offset) used to attack the four datasets, with example backdoor training images and their originally clean images. The backdoor pattern for MNIST is a small patch ($6\times6$) of ``chessboard'' located at the bottom right corner, with perturbation size 25/255. The backdoor pattern for F-MNIST is a pixel perturbation located at the bottom right corner with perturbation size 50/255. The backdoor pattern for GTSRB is a yellow square which mimics a sticker that can be physically attached to a traffic sign. To digitally embed such a pattern to a traffic sign image, we perturb in both the `red' and the `green' channel with perturbation size 60/255. Lastly, for CIFAR-100, we use the image-wide ``chessboard'' pattern, i.e. pattern A in the main paper, with perturbation size 3/255.

In Table \ref{tab:datasets_ASR_ACC}, we show the ASR and the clean test ACC of the classifiers trained on the four poisoned training sets, when there is no defense. All the attacks are successful, with high ASR and no clear degradation in the clean test ACC.

\begin{figure*}[t]
	\centering
	\begin{minipage}[b]{0.45\linewidth}
		\centering
		\includegraphics[width=\linewidth]{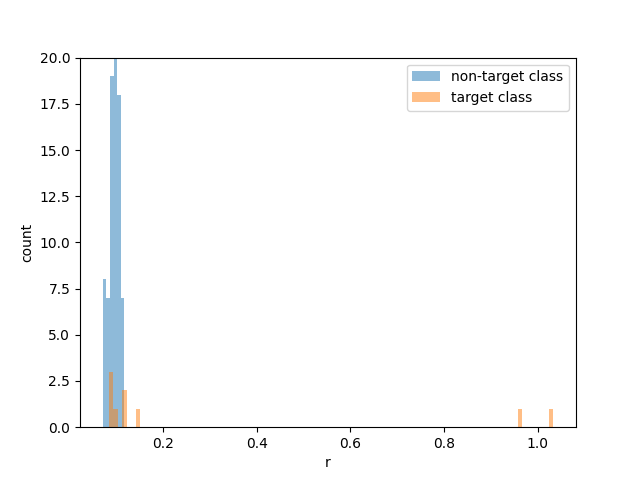}
		\subcaption{MNIST}
	\end{minipage}
	~
	\begin{minipage}[b]{0.45\linewidth}
		\centering
		\includegraphics[width=\linewidth]{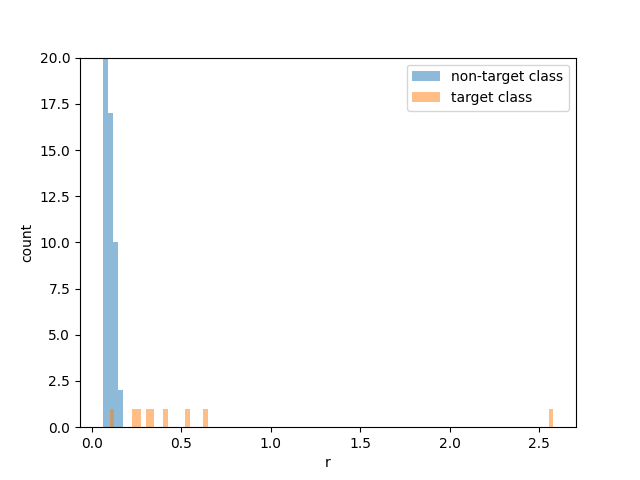}
		\subcaption{F-MNIST}
	\end{minipage}
	\begin{minipage}[b]{0.45\linewidth}
		\centering
		\includegraphics[width=\linewidth]{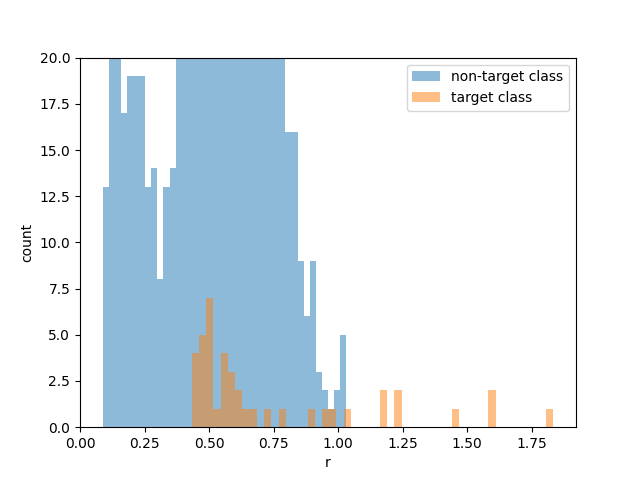}
		\subcaption{GTSRB}
	\end{minipage}
	~
	\begin{minipage}[b]{0.45\linewidth}
		\centering
		\includegraphics[width=\linewidth]{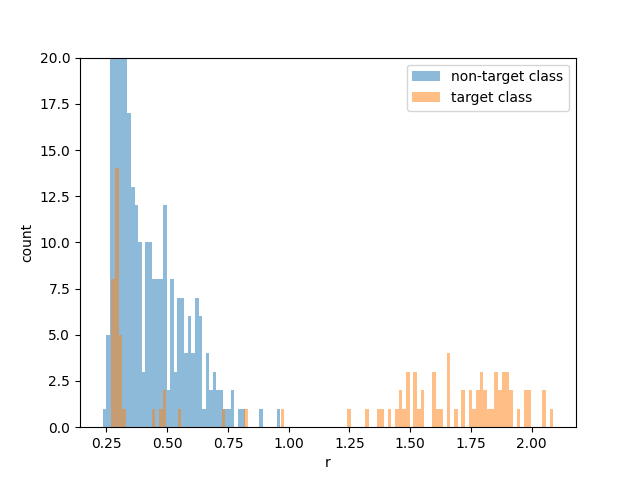}
		\subcaption{CIFAR-100}
	\end{minipage}
	\caption{Histogram of the reciprocal statistics when RE is applied against the attacks on MNIST, F-MNIST, GTSRB, and CIFAR-100, respectively.}
	\label{fig:datasets_r_hist}
\end{figure*}

\subsection{Defense Performance Evaluation}

As described in the main paper, as part of our RE defense, a classifier is trained on the training set which could be poisoned. For each attack, we perform such training using the same DNN architecture and training configurations as specified in Table \ref{tab:datasets_training}; except for CIFAR-100, we do not use data augmentation here.

Then we apply RE to the four attacks with the {\it same} (default) configurations as specified in the main paper, i.e. $\delta=10^{-4}$, $\pi=0.95$, and $\lambda=1$. As a benchmark for evaluating the detection performance, for each dataset, we also apply our RE when there is no backdoor poisoning. Our RE successfully detects all four attacks and infers no attack for the four clean datasets. Moreover, in Figure \ref{fig:datasets_r_hist}, for each attack, we show the histogram of the reciprocal statistics $r_{st}$ for the $K(K-1)$ $(s, t)$ class pairs. Clearly, for all four attacks, when the training set is backdoor poisoned, there is at least one reciprocal statistic corresponding to the {\it true backdoor} target class that is an outlier to the null model fitted with the reciprocal statistics corresponding to all the other non-backdoor target classes -- the success of RE detection is visualized.

\begin{table}[t]
	\begin{center}
		\caption{True positive rate (TPR) and false positive rate (FPR) of training set cleansing when applying RE against the attacks on MNIST, F-MNIST, GTSRB, and CIFAR-100, respectively.}
		\resizebox{0.45\textwidth}{!}{
			\begin{tabular}{ |c|c|c|c|c| }
				\hline 
				& MNIST & F-MNIST & GTSRB & CIFAR-100\\
				\hline
				TPR & 97.3 & 95.4 & 99.7 & 93.5\\
				\hline
				FPR & 1.4 & 5.8 & 4.8 & 5.2\\
				\hline
			\end{tabular}\label{tab:datasets_TPR_FPR}}
	\end{center}
\end{table}

In Table \ref{tab:datasets_TPR_FPR}, we show the TPR and FPR of training set cleansing when RE is applied against the four attacks. Our RE, with the same default settings, achieves both high TPR and low FPR in training set cleansing against these four attacks (against four different datasets).

\begin{table}[t]
	\begin{center}
		\caption{Attack success rate (ASR) and clean test accuracy (ACC) of retrained classifiers after RE training set cleansing, for attacks against MNIST, F-MNIST, GTSRB, and CIFAR-100, respectively.}
		\resizebox{0.45\textwidth}{!}{
			\begin{tabular}{ |c|c|c|c|c| }
				\hline 
				& MNIST & F-MNIST & GTSRB & CIFAR-100\\
				\hline
				ASR & 0.1 & 1.0 & 0.0 & 0.15\\
				\hline
				ACC & 99.1 & 90.8 & 98.6 & 71.8\\
				\hline
			\end{tabular}\label{tab:datasets_ASR_ACC_retrain}}
	\end{center}
\end{table}

After removing the detected training images, again, we retrain a classifier for each dataset using the same DNN architectures and training configurations specified in Table \ref{tab:datasets_training}. In Table \ref{tab:datasets_ASR_ACC_retrain}, we show that all the retrained classifiers have both low ASR and almost no degradation in clean test ACC compared with classifier trained on clean, unpoisoned training set (see the last row of Table \ref{tab:datasets_training}). {\it Our RE is generally effective in protecting these datasets from being backdoor poisoned.}

\begin{figure}[t]
	\centering
	\begin{minipage}[b]{\linewidth}
		\centering
		\includegraphics[width=0.32\linewidth]{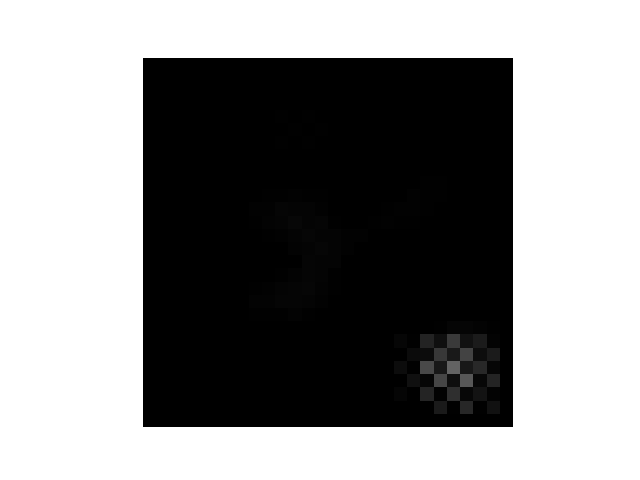}
		\includegraphics[width=0.32\linewidth]{figures/more_dataset/MNIST_with_bd.png}
		\includegraphics[width=0.32\linewidth]{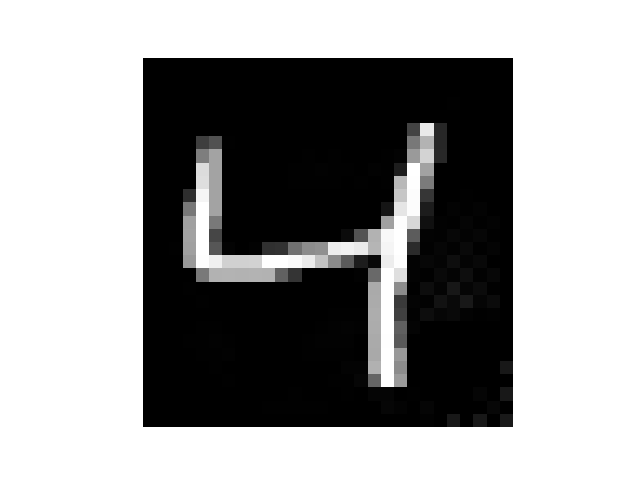}
		\subcaption{MNIST}
	\end{minipage}
	\begin{minipage}[b]{\linewidth}
		\centering
		\includegraphics[width=0.32\linewidth]{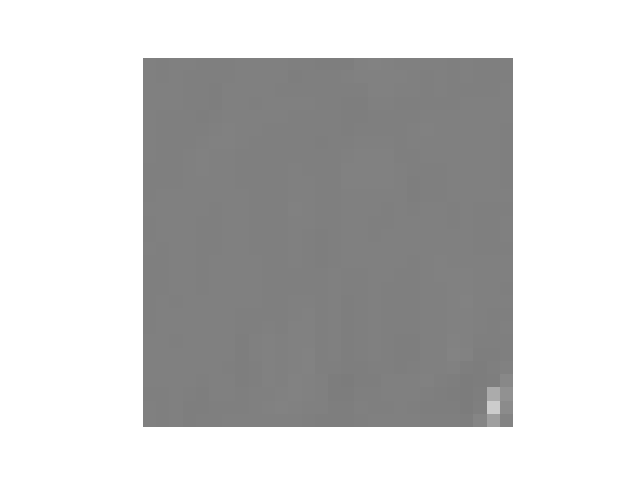}
		\includegraphics[width=0.32\linewidth]{figures/more_dataset/F-MNIST_with_bd.png}
		\includegraphics[width=0.32\linewidth]{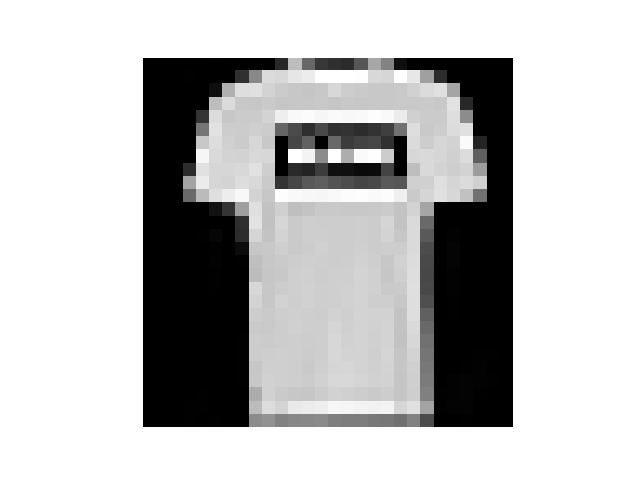}
		\subcaption{F-MNIST}
	\end{minipage}
	\begin{minipage}[b]{\linewidth}
		\centering
		\includegraphics[width=0.32\linewidth]{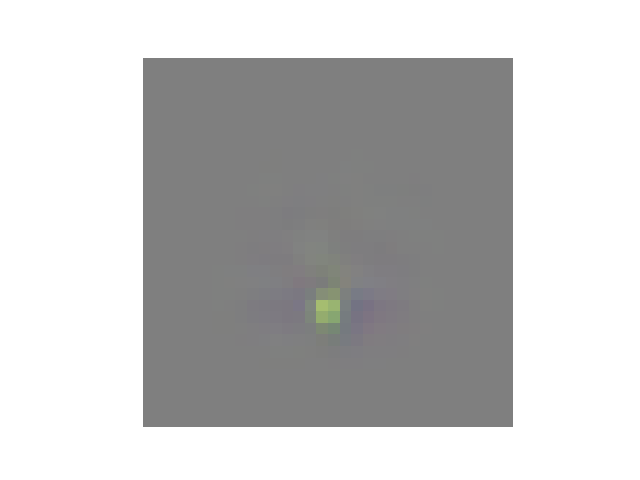}
		\includegraphics[width=0.32\linewidth]{figures/more_dataset/GTSRB_with_bd.png}
		\includegraphics[width=0.32\linewidth]{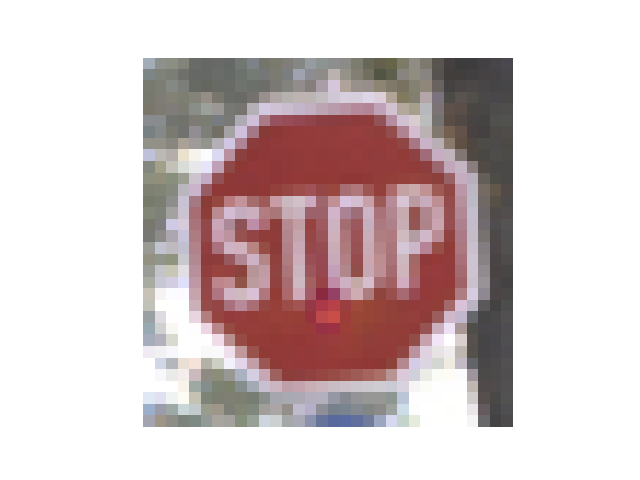}
		\subcaption{GTSRB}
	\end{minipage}
	\begin{minipage}[b]{\linewidth}
		\centering
		\includegraphics[width=0.32\linewidth]{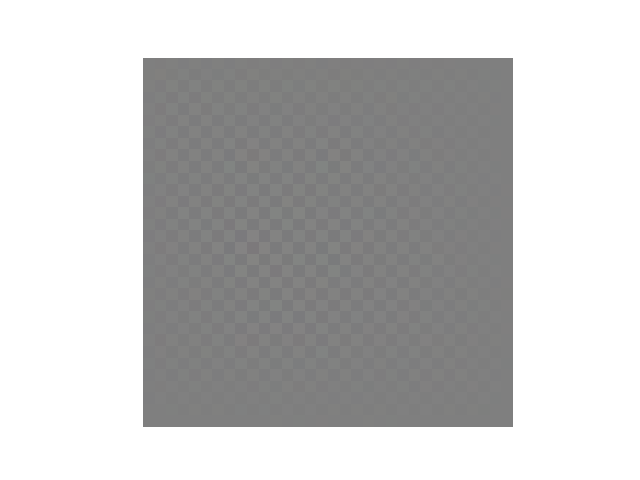}
		\includegraphics[width=0.32\linewidth]{figures/more_dataset/CIFAR100_with_bd.png}
		\includegraphics[width=0.32\linewidth]{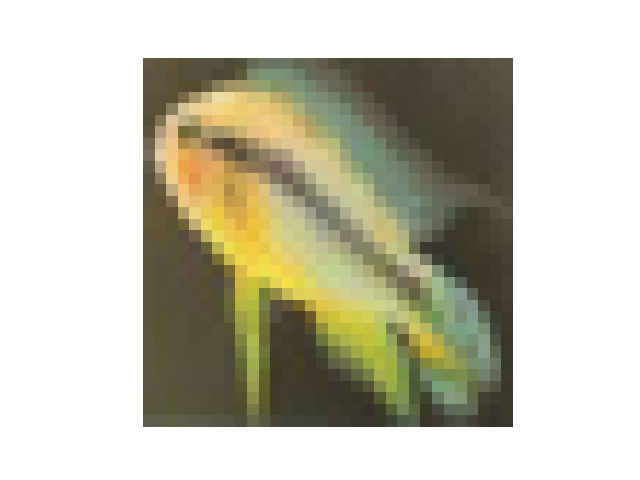}
		\subcaption{CIFAR-100}
	\end{minipage}
	\caption{Backdoor pattern (with 0.5 offset) estimated by RE (left), example backdoor training image (middle), and the image with the estimated pattern removed (right), for attacks against MNIST, F-MNIST, GTSRB, and CIFAR-100, respectively.}
	\label{fig:datasets_rm}
\end{figure}

Lastly, we show the backdoor pattern (with 0.5 offset) estimated by RE for each attack in Figure \ref{fig:datasets_rm}. For each attack, we also show a backdoor training image and the image with the estimated backdoor pattern ``removed''. RE successfully estimates and ``remove'' the backdoor pattern embedded in backdoor training images.

\section{Computing Infrastructure and Computational Complexity}

All the experiments in the main paper and this supplementary material are conducted with the following hardware specification:\\
CPU: Intel i9-9900K, 3.60GHz, 16MB\\
GPU: GeForce RTX2080-Ti, 11GB (dual card)

The computational complexity of our RE defense is ${\mathcal O}(K^2 N_{\rm b} N_{\rm i})$ backpropagations, where $K$ is the number of classes, $N_{\rm b}$ is the number of images per mini-batch\footnote{As described in the main paper, in each iteration of pattern estimation, we compute the gradient of Eq. (9) on two batches of training images randomly sampled from ${\mathcal D}_s$ and ${\mathcal D}_t$, respectively.} used in pattern estimation, $N_{\rm i}$ is the number of iterations until $\pi$-level misclassification is achieved. Using the hardware specified above, the average time cost for RE against the 18 attacks on CIFAR-10, when the wide ResNet architecture is used, is 2775 seconds (not including the initial training on the poisoned training set or the retraining). While the training time when there is no defense is 3060 seconds.

The bottle neck of our defense is its scalability when the number of classes $K$ increases. This is the common limitation of most reverse-engineering-based backdoor defenses, e.g \cite{NC, Tabor, Post-TNNLS}. However, this problem can be conquered by our new technique currently being developed, hence is not further discussed here.

\end{document}